\documentclass[10pt,journal,compsoc]{IEEEtran}

\usepackage{times}
\usepackage{soul}
\usepackage{url}

\usepackage{hyperref}

\hypersetup{
 colorlinks=true,
 linkcolor=red,
 citecolor=cyan,
 filecolor=magenta, 
 urlcolor=cyan,
 }
\usepackage[utf8]{inputenc}
\usepackage[small]{caption}
\usepackage{graphicx}
\usepackage{amsmath}
\usepackage{amsthm}
\usepackage{booktabs}
\usepackage[ruled,linesnumbered]{algorithm2e}
\usepackage{amssymb}
\usepackage{pifont}
\usepackage{times}
\usepackage{epsfig}
\usepackage{amssymb}
\usepackage{amsfonts}
\usepackage{subfigure}
\usepackage{algorithmic}
\usepackage{textcomp}
\usepackage{threeparttable}
\usepackage{multicol}
\usepackage{multirow}
\usepackage{xspace}
\usepackage{bm}
\usepackage{balance}
\usepackage{float}
\usepackage{color}
\usepackage{ragged2e}
\usepackage{amsthm,amsmath,amssymb}
\usepackage{mathrsfs}

\definecolor{qiu}{RGB}{250,10,10}

\def\ie{\emph{i.e.,~}}
\def\eg{\emph{e.g.,~}}

\def\wrt{w.r.t.~}
\def\etal{\emph{et al.~}}

\def\mata{\textcolor{red}}
\def\ournet{T-CPGA}
\definecolor{fan}{RGB}{181,68,52}

\def\newl{\vspace{0.06in}}

\newcommand{\cmark}{\ding{51}}%
\newcommand{\xmark}{\ding{55}}%

\ifCLASSOPTIONcompsoc
 \usepackage[nocompress]{cite}
\else
 \usepackage{cite}
\fi

\ifCLASSINFOpdf
\else
\fi

\begin{document}
\bstctlcite{IEEEexample:BSTcontrol}
%
\title{Imbalance-Agnostic Source-Free Domain Adaptation via Avatar Prototype Alignment}

\author{Hongbin Lin, Mingkui Tan, Yifan Zhang, Zhen Qiu, Shuaicheng Niu, Dong Liu, Qing Du and Yanxia Liu
\IEEEcompsocitemizethanks{
 \IEEEcompsocthanksitem{M. Tan, Y. Zhang and Z. Qiu are co-first authors. Corresponding to Y. Liu.}
 \IEEEcompsocthanksitem{H. Lin, Z. Qiu, S. Niu, D. Liu, Q. Du, Y. Liu and M. Tan are with South China University of Technology, Guangzhou 510641, China (e-mail: \{sehongbinlin, seqiuzhen, sensc, sesmildong\}@mail.scut.edu.cn; \{duqing, cslyx, mingkuitan\}@scut.edu.cn).}
 \IEEEcompsocthanksitem{Y. Zhang is with National University of Singapore, Singapore, 138600 (e-mail: yifan.zhang@u.nus.edu).}
 }
}

\markboth{Journal of \LaTeX\ Class Files,~Vol.~14, No.~8, August~2015}%
{Shell \MakeLowercase{\textit{et al.}}: Bare Advanced Demo of IEEEtran.cls for IEEE Computer Society Journals}
%



\IEEEtitleabstractindextext{%
\begin{abstract}
\justifying
Source-free Unsupervised Domain Adaptation (SF-UDA) aims to adapt a well-trained source model to an unlabeled target domain without access to the source data. One key challenge is the lack of source data during domain adaptation. To handle this, we propose to mine the hidden knowledge of the source model and exploit it to generate source avatar prototypes (\ie representative features for each source class). To this end, we propose a Contrastive Prototype Generation and Adaptation (CPGA) method. 
 CPGA consists of two stages: 1) Prototype generation: by exploring the classification boundary information of the source model, we train a prototype generator to generate source prototypes. 2) Prototype adaptation: based on the prototypes and target pseudo labels, we develop a robust contrastive prototype adaptation strategy to align each pseudo-labeled target data to the corresponding source prototypes. Extensive experiments on three UDA benchmark datasets demonstrate the superiority of CPGA. However, existing SF-UDA studies (including our CPGA) implicitly assume the class distributions of both source and target domains to be balanced. This hinders the applications of existing SF-UDA to real scenarios, in which the class distributions are usually skewed and agnostic. To address this issue, we study a more practical SF-UDA task, termed imbalance-agnostic SF-UDA, where the class distributions of both the unseen source domain and unlabeled target domain are unknown and could be arbitrarily skewed (\eg \textbf{long-tailed, or even inversely long-tailed}). This task is much more challenging than vanilla SF-UDA due to the co-occurrence of covariate shifts and unidentified class distribution shifts between the source and target domains. To address this task, we extend CPGA and propose a new Target-aware Contrastive Prototype Generation and Adaptation (T-CPGA) method. Specifically, for better prototype adaptation in the imbalance-agnostic scenario, T-CPGA applies a new pseudo label generation strategy to identify unknown target class distribution and generate accurate pseudo labels, by utilizing the collective intelligence of the source model and an additional contrastive language-image pre-trained model. Meanwhile, we further devise a target label-distribution-aware classifier to adapt the model to the unknown target class distribution. We empirically show that T-CPGA significantly outperforms CPGA and other SF-UDA methods in imbalance-agnostic SF-UDA, \eg 25.1\% and 22.5\% overall accuracy gains on Cl$\rightarrow$Pr and Cl$\rightarrow$Rw tasks of the imbalance-agnostic Office-Home dataset.
\end{abstract}
 
\begin{IEEEkeywords}
Source-free Unsupervised Domain Adaptation, Agnostic Class Distribution, Feature Prototype.
\end{IEEEkeywords}
}

\maketitle

\IEEEdisplaynontitleabstractindextext

%
\IEEEpeerreviewmaketitle

\ifCLASSOPTIONcompsoc
\IEEEraisesectionheading{\section{Introduction}\label{sec:introduction}}
\else
\section{Introduction}
\label{sec:introduction}
\fi

%
%
%
%
\IEEEPARstart{U}{nsupervised} domain adaptation (UDA) aims to promote the model performance on an unlabeled target domain, by adapting a model trained on a large-scale labeled source dataset to the unlabeled target domain. The key challenge of UDA is the distribution discrepancy between source and target domains~\cite{Sankaranarayanan2018GenerateTA,Hoffman2018CyCADACA}. To address this issue, existing methods propose to align source and target domains either by exploiting diverse discrepancy metrics (\eg maximum mean discrepancy~\cite{long2015learning,long2017deep}, high-order statistics of distributions~\cite{chen2020homm,sun2016deep} and inter/intra class distance~\cite{chen2019joint,Kang2019ContrastiveAN}), or by conducting domain adversarial learning~\cite{ganin2015unsupervised,long2018conditional,Zhang2020CollaborativeUD}.

However, in real-world applications, one may only access a source-trained model instead of source data due to the law of privacy protection~\cite{niu2022efficient,niu2023towards,lin2022prototype}. This makes existing UDA~\cite{dong2021and,li2021divergence,luo2020unsupervised} methods (that rely heavily on source data) fail. To handle this, Source-Free Unsupervised Domain Adaptation (SF-UDA)~\cite{liang2020shot} has been explored recently, where only a source model and unlabeled target data are available. To solve this problem, existing SF-UDA methods propose to refine the source model either by using the source model to generate pseudo-labeled target data (\eg SHOT~\cite{liang2020shot}), or by using generative adversarial networks (GANs)~\cite{goodfellow2014generative} to generate target-style images (\eg MA~\cite{Li2020ModelAU}). However, due to the domain discrepancy, the pseudo labels could be noisy. Moreover, directly generating target-style images is very difficult since GANs are hard to train on a small target dataset~\cite{karras2020training}.

\begin{table*}[t]

\setlength\tabcolsep{10pt}
 \begin{center}
 \caption{\label{tab:setting} Illustration of the related UDA settings. Compared to Source-Free Unsupervised Domain Adaptation (SF-UDA), Imbalanced SF-UDA~\cite{li2021imbalanced} particularly relies on the prior of the source class distribution (\eg label frequency) for training a balance source model, so it is not essentially SF-UDA as it influences the use of source data. In contrast, imbalance-agnostic SF-UDA only accesses an imbalance-agnostic source model without influencing the training of source models. Balance training refers to training a class-uniformed model with the class distribution prior, while standard training trains the source model only via vanilla loss (\eg Cross-Entropy loss).}
 \scalebox{0.84}{
 \begin{tabular}{lcccccccccccccc}
 \toprule
 \multirow{2}{*}{Setting} &
 \multicolumn{2}{c}{Source Model Training}&
 \multicolumn{3}{c}{Adaptation}\cr
 \cmidrule(lr){2-3} \cmidrule(lr){4-6}
 & Class Distribution Prior & Standard / Balance Training & Source Data & Target Data & Class Distribution Shift\cr
 \midrule
 UDA & \xmark & Standard & \cmark & \cmark &
 \begin{minipage}[c]{0.2\columnwidth}
		\centering
		\raisebox{-.5\height}{\includegraphics[width=\linewidth]{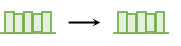}}
	\end{minipage}
 \\ 
 SF-UDA & \xmark & Standard & \xmark & \cmark &
 \begin{minipage}[c]{0.2\columnwidth}
		\centering
		\raisebox{-.5\height}{\includegraphics[width=\linewidth]{fig/11.png}}
	\end{minipage}
 \\ 
 Imbalanced SF-UDA\cite{li2021imbalanced} & Required & Balanced &\xmark &\cmark &
 \begin{minipage}[b]{0.4\columnwidth}
		\centering
		\raisebox{-.5\height}{\includegraphics[width=\linewidth]{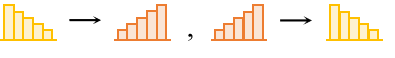}}
	\end{minipage}
 \\ 
 Imbalance-agnostic SF-UDA & \xmark & Standard &\xmark &\cmark &
 \begin{minipage}[c]{0.6\columnwidth}
		\centering
		\raisebox{-.5\height}{\includegraphics[width=\linewidth]{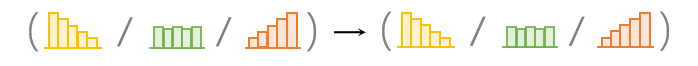}}
	\end{minipage}
 \\
 \bottomrule
 \end{tabular}
 }
 \end{center}
 \vspace{-0.1in}
 
\end{table*}

To handle the absence of source data, our insight is to mine the hidden knowledge within the source model for generating feature prototypes of each source class. In light of this, we propose a Contrastive Prototype Generation and Adaptation (CPGA) method, including two stages: 1) Prototype generation: by exploring the classification boundary information in the source classifier, we train a prototype generator to generate source prototypes based on contrastive learning.
2) Prototype adaptation: to mitigate domain discrepancies, based on the generated feature prototypes and target pseudo labels, we develop a new contrastive prototype adaptation strategy to align each pseudo-labeled target data to the source prototype with the same class. To alleviate label noise, we enhance the alignment via confidence reweighting and early learning regularization. 
Extensive experiments verify the effectiveness and superiority of our CPGA.

Despite the success of CGPA in solving vanilla SF-UDA, as shown in Table~\ref{tab:setting}, conventional SF-UDA methods~\cite{liang2020shot,Li2020ModelAU,xia2021adaptive} 
 implicitly assume that the training data of the source model and the target data follow relatively balanced class distributions. 
Nevertheless, practical data may follow any class distribution, \eg{a long-tailed class distribution~\cite{zhang2021deep,kang2019decoupling,zhang2022self}}. In this scenario, SF-UDA becomes more challenging, and vanilla SF-UDA methods suffer performance degradation due to the issues of class imbalance and unknown class distribution shifts.

To conquer this, ISFDA~\cite{li2021imbalanced} explores handling Imbalanced SF-UDA where the class distributions of both domains are inverse (\eg{long-tailed source domain and inversely long-tailed target domain}) as shown in Table~\ref{tab:setting}. Specifically, it first resorts to the prior of the source class distribution to train a class-balanced model. Then, it conducts label refine curriculum adaptation and representation optimization to overcome the joint presence of covariate and class distribution shifts. However, the class-balanced source model is not always available in real scenarios since it relies on the prior knowledge of the source class distribution. Due to the lack of source data, an imbalance-agnostic source model is more probably given, \ie the source model may be class-biased.
More critically, the target domain is not necessarily following the class distribution that is just inverse to that of the source domain. 

To address these issues, we explore a more practical task, called imbalance-agnostic SF-UDA, where the class distributions of both the unseen source domain and unlabeled target domain are unknown and can be arbitrarily skewed (\eg \textbf{long-tailed, inversely long-tailed}) as shown in Table~\ref{tab:setting}.
In addition to the challenges in SF-UDA, this task poses an additional challenge: it is unknown how to adapt the imbalance-agnostic source model to the unlabeled target domain under unidentified class distribution shifts. Apparently, dealing with the co-occurrence of data distribution shifts and unidentified class distribution shifts is non-trivial, which leads to the performance degradation of existing SF-UDA methods~\cite{li2021imbalanced,Qiu2021CPGA}.
Compared with Imbalanced SF-UDA, imbalance-agnostic SF-UDA does not rely on the source class distribution prior and considers the existence of unidentified class distribution shifts.
 
To handle imbalance-agnostic SF-UDA, we extend CPGA and propose a new Target-aware Contrastive Prototype Generation and Adaptation (\ournet) method. To alleviate the negative effect of unidentified class distribution shifts, we are motivated to leverage the zero-shot prediction abilities of CLIP (Contrastive Language-Image Pre-training)~\cite{radford2021learning} to help identify unknown target class distribution. Specifically, we aggregate the knowledge of the source model and CLIP to perceive the unlabeled target domain. This way helps us obtain more reliable target pseudo labels, which enable contrastive domain alignment via feature prototypes even under unknown class distribution shifts.
Specifically, as CPGA, \ournet~also contains two stages:
1) Prototype generation: we keep the same contrastive source prototype generation strategy with CPGA to handle the lack of source data.
2) Prototype adaptation: instead of only relying on the source model, \ournet~generates target pseudo labels via the automatically weighted ensemble of self-supervised pseudo-labeling~\cite{liang2020shot} and CLIP zero-shot prediction. Meanwhile, rather than assigning confidence weights for target data based on source predictions as CPGA, we further reweight the target sample confidence to avoid noisy pseudo labels. To alleviate the negative effect of unidentified class distribution shifts, we further devise an additional target label-distribution-aware classifier to match the class distribution of the target domain. 
In this way, we are able to adapt a class distribution-agnostic source model to an unlabeled target domain even if both domains are class-imbalanced and agnostic. Extensive experiments on three imbalanced domain adaptation benchmark datasets demonstrate the effectiveness and superiority of \ournet~in handling imbalance-agnostic SF-UDA.

Our primary contributions are summarized as follows:
\begin{itemize}
 \item We introduce a novel CPGA method for addressing SF-UDA. Compared with previous SF-UDA methods, CPGA innovatively generates source feature prototypes to handle the absence of source data. More critically, these feature prototypes can also enhance the performance of conventional UDA methods, allowing them to achieve comparable or even superior results to those obtained through the illegitimate use of source data in SF-UDA. 
 
 \item We study a more practical task called imbalance-agnostic SF-UDA. Compared with vanilla SF-UDA, it assumes that the class distributions of both source and target domains are unknown and can be arbitrarily skewed. Hence, it accounts for unidentified class distribution shifts during adaptation, making it more applicable to real-world SF-UDA scenarios.
 
 \item We further propose a T-CPGA method to handle imbalance-agnostic SF-UDA. 
 This method introduces a new pseudo label generation strategy that is crucial for accurately generating pseudo labels for unlabeled target data, even under unknown class shifts. Specifically, this strategy identifies unknown target class distributions, and thus is essential for effective adaptation in imbalance-agnostic SF-UDA. 
 
\end{itemize}

A short version of this work was published in IJCAI 2021~\cite{Qiu2021CPGA}. This paper extends the previous version in the following aspects: 
1) It explores a novel task called imbalance-agnostic SF-UDA, which considers a more practical scenario where the class distributions of both the source and target domains can be imbalanced.
2) To solve unidentified class distribution shifts, \ournet~introduces a new pseudo label generation strategy and a target-aware classifier to better match the target class distribution.
3) The paper provides extensive new empirical evaluations, demonstrating that \ournet~achieves clearly better performance over CPGA (\eg{the average of 25.1\% and 22.5\% overall accuracy  gains on Cl$\rightarrow$Pr and Cl$\rightarrow$Rw of the imbalance-agnostic Office-Home dataset}).

\section{Related Work}
This section commences with a comprehensive literature review of relevant domain adaptation tasks, including Source-Free Unsupervised Domain Adaptation (SF-UDA) and Class-Imbalanced Domain Adaptation (CI-UDA). Then, we compare our task with the most pertinent Imbalanced SF-UDA~\cite{li2021imbalanced}. Due to page limitations, we provide the review of vanilla UDA in Appendix \mata{A}.

\subsection{Source-Free Unsupervised Domain Adaptation}
Unlike conventional UDA, SF-UDA methods~\cite{kim2020progressive,Li2020ModelAU} seek to adapt a source model to an unlabeled target domain without access to any source data. To handle this task, existing methods seek to refine the source model either by pseudo label generation (\ie SHOT~\cite{liang2020shot} and SHOT++~\cite{SHOT+}) or target-style images generation (\ie MA~\cite{Li2020ModelAU}). 
Nonetheless, pseudo labels would be noisy due to the domain discrepancy, which is ignored by SHOT. To address this issue, SHOT++ employs semi-supervised learning to improve the accuracy of less-confident predictions. As for MA, it may be plagued by the training difficulties of GAN-based approaches~\cite{karras2020training}.
Recent SF-UDA methods aim to alleviate the domain discrepancy via learning domain-invariant feature representations. For instance, NRC~\cite{yang2021exploiting} and G-SFDA~\cite{yang2021generalized} focus on leveraging neighborhood structures to encourage consistency in feature predictions. Alternatively, CAiDA~\cite{Dong2021caida} guides anchor points to search for semantically nearest confident anchors to generate pseudo labels and enhance feature representations.

Compared with the above methods, our CPGA proposes to generate source feature prototypes for each class to handle the lack of source data. Additionally, CPGA alleviates the negative effect of pseudo label noise via confidence reweighting and early learning regularization. 

\subsection{Imbalanced Unsupervised Domain Adaptation}
The objective of CI-UDA is to conduct domain alignment between a labeled source domain and an unlabeled target domain in the presence of class distribution shifts. 
Existing methods seek to overcome class distribution shifts by class-wise importance reweighting~\cite{self2022}, balanced sampling~\cite{tan2020class,li2021imbalanced} or representation learning~\cite{ShiZ022,TsaiHCYW16,tanwisuth2021prototype}.
Specifically, SIDA~\cite{self2022} employs self-adaptive imbalanced cross-entropy loss to adjust its model to varying degrees of imbalanced target scenarios. COAL~\cite{tan2020class} utilizes balanced sampling and self-training to address class distribution shifts and conducts prototype-based conditional alignment to mitigate domain shifts.
Regarding representation learning methods, CDM~\cite{TsaiHCYW16} 
exploits latent sub-domains within and across data domains to learn class-balanced feature representations for joint adaptation. Besides, PCT~\cite{tanwisuth2021prototype} aims to learn robust and domain-invariant representations by minimizing the expected pairwise cost between target features and imbalance-robust source prototypes. PAT~\cite{ShiZ022} reduces domain discrepancy by aligning centroids and generating adversarial samples for minority classes to handle the class imbalance issue.

In the context of CI-UDA, existing methods construct class-imbalanced UDA scenarios by sub-sampling datasets with imbalanced source domains and uniform or reversely-imbalanced target domains. Compared with imbalance-agnostic SF-UDA, they only account for a portion of imbalance scenarios. Moreover, CI-UDA relies on the accessibility of source data.

\subsection{Imbalanced Source-free Domain Adaptation}
ISFDA~\cite{li2021imbalanced} is a relevant study that investigates imbalanced source-free domain adaptation in which the class distributions between the source and target domains are opposite (e.g., long-tailed source and inversely long-tailed target). This study assumes that using class-balanced sampling to train the source model is permissible and introduces secondary label correction to handle class distribution shifts. However, the source-trained model is generally provided in advance and cannot be further trained for class re-balancing. In other words, the source model is more likely to be an imbalance-agnostic model trained via the standard cross-entropy loss. Moreover, ISFDA only focuses on the task with opposite class distributions. However, the source class distribution is not necessary to be inverse to the target class distribution in practice. Therefore, we relax the assumption and propose a more challenging but practical task, called imbalance-agnostic SF-UDA, where we seek to adapt an imbalance-agnostic source model to an imbalance-agnostic target domain with access to only unlabeled target data.

\begin{figure*}[t]
\centering
\includegraphics[width=15.8cm]{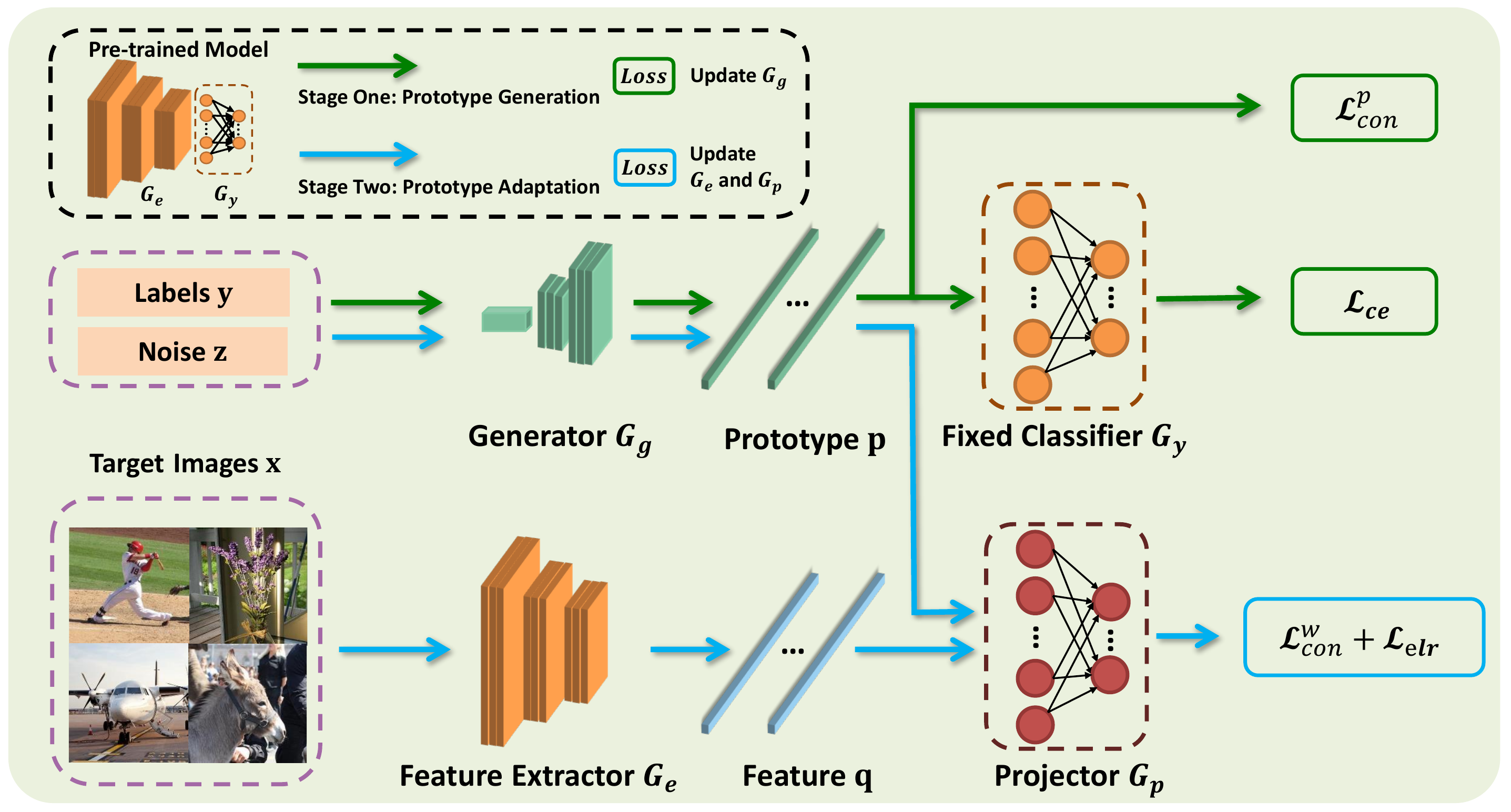}
\caption{An overview of CPGA. CPGA~contains two stages: (1) \textbf{Prototype generation}: under the guidance of the fixed classifier, a generator $G_{g}$ is trained to generate feature prototypes via $\mathcal{L}_{ce}$ and $\mathcal{L}_{con}^{p}$. (2) \textbf{Prototype adaptation}: in each training batch, we use the learned prototype generator to generate one prototype for each class. Based on the generated prototypes and pseudo labels obtained by clustering, we align each pseudo-labeled target feature to the corresponding class prototype by training a domain-invariant feature extractor via $\mathcal{L}_{con}^{w}$ and $\mathcal{L}_{elr}$. Note that the classifier $G_{y}$ is fixed during the whole training phase.}
\label{fig:overall}
\end{figure*}

\section{Problem Definition}
\noindent{\textbf{Source-Free Unsupervised Domain Adaptation (SF-UDA).}} We first study the task of SF-UDA, where only a well-trained source model and unlabeled target data are accessible.
Specifically, this work considers a multi-class classification task where the source and target domains share the same label space with $K$ classes. The pre-trained source model is assumed to consist of a feature extractor $G_{e}$ and a classifier $G_{y}$. Additionally, the unlabeled target domain is denoted by $\mathcal{D}_t=\{\textbf{x}_i\}_{i\small{=}1}^{n_t}$, where $n_{t}$ is the number of target samples. The primary objective is to adapt the source model to the target domain by leveraging only the unlabeled target data. The task of SF-UDA presents a challenge due to the lack of source data and target annotations. Conventional UDA methods that rely on source data are unable to tackle this task. To address the challenge of SF-UDA, we propose a  Contrastive Prototype Generation and Adaptation (CPGA) method (cf.~Section~\ref{method:cpga}). 

\newl

\noindent{\textbf{Imbalance-Agnostic SF-UDA.}}
Existing SF-UDA methods implicitly assume that the training class distributions of the source domain on which the source model is pre-trained and the target domain follow a balanced class distribution. However, in real-world applications, this assumption may not necessarily hold, and the source and target domains are likely to follow any class distribution (\eg being long-tailed, inversely long-tailed, or relatively class-balanced). For this reason, we study a more practical task, called imbalance-agnostic SF-UDA, where a class distribution-agnostic model trained via vanilla cross-entropy loss and a class distribution-agnostic unlabeled target domain are available. To resolve this task, we extend CPGA and propose Target-aware Contrastive Prototype Generation and Adaptation (cf.~Section~\ref{method:Tcpga}). For simplicity, we use the same notations as the above sections.




\section{CPGA: Contrastive Prototype Generation and Adaptation}
\label{method:cpga}

\subsection{Overall Scheme}
The key challenge of SF-UDA is the lack of source data.
Inspired by that feature prototypes can represent a group of semantically similar instances~\cite{snell2017prototypical}, we explore generating feature prototypes to represent each source class and adopt them for class-wise domain alignment. 
As shown in Figure~\ref{fig:overall}, CPGA consists of two stages: prototype generation and prototype adaptation.

In stage one (Section~\ref{stage1}), motivated by that the classifier of the source model contains class distribution information~\cite{Xu2020GenerativeLD}, we train a class conditional generator $G_{g}$ to learn such class information and generate feature prototypes for each class. Meanwhile, the source classifier $G_{y}$ is exploited to judge whether $G_{g}$ generates correct feature prototypes regarding classes. By training the generator $G_g$ to confuse $G_{y}$ via both cross-entropy $\mathcal{L}_{ce}$ and contrastive loss $\mathcal{L}_{con}^{p}$, we are able to generate intra-class compact and inter-class separated feature prototypes.
Meanwhile, to overcome the lack of target labels, we resort to a self pseudo-labeling strategy to generate pseudo labels for each target data. 

In stage two (Section~\ref{stage2}), we propose to adapt the source model to the target domain by aligning the pseudo-labeled target features to the corresponding source class prototypes. Specifically, we conduct class-wise alignment using a contrastive loss $\mathcal{L}_{con}^{w}$ with a domain projector $G_p$. Besides, we introduce an early learning regularization term $\mathcal{L}_{elr}$ to mitigate the effects of noisy pseudo labels on the adaptation process.

The overall procedure of CPGA is summarized as:
\begin{equation}
\label{loss:generator}
\min_{\theta_{g}} \mathcal{L}_{ce}(\theta_{g}) + \mathcal{L}_{con}^{p}(\theta_{g}),
\end{equation}
\vspace{-3.5mm}
\begin{equation}
\label{loss:extractor}
\min_{\{\theta_{e}, \theta_{p}\}} \mathcal{L}_{con}^{w}(\theta_{e}, \theta_{p}) + \lambda \mathcal{L}_{elr}(\theta_{e}, \theta_{p}),
\end{equation}
where $\theta_{g}$, $\theta_{e}$ and $\theta_{p}$ denotes the parameters of the generator $G_{g}$, the feature extractor $G_{e}$ and the projector $G_{p}$, respectively. Moreover, $\lambda$ is a trade-off parameter to balance losses. 
 
\begin{figure}[t]
\centering
\begin{minipage}{0.49\linewidth}
\subfigure[Training with $\mathcal{L}_{ce}$]{
\includegraphics[width=3.5cm,clip,trim={3.6cm 6.5cm 2.5cm 5.5cm}]{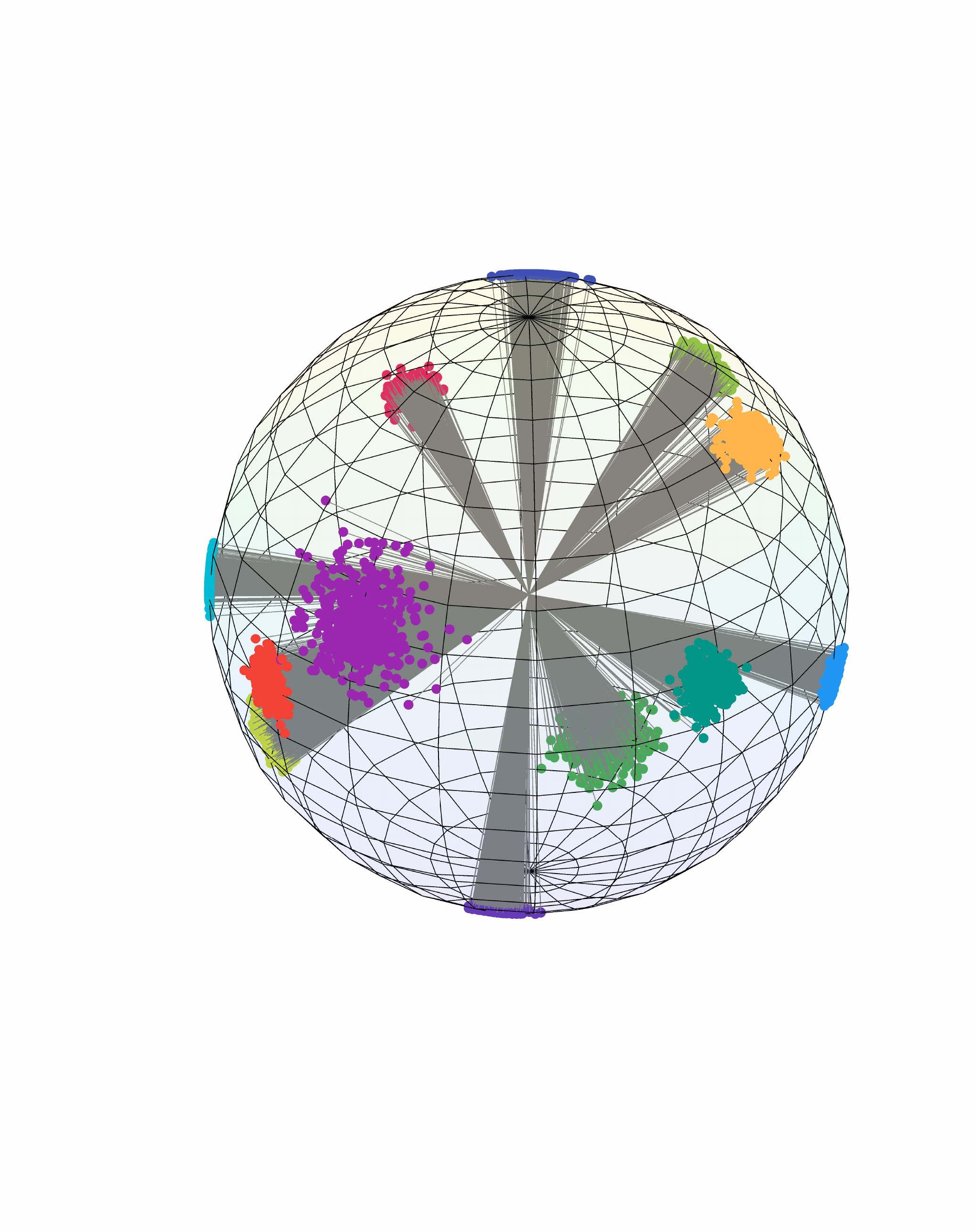}
\label{vis:lce}
}
\end{minipage}
\begin{minipage}{0.49\linewidth}
\subfigure[Training with $\mathcal{L}_{ce}\small{+}\mathcal{L}_{con}^{p}$]{
\includegraphics[width=3.5cm,clip,trim={3.6cm 6.5cm 2.5cm 5.5cm}]{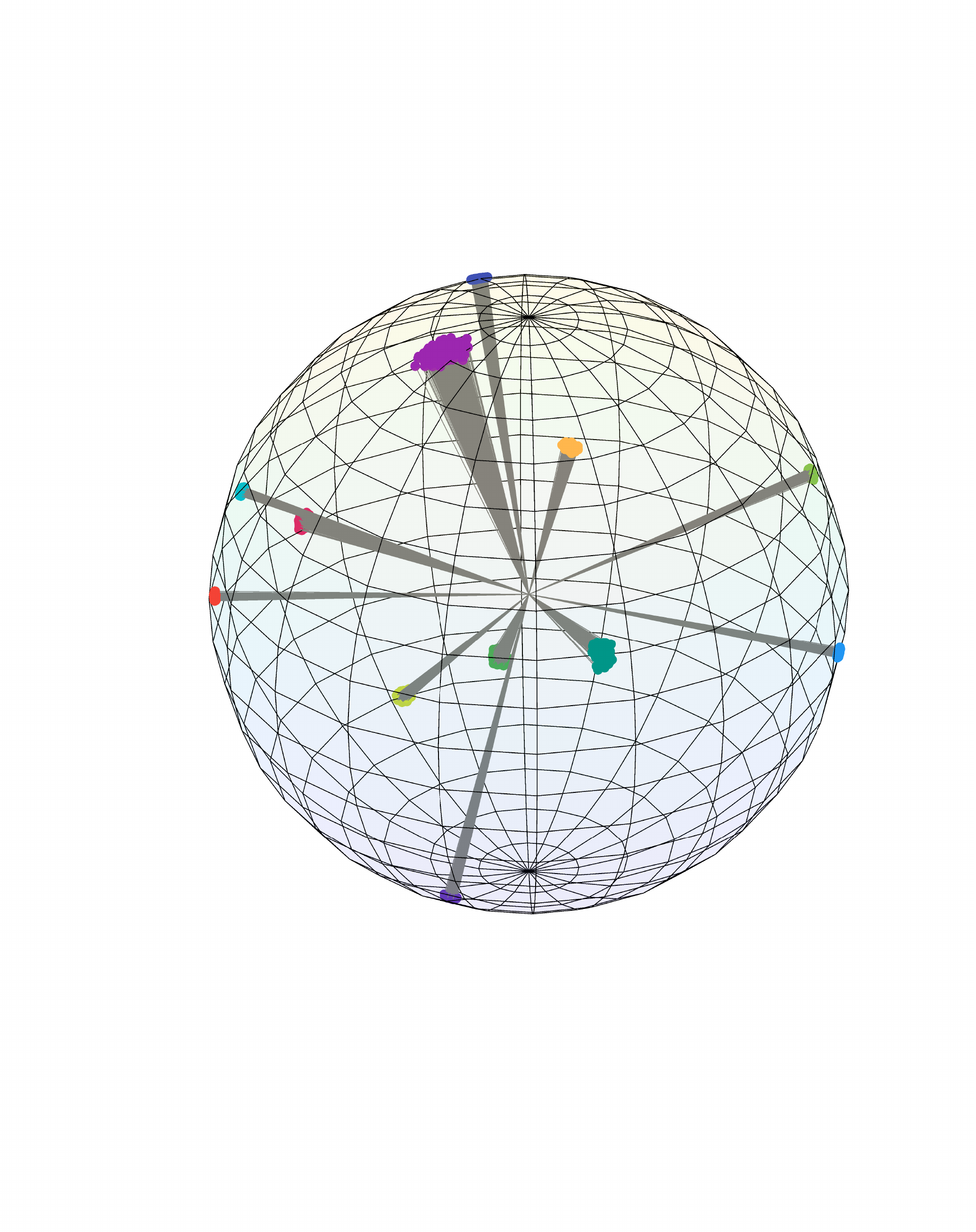}
\label{vis:lcec}
}
\end{minipage}
\label{figdata}
\caption{Visualizations of the generated feature prototypes by the generator trained with different losses, which shows the corresponding visual results of Table~\ref{tab:prototypes}.
Compared with training with only cross-entropy $\mathcal{L}_{ce}$, the contrastive loss $\mathcal{L}_{con}^{p}$
encourages the prototypes of the same category to be more compact and those of different categories to be more separated. Better viewed in color.}
\label{vis:lce_total}
\end{figure}

\subsection{Contrastive Prototype Generation}\label{stage1}
The absence of the source data makes UDA challenging. To handle this, we generate feature prototypes for each class by exploring the class distribution information hidden in the source classifier~\cite{Xu2020GenerativeLD}. To this end, we use the source classifier $G_{y}$ to train the class conditional generator $G_{g}$. 
To be specific, as shown in Figure~\ref{fig:overall}, given a uniform noise $\textbf{z}\sim U(0,1)$ and a label $\textbf{y}\in\mathbb{R}^K$ as inputs, the generator $G_{g}$ first generates the feature prototype $\textbf{p} = G_{g}(\textbf{y},\textbf{z})$ (more details of the generator and the generation process can be found in Appendix \mata{C}). Then, the classifier $G_y$ judges whether the generated prototype belongs to $\textbf{y}$ and trains the generator via the cross-entropy loss:
\begin{equation}
\label{eq:ce}
\mathcal{L}_{ce} = -\textbf{y}\log G_{y}(\textbf{p}),
\end{equation}
where $\textbf{p}$ is the generated prototype and $G_{y}(\textbf{p})$ denotes the prediction of the classifier.
In this way, the generator is capable of generating feature prototypes for each category. 

However, as shown in Figure~\ref{vis:lce}, training the generator with only cross-entropy may make the feature prototypes not well compact and prototypical. As a result, domain alignment with these prototypes may make the adapted model less discriminative, leading to limited performance (See Table~\ref{tab:prototypes}). To address this, motivated by InfoNCE~\cite{Oord2018RepresentationLW,Zhang2021UnleashingTP}, we further impose a contrastive loss to encourage more prototypical prototypes:
\begin{equation}
\label{eq:C_p}
\mathcal{L}_{con}^{p}\! \small{=} \small{-}\log \frac{\exp(\phi(\textbf{p}, \textbf{o}^+)/\tau)}{ \exp(\phi(\textbf{p}, \textbf{o}^+)/\tau) \small{+} \sum_{j=1}^{K\small{-}1}\exp(\phi(\textbf{p}, \textbf{o}_j^-)/\tau)},
\end{equation}
where $\textbf{p}$ denotes any anchor prototype. For each anchor, we sample the positive pair $\textbf{o}^+$ by randomly selecting a generated prototype with the same category to the anchor $\textbf{p}$, and sample $K\small{-}1$ negative pairs $\textbf{o}^-$ that have diverse classes with the anchor. Here, in each training batch, we generate at least 2 prototypes for each class in stage one. Moreover, $\phi(\cdot,\cdot)$ denotes the cosine similarity and $\tau$ is a temperature factor.

As shown in Figure~\ref{vis:lcec}, by training the generator with $\mathcal{L}_{ce}+\mathcal{L}_{con}^{p}$, the generated prototypes are more representative (\ie intra-class compact and inter-class separated). Interestingly, we empirically observe that the inter-class cosine distance will converge closely to 1 (\ie cosine similarity close to 0) by training with $\mathcal{L}_{ce} +\mathcal{L}_{con}^{p}$ (See Table~\ref{tab:prototypes}), if the feature dimensions are larger than the number of classes. That is, the generated prototypes of different categories are approximatively orthometric in the high-dimensional feature space.

\begin{algorithm}[t]
 \small
 \caption{Training of CPGA}\label{al:training}
 \KwIn{Unlabeled target data $\mathcal{D}_t\small{=}\{\textbf{x}_i\}_{i=1}^{n_t}$; Source model $\{G_{e}, G_{y}\}$; Training epoch $E$, $M$; Parameters $\beta$, $\tau$, $\lambda$.}
 Initialize Projector $G_{p}$, Generator $G_g$;
 
 \tcp{**~Stage 1: Prototype Generation~**~//}
 \For{$e = 1 \to E$}{
 Generate prototypes $\textbf{p}$ based on $G_{g}$;

 \tcp{Learn representative prototypes}
 Compute $\mathcal{L}_{ce}$ and $\mathcal{L}_{con}^{p}$ based on Eqns.~(\ref{eq:ce}) and (\ref{eq:C_p});

 Update generator $G_{g}$ based on Eqn.~(\ref{loss:generator});
 }
 Generate prototypes $\textbf{p}$ based on the learned $G_{g}$;

 \tcp{**~Stage 2: Prototype Adaptation~**~//}
 \For{$m = 1 \to M$}{
 Extract target data features $G_{e}(\textbf{x})$ based on $G_e$;
 
 Obtain target pseudo labels based on Eqn.~(\ref{pse_sca});
 
 Obtain contrastive features $\textbf{h}_{t}$ based on $G_{p}$;
 
 \tcp{Conduct class-wise domain alignment}
 Compute $\mathcal{L}_{con}^{w}$ based on Eqn.~(\ref{eq:C_p});

 \tcp{Prevent memorizing label noise}
 Compute $\mathcal{L}_{elr}$ based on Eqn.~(\mata{B.2}) (cf. Appendix \mata{B});
 
 Update target feature extractor $G_{e}$ based on Eqn.~(\ref{loss:extractor});
 }
 
 \KwOut{$G_{e}$ and $G_{y}$.}
\end{algorithm}

\subsection{Contrastive Prototype Adaptation}\label{stage2}
\noindent{\textbf{Pseudo label generation.}} Domain alignment can be conducted based on the generated source prototypes, However, the alignment is non-trivial due to the lack of target annotations, which makes the class-wise alignment difficult~\cite{pei2018multi,Kang2019ContrastiveAN}. To address this, a feasible way is to leverage a self-supervised pseudo-labeling strategy~\cite{liang2020shot} to generate pseudo labels for the target data. 
To be specific, let $\textbf{q}_i = G_{e}(\textbf{x}_i)$ denote the feature vector and let $\hat{y}_i^k = G_{y}^{k}(\textbf{q})$ be the predicted probability of the classifier regarding the class $k$. We first attain the initial centroid for each class $k$ by: $\textbf{c}_{k} = \frac{\sum_{i=1}^{n_{t}} \hat{y}_i^k \textbf{q}_i}{\sum_{i=1}^{n_{t}} \hat{y}_i^k}$, 
where $n_{t}$ is the number of target data. These centroids help to characterize the distribution of different categories~\cite{liang2020shot}. Then, the prediction of the $i$-th target data is obtained by: $\hat{\textbf{y}}_{i} = \sigma(\phi(\textbf{q}_i, \textbf{C})/\tau)$, 
where $\sigma(\cdot)$, $\phi(\cdot,\cdot)$ and $\textbf{C}\small{=}[\textbf{c}_0,...,\textbf{c}_{K-1}]$ denote the softmax function, cosine similarity and class centroid matrix, respectively. 
Moreover, the pseudo label is computed:
\begin{equation}\label{pse_sca}
\bar{y}_{i} = \mathop{\arg\max}_{k}{\hat{\textbf{y}}_{i}},
\end{equation}
where $\bar{y}_{i}\in \mathbb{R}^1$ is a scalar index.
During the training process, we update the centroid of each class by 
$\textbf{c}_{k} = \frac{\sum_{i=1}^{n_{t}} \mathbb{I}(\bar{y}_{i}\small{=}k) \textbf{q}_i}{\sum_{i=1}^{n_{t}}\mathbb{I}(\bar{y}_{i}\small{=}k)}$ and then update pseudo labels based on Eqn.~(\ref{pse_sca}) in each epoch,
where $\mathbb{I}(\cdot)$ is the indicator function. 

Based on the generated prototypes and target pseudo labels, we conduct prototype adaptation to alleviate domain shifts. Here, in each training batch, we generate one prototype for each class. 
However, due to domain shifts, the pseudo labels can be quite noisy, making the adaptation difficult. 
To address this, we propose a new contrastive prototype adaptation strategy, which consists of two key components: (1) weighted contrastive alignment and (2) early learning regularization. 

\newl

\noindent\textbf{Weighted contrastive alignment.} 
Relying on the pseudo-labeled target data, we then conduct class-wise contrastive learning to align the target data to the corresponding source feature prototype.
However, the pseudo labels may be noisy, making contrastive alignment degraded. To handle this issue, we differentiate pseudo-labeled target data and assign higher importance to reliable ones. Motivated by~\cite{Chen2019ProgressiveFA} that reliable samples are generally closer to the class centroid, we compute the confidence weight by:
\begin{equation}
w_{i} = \frac{\exp(\phi(\textbf{q}_i, \textbf{c}_{\bar{y}_{i}})/\tau)}{\sum_{k=1}^{K}\exp(\phi(\textbf{q}_i, \textbf{c}_{k} )/\tau)}, 
\end{equation} 
where the feature with higher similarity to the corresponding centroid will have higher importance.
Then, we can conduct weighted contrastive alignment. To this end, inspired by~\cite{chen2020simple}, we first use a non-linear projector $G_p$ to project the target features and source prototypes to a $l_2$-normalized contrastive feature space. Specifically, the target contrastive feature is denoted as $\textbf{u} = G_{p}(\textbf{q})$, while the prototype contrastive feature is denoted as $\textbf{v} = G_{p}(\textbf{p})$. Then, for any target feature $\textbf{u}_{i}$ as an anchor, we conduct prototype adaptation via a weighted contrastive loss:
\begin{equation}
\label{eqn_lwc}
\mathcal{L}_{con}^{w} \!=\! -\! w_{i}\!\log\frac{\exp(\textbf{u}_{i}^{\top} \textbf{v}^+/\tau )}{\exp(\textbf{u}_{i}^{\top} \textbf{v}^+/\tau) \small{+} \sum_{j=1}^{K\small{-}1}\exp(\textbf{u}_{i}^{\top} \textbf{v}^-_{j}/\tau)},
\end{equation}
where the positive pair $\textbf{v}^+$ is the prototype with the same class to the anchor $\textbf{u}_{i}$, while the negative pairs $\textbf{v}^-$ are the prototypes with different classes. 
\newl

\noindent\textbf{Early learning regularization.} 
As deep neural networks (DNNs) tend to first memorize the clean samples with correct labels and subsequently learn the noisy data with incorrect labels~\cite{arpit2017closer}, the model in the “early learning” phase can be more predictable to the noisy data. Therefore, inspired by~\cite{liu2020early}, we regularize the learning process via the early learning regularization term $\mathcal{L}_{elr}$ to further prevent the model from memorizing pseudo label noise. Please refer to Appendix \mata{B} for more details on $\mathcal{L}_{elr}$.

\begin{algorithm}[t]
 \small
 \caption{Training of T-CPGA}\label{al:training_tcpga}
 \KwIn {Unlabeled target data $\mathcal{D}_t\small{=}\{\textbf{x}_i\}_{i=1}^{n_t}$; Source model $\{G_{e}, G_{y}\}$; target label-distribution-aware classifier $G_{t}$; Training epoch $E$, $M$; Parameters $\beta$, $\tau$, $\lambda$.}
 Initialize Projector $G_{p}$; Generator $G_g$.
 
 \tcp{**~Stage 1: Prototype Generation~**~//}
 \For{$e = 1 \to E$}{
 Generate prototypes $\textbf{p}$ based on $G_{g}$; 

 
 Compute $\mathcal{L}_{ce}$ and $\mathcal{L}_{con}^{p}$ based on Eqns.~(\ref{eq:ce}) and (\ref{eq:C_p});

 Update generator $G_{g}$ based on Eqn.~(\ref{loss:generator});
 }
 
 Generate prototypes $\textbf{p}$ based on the learned $G_{g}$;

 \tcp{**~Stage 2: Prototype Adaptation~**~//}
 \For{$m = 1 \to M$}
 {Extract target data features $G_{e}(\textbf{x})$ based on $G_e$;

\scriptsize \tcp{Conduct target-aware pseudo label generation.}
\normalsize 
 
 Obtain pseudo labels based on Eqn.~(\ref{eqn:tcpga_pseudo});
 
 Obtain contrastive features $\textbf{h}_{t}$ based on $G_{p}$;

 Obtain confidence weights $w_{i}^{t}$ based on Eqn.~(\ref{weight_confi});
 
\scriptsize \tcp{Target-aware weighted contrastive alignment.}
\normalsize 
 
 Compute $\mathcal{L}_{con}^{wt}$ based on $w_{i}^{t}$ and Eqn.~(\ref{eqn_lwc_tcpga});
 
\scriptsize \tcp{Train the label-distribution-aware classifier $G_{t}$ to match the target distribution.}
\normalsize 

 Train target-aware classifier $G_{t}$ via $\mathcal{L}_{ce}^{t}$ (Eqn.~(\ref{loss:com}))
 
 Compute $\mathcal{L}_{elr}$ based on Eqn.~({\mata{B.2}}) (cf. Appendix \mata{B});

 Update $G_{e}$, $G_{y}$ and $G_{t}$ based on Eqn.~(\ref{loss:tcpga}).}

 \KwOut{$G_{e}$, $G_{y}$ and $G_{t}$.}
\end{algorithm}

\section{T-CPGA: Target-aware Contrastive Prototype Generation and Adaptation}
\label{method:Tcpga}

\subsection{Overall Scheme} 
In this section, we seek to adapt a class distribution-agnostic source model to a class distribution-agnostic target domain with access to only unlabeled target data. This task poses a new challenge in SF-UDA, as it involves adapting the source model to an unlabeled target domain under unidentified class distribution shifts. 
Existing SF-UDA methods (\eg SHOT~\cite{liang2020shot} and our CPGA) are unable to tackle this task, since they rely on the source model to generate pseudo labels for unlabeled target data, but the source model is class distribution-agnostic (\ie{source data are unknown and may be arbitrarily skewed}) and may generate noisy pseudo labels. Moreover, existing SF-UDA methods use a fixed source classifier, which may not provide accurate predictions for target data under unidentified class distribution shifts. 

To address these issues, by extending CPGA, we propose a Target-aware Contrastive Prototype Generation and Adaptation (\ournet)~method. We summarize the overall training scheme of \ournet~in Algorithms~\ref{al:training_tcpga}, which is made up of two stages. To handle the lack of source data, \ournet~holds the same first stage as CPGA. As for the second stage, it is unreliable for an imbalance-agnostic source model to generate accurate pseudo labels for unlabeled target data due to unidentified class distribution shifts. 
Inspired by the unknown class distribution identification ability of CLIP~\cite{radford2021learning} (cf.~Section~\ref{clip_ldd}), we leverage its zero-shot prediction capabilities to identify unknown target class distribution and adjust our pseudo-labeling strategy.
In addition, since the fixed classifier $G_y$ is biased toward the source label distribution which is probably different from the target ones, we develop an additional target label-distribution-aware classifier $G_{t}$  to adjust the bias. The overall training objective of the second stage is summarized as:
\begin{equation}
\label{loss:tcpga}
\min_{\{\theta_{e}, \theta_{p}, \theta_{t}\}} \mathcal{L}_{con}^{wt}(\theta_{e}, \theta_{p}) + \mathcal{L}_{ce}^{t}(\theta_{e}, \theta_{t}) + \lambda \mathcal{L}_{elr}(\theta_{e}, \theta_{p}),
\end{equation}
where $\theta_{e}$, $\theta_{p}$ and $\theta_{t}$ denotes the parameters of the feature extractor $G_{e}$, the projector $G_{p}$ and the target label-distribution-aware classifier $G_{t}$, respectively.  We will depict $\mathcal{L}_{con}^{wt}$ and $\mathcal{L}_{ce}^{t}$ in the following sub-sections.

\begin{figure}
 \centering
 \includegraphics[width=7.5cm]{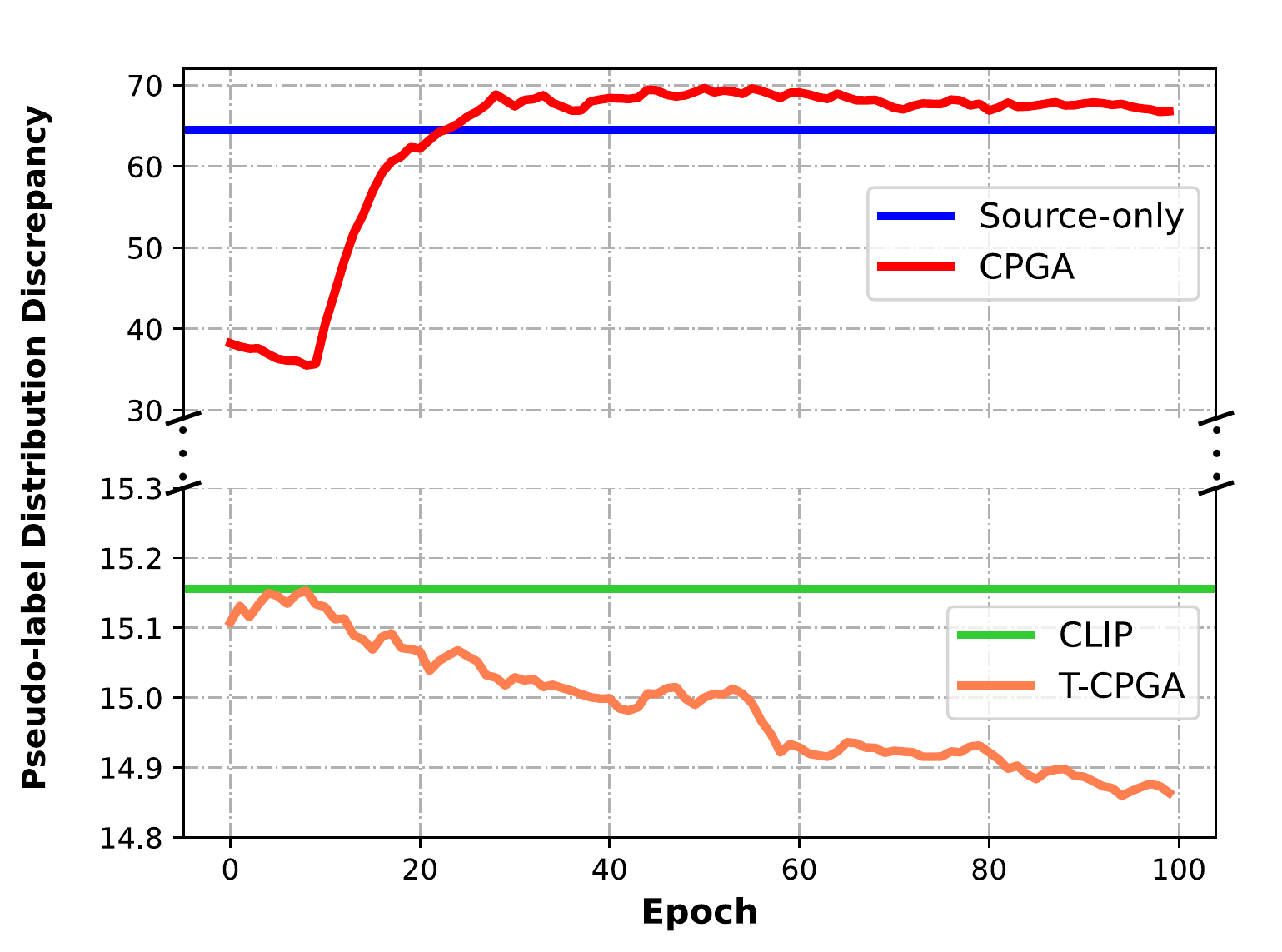}
 \caption{
 Pseudo-label distribution discrepancy for different methods on the \textbf{VisDA-I} dataset (long-tailed $\rightarrow$ inversely long-tailed, imbalance ratio 100). The pseudo-label distribution discrepancy means the difference in the amount of each category between ground truths and pseudo labels (or predictions) of compared methods. 
 The results show that T-CPGA can iteratively achieve more accurate pseudo labels with a better initialization via CLIP, while CPGA overfits noisy labels when it exists unidentified class distribution shifts.}
 \label{fig:clip_motivation}
\end{figure}

\subsection{Target-aware Contrastive Prototype Alignment}\label{sec5-2}

\noindent{\textbf{Target-aware pseudo label generation.}}
\label{clip_ldd}
As mentioned in Section~\ref{stage2}, CPGA generates pseudo labels for target samples based on Eqn.~(\ref{pse_sca}). Unfortunately, as the class distribution-agnostic source model may be biased toward majority classes in imbalanced scenarios, this strategy may fail to provide precise pseudo labels and leads to severe domain misalignment. 
To examine this issue, we introduce a metric called pseudo-label distribution discrepancy. It is calculated by comparing the per-category number of pseudo labels $\{{\textbf{y}_{pl}^i}\}_{i=1}^K$ to the ground truth labels $\{{\textbf{y}_{gt}^i}\}_{i=1}^K$, \ie{pseudo-label distribution discrepancy $d_{pdd}=\sum_{i=1}^K\frac{\left|\textbf{y}_{pl}^i-\textbf{y}_{gt}^i\right|}{\textbf{y}_{gt}^i}$.} A smaller pseudo-label distribution discrepancy value indicates that the generated pseudo labels are more reliable.

As shown in Figure~\ref{fig:clip_motivation}, a class-imbalanced source model trained on long-tailed source data  exhibits a significant pseudo-label discrepancy due to the data/class distribution shifts when applied to an inversely long-tailed target domain. This highlights the challenge of relying solely on the source model to generate pseudo labels, as it may lead to pseudo label noise and deviation from the ground truth. In contrast, CPGA exhibits a smaller pseudo-label distribution discrepancy, but eventually memorizes the noisy pseudo labels. As we mentioned before, DNNs would memorize clean samples at first, and then the noisy data with wrong labels~\cite{arpit2017closer}. Once the model memorizes the noisy data, it is prone to severe performance degradation. 

To handle this issue, we resort to CLIP~\cite{radford2021learning} (Contrastive Language-Image Pre-training), a powerful model for zero-shot prediction. In particular, CLIP's zero-shot prediction can provide relatively accurate predictions for unlabeled data even under unidentified class distribution shifts. As shown in Figure~\ref{fig:clip_motivation}, CLIP has a much smaller pseudo-label distribution discrepancy, which inspires us to leverage its zero-shot prediction abilities to identify unknown target class distributions. Despite the relatively reliable predictions,  solely using CLIP is sub-optimal since it does not take advantage of labeled source data for improvement (cf. Figure~\ref{fig:clip_motivation}). A feasible solution is to aggregate the knowledge of the source-trained model to constantly refine pseudo labels. 
Specifically, considering the unequal predictive power of the source-trained model and CLIP, we apply a dynamic ensemble strategy. Inspired by previous work~\cite{li2021imbalanced} that the predictions are more reliable when the discrepancy between the largest and second-largest predicted probabilities widens,   we propose to automatically assign ensemble weights based on the difference between their largest and the second-largest predicted probability. 
To be specific, let $\psi(\cdot)$ denote the CLIP model and $\sigma(\cdot)$ denote the softmax function.
We first compute the weights by: 
\begin{equation}
\begin{split} 
a_c &= \mathop{\max}_{k_1}\sigma(\psi(\textbf{x}_i)) - \mathop{\max}_{k_2,k_2\neq k_1} \sigma(\psi(\textbf{x}_i)), \\
a_p &= \mathop{\max}_{k_1}\hat{\textbf{y}}_i - \mathop{\max}_{k_2,k_2\neq k_1} \hat{\textbf{y}}_i,  
\end{split}
\end{equation}
where $a_c$ and $a_p$ are the weights for the CLIP and the predictions $\hat{\textbf{y}}_i$, respectively. Moreover, $k_1$ and $k_2$ are element indexes regarding different classes. To guarantee the sum of the ensemble prediction to be 1, we obtain the normalized weights $\bar{a}_c$ and $\bar{a}_p$ via a softmax function: $[\bar{a}_c, \bar{a}_p] = \sigma([a_c, a_p])$. 
Lastly, the final prediction of the $i$-th target data can be formulated:
\begin{equation}
\label{eqn:tcpga_pseudo}
\tilde{\textbf{y}}_{i} = \bar{a}_c\sigma(\psi(\textbf{x}_i)) + \bar{a}_p\hat{\textbf{y}}_i.
\end{equation} 
Afterward, we  can  obtain the pseudo label by: 
$\bar{y}_{i} = \mathop{\arg\max}_{k}{\tilde{\textbf{y}}_{i}}$, where $\bar{y}_{i}$  is a scalar index. 
As shown in Figure~\ref{fig:clip_motivation}, \ournet~is capable of producing relatively precise pseudo labels in the initial stage, while also improving the quality of the generated pseudo labels in the subsequent stage. 

\newl
\noindent{\textbf{Target-aware weighted contrastive alignment.}}
As mentioned in Section~\ref{stage2}, to mitigate the negative effect of pseudo label noise, we propose to differentiate target data based on their similarity to the corresponding centroid in CPGA. Nevertheless, due to unidentified class distribution shifts, such a strategy may be less reliable. 
To handle this, since target pseudo labels are obtained via ensemble intelligence, the confidence weights are modified as the maximum element of the prediction $\tilde{\textbf{y}}_{i}$:
\begin{equation}\label{weight_confi}
w_{i}^{t} = \max_{k}\tilde{\textbf{y}}_{i}, 
\end{equation} 
where $k$ is an element index.  
Eventually, the weighted contrastive loss of \ournet~is modified to:
\begin{equation}
\label{eqn_lwc_tcpga}
\mathcal{L}_{con}^{wt} \!=\! -\! w_{i}^{t}\!\log\frac{\exp(\textbf{u}_{i}^{\top} \textbf{v}^+/\tau )}{\exp(\textbf{u}_{i}^{\top} \textbf{v}^+/\tau) \small{+} \sum_{j=1}^{K\small{-}1}\exp(\textbf{u}_{i}^{\top} \textbf{v}^-_{j}/\tau)},
\end{equation}

\newl

\noindent{\textbf{Target label-distribution-aware classifier.}}
\label{classifier_ldd}
In CPGA, the final prediction is made by the fixed source classifier. Although contrastive alignment facilitates the alignment of target features to the source prototypes and thereby the source classifier, a fixed source classifier may not be capable of predicting target samples well in the presence of class distribution shifts across domains.
To address this issue, we develop an additional target label-distribution-aware classifier $G_{t}$ that is designed to particularly fit the target class distribution. Specifically,  we train $G_{t}$ using the cross-entropy loss to estimate the target pseudo label distribution:
\begin{equation}
\label{loss:com}
\mathcal{L}_{ce}^{t} = - \tilde{\textbf{y}}_{i}\log G_{t}(\textbf{q}_i),
\end{equation}
Compared with the fixed source classifier, the target-aware classifier $G_{t}$  matches the target class distribution better. Despite this, the existence of noisy pseudo labels may impede the classification performance of $G_{t}$.
To address this issue, we complementarily use $G_{y}$ and $G_{t}$ to get more accurate predictions via average ensemble, wherein $G_{y}$ demonstrates stronger classification ability thanks to sufficiently labeled source data, whereas $G_{t}$ conforms better to the target class distribution.

\begin{table}[t]
\setlength\tabcolsep{4pt}
 \begin{center}
 \caption{\label{tab:office}Overall Accuracy (\%) on the \textbf{Office-31} (ResNet-50).}
 \scalebox{0.8}{
 \begin{tabular}{lcccccccl}
 \toprule
 Method & Source-free & A$\rightarrow$D & A$\rightarrow$W & D$\rightarrow$W & W$\rightarrow$D & D$\rightarrow$A & W$\rightarrow$A & Avg.\\
 \midrule
 ResNet-50~\cite{He2016DeepRL} & \xmark & 68.9 & 68.4 & 96.7 & 99.3 & 62.5 & 60.7 & 76.1 \\
 MCD~\cite{saito2018maximum} & \xmark & 92.2 & 88.6 & 98.5 & 100.0 & 69.5 & 69.7 & 86.5 \\
 CDAN~\cite{long2018conditional} & \xmark & 92.9 & 94.1 & 98.6 & 100.0 & 71.0 & 69.3 & 87.7 \\
 MDD~\cite{zhang2019bridging} & \xmark & 90.4 & 90.4 & 98.7 & 99.9 & 75.0 & 73.7 & 88.0 \\
 CAN~\cite{Kang2019ContrastiveAN} & \xmark & 95.0 & 94.5 & 99.1 & 99.6 & 70.3 & 66.4 & 90.6 \\
 DMRL~\cite{wu2020dual} & \xmark & 93.4 & 90.8 & 99.0 & 100.0 & 73.0 & 71.2 & 87.9 \\
 BDG~\cite{yang2020bi} & \xmark & 93.6 & 93.6 & 99.0 & 100.0 & 73.2 & 72.0 & 88.5 \\
 MCC~\cite{jin2020minimum} & \xmark & 95.6 & 95.4 & 98.6 & 100.0 & 72.6 & 73.9 & 89.4 \\
 SRDC~\cite{tang2020unsupervised} & \xmark & 95.8 & 95.7 & 99.2 & 100.0 & 76.7 & 77.1 & 90.8 \\
 \midrule
 PrDA~\cite{kim2020progressive} & \cmark & 92.2 & 91.1 & 98.2 & 99.5 & 71.0 & 71.2 & 87.2 \\
 SHOT~\cite{liang2020shot} & \cmark & 93.1 & 90.9 & {98.8} & 99.9 & 74.5 & 74.8 & 88.7 \\
 BAIT~\cite{Yang2020UnsupervisedDA} & \cmark & 92.0 & {94.6} & 98.1 & {100.0} & 74.6 & 75.2 & 89.1 \\
 MA~\cite{Li2020ModelAU} & \cmark & 92.7 & 93.7 & 98.5 & 99.8 & 75.3 & {77.8} & 89.6 \\
 \midrule
 CPGA~(ours) & \cmark & {94.4} & 94.1 & 98.4 & 99.8 & {76.0} & 76.6 & \textbf{89.9} \\
 \bottomrule
 \end{tabular}
 }
 \end{center}
\end{table}

\begin{table*}[t]
\setlength\tabcolsep{8pt}
 \begin{center}
 \caption{\label{tab:visda} Per-class Accuracy (\%) on the large-scale \textbf{VisDA} dataset (ResNet-101).}
 \scalebox{0.75}{
 \begin{tabular}{lcccccccccccccc}
 \toprule
 Method & Source-free & plane & bicycle & bus & car & horse & knife & mcycl & person & plant & sktbrd & train & truck & Per-class\\
 \midrule
 ResNet-101~\cite{He2016DeepRL} & \xmark & 55.1 & 53.3 & 61.9 & 59.1 & 80.6 & 17.9 & 79.7 & 31.2 & 81.0 & 26.5 & 73.5 & 8.5 & 52.4 \\
 CDAN~\cite{long2018conditional} & \xmark & 85.2 & 66.9 & 83.0 & 50.8 & 84.2 & 74.9 & 88.1 & 74.5 & 83.4 & 76.0 & 81.9 & 38.0 & 73.9 \\
 SAFN~\cite{xu2019larger} & \xmark & 93.6 & 61.3 & 84.1 & 70.6 & 94.1 & 79.0 & 91.8 & 79.6 & 89.9 & 55.6 & 89.0 & 24.4 & 76.1 \\
 SWD~\cite{lee2019sliced} & \xmark & 90.8 & 82.5 & 81.7 & 70.5 & 91.7 & 69.5 & 86.3 & 77.5 & 87.4 & 63.6 & 85.6 & 29.2 & 76.4 \\
 TPN~\cite{pan2019transferrable} & \xmark & 93.7 & 85.1 & 69.2 & 81.6 & 93.5 & 61.9 & 89.3 & 81.4 & 93.5 & 81.6 & 84.5 & 49.9 & 80.4 \\
 PAL~\cite{hu2020panda} & \xmark & 90.9 & 50.5 & 72.3 & 82.7 & 88.3 & 88.3 & 90.3 & 79.8 & 89.7 & 79.2 & 88.1 & 39.4 & 78.3 \\ 
 MCC~\cite{jin2020minimum} & \xmark & 88.7 & 80.3 & 80.5 & 71.5 & 90.1 & 93.2 & 85.0 & 71.6 & 89.4 & 73.8 & 85.0 & 36.9 & 78.8 \\
 CoSCA~\cite{Dai_2020_ACCV} & \xmark & 95.7 & 87.4 & 85.7 & 73.5 & 95.3 & 72.8 & 91.5 & 84.8 & 94.6 & 87.9 & 87.9 & 36.8 & 82.9 \\
 \midrule
 PrDA~\cite{kim2020progressive} & \cmark & 86.9 & 81.7 & {84.6} & 63.9 & {93.1} & 91.4 & 86.6 & 71.9 & 84.5 & 58.2 & 74.5 & 42.7 & 76.7\\
 SHOT~\cite{liang2020shot} & \cmark & 92.6 & 81.1 & 80.1 & 58.5 & 89.7 & 86.1 & 81.5 & 77.8 & 89.5 & 84.9 & 84.3 & 49.3 & 79.6\\
 MA~\cite{Li2020ModelAU} & \cmark & 94.8 & 73.4 & 68.8 & {74.8} & {93.1} & 95.4 & 88.6 & {84.7} & 89.1 & 84.7 & 83.5 & 48.1 & 81.6\\
 BAIT~\cite{Yang2020UnsupervisedDA} & \cmark & 93.7 & 83.2 & 84.5 & 65.0 & 92.9 & 95.4 & 88.1 & 80.8 & 90.0 & 89.0 & 84.0 & 45.3 & 82.7 \\
 \midrule
 CPGA~(Ours) & \cmark & {95.6} & {89.0} & 75.4 & 64.9 & 91.7 & {97.5} & 89.7 & 83.8 & {93.9} & {93.4} & 87.7 & {69.0} & \textbf{86.0} \\
 \bottomrule
 \end{tabular}
 }
 \end{center}
 \vspace{-0.1in}
\end{table*}

\section{Experiment of vanilla sf-uda}
In this section, we empirically evaluate the effectiveness of CPGA for tackling vanilla SF-UDA. Moreover, we conduct ablation studies on the proposed two modules (\ie prototype generation and prototype adaptation).

\newl

\label{vanilla_data}
\noindent\textbf{Datasets.}\label{des:dataset}
We conduct  experiments on three benchmark datasets:
(1) \textbf{Office-31}~\cite{Saenko2010AdaptingVC}  is a standard domain adaptation dataset consisting of three distinct domains, \ie Amazon (A), Webcam (W) and DSLR (D). Three domains share 31 categories and contain 2817, 795 and 498 samples, respectively.
(2) \textbf{VisDA}~\cite{Peng2017VisDATV} is a large-scale dataset that concentrates on the 12-class synthesis-to-real object recognition task. The dataset has a source domain containing 152k synthetic images and a target domain with 55k real object images. 
(3) \textbf{Office-Home}~\cite{Venkateswara2017DeepHN} is a medium-sized dataset consisting of four distinct domains, \ie Artistic images (Ar), CLIP Art (Cl), Product images (Pr) and Real-world images (Rw).  The dataset contains 65 categories in each of the four domains.

\newl

\vspace{0.05in}
\begin{table}[t]
\setlength\tabcolsep{1pt}
 \begin{center}
 \caption{\label{tab:proto-office}Comparisons of the existing domain adaptation methods with source data or prototypes on \textbf{Office-31} (ResNet-50).}
 \scalebox{0.85}{
 \begin{tabular}{lccccccc}
 \toprule
 Method & A$\rightarrow$D & A$\rightarrow$W & D$\rightarrow$W & W$\rightarrow$D & D$\rightarrow$A & W$\rightarrow$A & Avg.\\
 \midrule
 DANN (with source data) & 79.7 & 82.0 & 96.9 & 99.1 & 68.2 & 67.4 & \textbf{82.2} \\
 DANN (with prototypes) & {83.7} & 81.1 & {97.5} & {99.8} & 63.4 & 63.6 & 81.5 \\
 \midrule
 DMAN (with source data) & 83.3 & 85.7 & 97.1 & 100.0 & 65.1 & 64.4 & 82.6 \\
 DMAN (with prototypes) & {86.3} & 84.2 & {97.7} & {100.0} & 64.7 & {64.5} & \textbf{82.9} \\
 \midrule
 ADDA (with source data) & 82.9 & 79.9 & 97.4 & 99.4 & 64.9 & 63.6 & 81.4 \\
 ADDA (with prototypes) & {83.5} & {81.9} & 97.2 & {100.0} & 63.8 & 63.0 & \textbf{81.6} \\
 \bottomrule
 \end{tabular}
 }
 \end{center}
\end{table}

\noindent\textbf{Baselines.}
We compare CPGA~with three types of baselines:
(1) source-only model: ResNet~\cite{He2016DeepRL};
(2) UDA methods:
 MCD~\cite{saito2018maximum}, CDAN~\cite{long2018conditional}, TPN~\cite{pan2019transferrable}, SAFN~\cite{xu2019larger}, SWD~\cite{lee2019sliced}, MDD~\cite{zhang2019bridging}, CAN~\cite{Kang2019ContrastiveAN}, DMRL~\cite{wu2020dual}, BDG~\cite{yang2020bi}, PAL~\cite{hu2020panda}, MCC~\cite{jin2020minimum}, SRDC~\cite{tang2020unsupervised}; (3) SF-UDA methods: SHOT~\cite{liang2020shot}, PrDA~\cite{kim2020progressive}, MA~\cite{Li2020ModelAU} and BAIT~\cite{Yang2020UnsupervisedDA}.

\newl


\noindent\textbf{Implementation details.}
We implement our method in PyTorch. We use a ResNet~\cite{He2016DeepRL} model pre-trained on ImageNet as the backbone for all methods. Following~\cite{liang2020shot}, we replace the original fully connected (FC) layer with a task-specific FC layer followed by a weight normalization layer. 
The projector consists of three FC layers with hidden feature dimensions of 1024, 512 and 256. 
We train the source model via label smoothing technique~\cite{Mller2019WhenDL} and train CPGA using SGD optimizer. To get more compact feature representations, we further train the extractor via the neighborhood clustering term~\cite{Saito2020UniversalDA}. More implementation details are put in Appendix \mata{C} due to the page limitation.

\begin{table}[t]
\setlength\tabcolsep{17pt}
 \begin{center}
 \caption{\label{tab:ablation}Ablation study of the losses (\ie $\mathcal{L}_{con}^w$ and $\mathcal{L}_{elr}$) in terms of  per-class accuracy (\%) on \textbf{VisDA}. Here, $\mathcal{L}_{con}$ indicates $\mathcal{L}_{con}^{w}$ without the confidence weight $w$.}
 \scalebox{0.7}{
 \begin{tabular}{cccc|c}
 \toprule
 Backbone & $\mathcal{L}_{con}$& $\mathcal{L}_{con}^{w}$ & $\mathcal{L}_{elr}$ & Per-class (\%) \\
 \midrule
 \cmark & & & & 52.4 \\
 \cmark & \cmark & & & 80.9 \\
 \cmark & & \cmark & & 83.6\\
 \cmark & & \cmark & \cmark & \textbf{86.0} \\
 \bottomrule
 \end{tabular}
 }
 \end{center}
 \vspace{0.03in}
\end{table}

\begin{table}[t]
\renewcommand\arraystretch{1.4}
\setlength\tabcolsep{4pt}
 \begin{center}
 \caption{\label{tab:prototypes}Ablation studies on prototype generation in stage one with different losses. Inter-class distance and intra-class distance are based on cosine distance (range from 0 to 2). We report per-class accuracy (\%) after training the model on \textbf{VisDA}.}
 \scalebox{0.85}{\begin{tabular}{cccc}
 \toprule
 Objective & Inter-class distance & Intra-class distance & Per-class (\%)\\
 \midrule
 $\mathcal{L}_{ce}$ & 0.7860 & $3.343\times e^{-4}$ & 85.0\\
 $\mathcal{L}_{ce} + \mathcal{L}_{con}^{p}$ & 1.0034 & $2.670\times e^{-6}$ & 86.0\\
 \bottomrule
 \end{tabular}
 }
 \end{center}
 \vspace{0.03in}
\end{table}

\subsection{Results of Vanilla SF-UDA}
As shown in Table~\ref{tab:office},  the proposed CPGA~achieves the best performance on \textbf{Office-31}, compared with SF-UDA methods \wrt the average accuracy over 6 transfer tasks. 
Note that even when compared with the state-of-the-art methods using source data (\eg SRDC), our CPGA~is still able to obtain a competitive result.
Besides, Table~\ref{tab:visda} demonstrates that CPGA outperforms all the state-of-the-art methods \wrt the average accuracy (\ie per-class accuracy) on the challenging \textbf{VisDA} dataset. Specifically, CPGA achieves the highest accuracy regarding eight classes of the VisDA dataset, while also obtaining comparable results in the remaining classes.
Moreover, our CPGA also surpasses the baseline methods with source data (\eg CoSCA), which demonstrates the superiority of our proposed method.
Due to the page limitation, we put the results on \textbf{Office-Home} in   Appendix \mata{D}.

\subsection{Ablation Studies of Vanilla SF-UDA}
To evaluate the effectiveness of the proposed two modules (\ie prototype generation and prototype adaptation), we conduct a series of ablation studies on VisDA. Moreover, we put the analysis of hyper-parameters in  Appendix \mata{D}.

\setlength\tabcolsep{4pt}
\begin{table*}[t]
 \begin{minipage}[t]{0.47\textwidth}
 \begin{center}
 \caption{\label{tab:oh_cl2pr_oa} \textbf{Overall} Accuracy (\%) of \textbf{Cl$\rightarrow$Pr} Task with {different class distribution shifts} on the \textbf{Office-Home-I} dataset (ResNet-50). SF and CI indicate source-free and class-imbalanced.}
 \scalebox{0.6}{ 
 \begin{tabular}{lcccccccccccccccc}
 \toprule
 Method & 
 SF & CI
 & {FLT$\rightarrow$FLT} &
 {FLT$\rightarrow$BLT} &
 {FLT$\rightarrow$Bal} &
 {BLT$\rightarrow$FLT} & 
 {BLT$\rightarrow$BLT} & 
 {BLT$\rightarrow$Bal} & 
 {Avg.} \\
 \midrule
 ResNet-50~\cite{He2016DeepRL} & \xmark & \xmark &53.88	&43.93	&48.19	&44.51	&54.26	&51.39	&49.36
 \\
 DANN~\cite{ganin2015unsupervised} & \xmark & \xmark &65.90	&45.10	&51.50	&43.40	&66.90	&59.00	&55.30
 \\
  MDD~\cite{zhang2019bridging} & \xmark & \xmark &69.41	&48.92	&55.24	&46.32	&68.21	&61.86	&58.33
 \\
 MCC~\cite{jin2020minimum} & \xmark & \xmark &53.28	&42.92	&47.02	&39.11	&54.41	&48.19	&47.49
 \\
 ToAlign~\cite{wei2021toalign} & \xmark & \xmark &69.66	&56.22	&65.40	&52.42	&71.54	&64.50	&63.29
 \\
 \midrule
 COAL~\cite{tan2020class} & \xmark & \cmark &64.06	&58.74	&63.37	&57.11	&61.81	&64.05	&61.52
 \\
 PCT~\cite{tanwisuth2021prototype} & \xmark & \cmark &67.94	&59.29	&66.97	&55.34	&70.73	&67.24	&64.58
 \\
 \midrule
 SHOT~\cite{liang2020shot} & \cmark & \xmark &69.66	&58.74	&66.50	&56.35	&70.43	&72.18	&65.64 
 \\
 BAIT~\cite{Yang2020UnsupervisedDA} & \cmark & \xmark &65.98	&53.20	&61.84	&54.18	&64.84	&61.84	&60.31
 \\
 NRC~\cite{yang2021exploiting} & \cmark & \xmark &71.77	&64.58	&72.85	&59.43	&69.57	&72.70	&68.48
 \\
 CPGA~(Ours) & \cmark & \xmark &65.73	&56.17	&60.37	&53.78	&66.00	&64.16	&61.03
 \\
 \midrule
 ISFDA~\cite{li2021imbalanced} & \cmark & \cmark &67.59	&66.35	&73.15	&56.75	&68.06	&71.05	&67.16
 \\ 
 T-CPGA (Ours) & \cmark & \cmark &\textbf{84.88} & \textbf{86.25} & \textbf{87.38} & \textbf{84.78} & \textbf{86.20} & \textbf{87.20} & \textbf{86.12} 
 \\


 \bottomrule
 \end{tabular}}
 
 \end{center}
 \end{minipage}
 \hfill
 \begin{minipage}[t]{0.47\textwidth}
 \begin{center}
 \caption{\label{tab:oh_cl2rw_oa} \textbf{Overall} Accuracy (\%) of \textbf{Cl$\rightarrow$Rw} Task with different class distribution shifts on the \textbf{Office-Home-I} dataset (ResNet-50). SF and CI indicate source-free and class-imbalanced.}
 \scalebox{0.6}{ 
 \begin{tabular}{lcccccccccccccccc}
 \toprule
 Method & 
 SF & CI
 & {FLT$\rightarrow$FLT} &
 {FLT$\rightarrow$BLT} &
 {FLT$\rightarrow$Bal} &
 {BLT$\rightarrow$FLT} & 
 {BLT$\rightarrow$BLT} & 
 {BLT$\rightarrow$Bal} & 
 {Avg.} \\
 \midrule
 ResNet-50~\cite{He2016DeepRL} & \xmark & \xmark &54.43	&44.93	&50.01	&45.41	&58.66	&54.05	&51.25

 \\
 DANN~\cite{ganin2015unsupervised} & \xmark & \xmark &66.80	&44.70	&57.20	&46.00	&71.70	&65.20	&58.60

 \\
 MDD~\cite{zhang2019bridging} & \xmark & \xmark &69.67	&50.36	&58.30	&48.28	&71.43	&69.82	&61.31

 \\
 MCC~\cite{jin2020minimum} & \xmark & \xmark &54.99	&43.50	&51.46	&40.78	&63.29	&51.18	&50.87

 \\
 ToAlign~\cite{wei2021toalign} & \xmark & \xmark &71.35	&55.95	&69.02	&53.71	&72.55	&70.37	&65.49

 \\
 \midrule
 COAL~\cite{tan2020class} & \xmark & \cmark &61.94	&58.82	&68.21	&58.98	&68.32	&68.37	&64.11

 \\
 PCT~\cite{tanwisuth2021prototype} & \xmark & \cmark &70.23	&59.86	&70.48	&56.42	&71.03	&69.06	&66.18

 \\
 \midrule
 SHOT~\cite{liang2020shot} & \cmark & \xmark &68.95	&61.85	&74.27	&60.02	&72.39	&72.96	&68.41

 \\
 BAIT~\cite{Yang2020UnsupervisedDA} & \cmark & \xmark &65.28	&51.56	&61.72	&51.16	&70.79	&63.00	&60.59

 \\
 NRC~\cite{yang2021exploiting} & \cmark & \xmark &65.44	&63.77	&72.14	&61.85	&70.71	&75.24	&68.19

 \\
 CPGA~(Ours) & \cmark & \xmark &62.61	&59.46	&66.67	&54.91	&70.31	&66.86	&63.47

 \\

 \midrule
 ISFDA~\cite{li2021imbalanced} & \cmark & \cmark &68.40	&67.60	&71.06	&61.77	&70.79	&71.77	&68.56

 \\ 
 T-CPGA (Ours) & \cmark & \cmark & \textbf{85.16} & \textbf{85.79} & \textbf{87.15} & \textbf{85.00} & \textbf{85.87} & \textbf{87.03} & \textbf{86.00}

 \\


 \bottomrule
 \end{tabular}}
 \end{center}
 \end{minipage}
\end{table*}

\begin{table*}[t]
\setlength\tabcolsep{12.5pt}
 \begin{center}
 \caption{\label{tab:overall-domainnet} \textbf{Overall} Accuracy (\%) on the \textbf{DomainNet-S} dataset (ResNet-50). SF and CI indicate source-free and class-imbalanced.}
 \scalebox{0.7}{ 
 \begin{tabular}{lcccccccccccccccc}
 \toprule
 Method & 
 SF & CI
 & {C$\rightarrow$P} &
 {C$\rightarrow$R} &
 {C$\rightarrow$S} &
 {P$\rightarrow$C} & 
 {P$\rightarrow$R} & 
 {P$\rightarrow$S} & 
 {R$\rightarrow$C} &
 {R$\rightarrow$P} &
 {R$\rightarrow$S} &
 {S$\rightarrow$C} & 
 {S$\rightarrow$P} & 
 {S$\rightarrow$R} &
 {Avg.} \\
 \midrule
 ResNet-50~\cite{He2016DeepRL} & \xmark & \xmark &56.96	&76.58	&58.02	&57.92	&82.49	&65.74	&66.46	&74.56	&60.19	&60.15	&62.70	&74.43	&66.35

 \\
 DANN~\cite{ganin2015unsupervised} & \xmark & \xmark &61.70	&81.70	&65.50	&59.30	&77.30	&61.00	&74.10	&77.30	&71.60	&73.10	&69.00	&79.60	&70.93

 \\
  MDD~\cite{zhang2019bridging} & \xmark & \xmark &70.30	&86.72	&72.70	&62.31	&85.80	&69.20	&79.58	&79.24	&73.24	&77.41	&74.87	&84.29	&76.31

 \\
 MCC~\cite{jin2020minimum} & \xmark & \xmark &51.94	&81.06	&60.23	&63.12	&84.19	&57.40	&66.52	&61.16	&55.77	&62.62	&55.55	&74.85	&64.53

 \\
 ToAlign~\cite{wei2021toalign} & \xmark & \xmark &70.20	&86.98	&71.86	&67.20	&84.86	&73.74	&78.71	&80.10	&73.70	&77.29	&74.22	&83.93	&76.90

 \\
 \midrule

 COAL~\cite{tan2020class} & \xmark & \cmark &73.50	&84.65	&71.03	&69.99	&87.20	&67.15	&75.99	&79.37	&61.61	&77.23	&75.35	&85.22	&75.69

 \\
 PCT~\cite{tanwisuth2021prototype} & \xmark & \cmark &73.24	&89.21	&75.24	&75.07	&88.47	&75.51	&78.58	&81.17	&74.82	&79.74	&78.58	&86.77	&79.70

 \\
 \midrule
 SHOT~\cite{liang2020shot} & \cmark & \xmark &76.90	&89.07	&72.57	&74.63	&88.92	&74.28	&76.67	&77.62	&71.24	&74.81	&75.39	&86.92	&78.25

 \\
 BAIT~\cite{Yang2020UnsupervisedDA} & \cmark & \xmark &81.95	&90.48	&76.74	&76.30	&87.28	&76.28	&77.97	&82.16	&74.20	&81.68	&79.20	&88.05	&81.02

 \\
 NRC~\cite{yang2021exploiting} & \cmark & \xmark &77.93	&90.47	&76.07	&78.22	&90.31	&75.74	&80.07	&78.62	&74.49	&80.82	&80.82	&91.06	&81.22

 \\
 CPGA~(Ours) & \cmark & \xmark &68.06	&84.91	&66.57	&69.06	&84.72	&69.53	&74.32	&79.34	&63.78	&75.31	&74.32	&84.13	&74.50

 \\
 \midrule
 ISFDA~\cite{li2021imbalanced} & \cmark & \cmark &77.38	&89.30	&73.78	&77.91	&89.73	&72.61	&80.07	&80.44	&72.07	&77.60	&76.76	&87.31	&79.58

 \\ 
 T-CPGA (Ours) & \cmark & \cmark & \textbf{86.59}	& \textbf{93.30}	& \textbf{85.08}	& \textbf{89.36}	& \textbf{92.99}	&\textbf{85.54}	&\textbf{90.10}	& \textbf{86.59}	&\textbf{85.49}	&\textbf{89.73}	&\textbf{86.73}	&\textbf{93.03}	&\textbf{88.71}

 \\


 \bottomrule
 \end{tabular}}
 \end{center}
\end{table*}
\newl

\noindent\textbf{Effectiveness of prototype generation.}
In this section, we verify the benefits of our generated prototypes to existing UDA methods (\eg DANN~\cite{ganin2015unsupervised}, ADDA~\cite{Tzeng2017AdversarialDD} and DMAN~\cite{zhang2019whole}), which cannot resolve SF-UDA previously.
Specifically, we use the generated prototypes to replace their source data for domain alignment. As shown in Table~\ref{tab:proto-office},
these methods based on prototypes achieve competitive performance compared with the counterparts using source data, or even perform better in some tasks of Office-31. This results demonstrates the benefits and applicability of our prototype generation scheme to existing UDA methods.

\newl

\noindent\textbf{Ablation studies on prototype generation.}
To study the impact of our contrastive loss $\mathcal{L}_{con}^p$, we compare the results of models with and without $\mathcal{L}_{con}^p$. 
As shown in  Table~\ref{tab:prototypes}, compared with training by only the cross-entropy loss $\mathcal{L}_{ce}$, optimizing the generator via $\mathcal{L}_{ce} \small{+} \mathcal{L}_{con}^{p}$ makes the inter-class features separated (\ie larger inter-class distance) and intra-class features compact (\ie smaller intra-class distance).
As a result,  $\mathcal{L}_{con}^p$ enhances the final adaptation performance by 1\% accuracy gains. 
\newl

\noindent\textbf{Ablation studies on prototype adaptation.}
We next ablate the losses in prototype adaptation.
As shown in Table~\ref{tab:ablation}, compared with the conventional contrastive loss $\mathcal{L}_{con}$, our weighted contrastive loss $\mathcal{L}_{con}^w$ can achieve more promising performance on VisDA. This result verifies the ability of our method to alleviate pseudo label noise.
Besides, $\mathcal{L}_{elr}$ can also improve the performance, since it prevents the model from memorizing pseudo label noise.
When combining all the losses (\ie $\mathcal{L}_{con}^w$ and $\mathcal{L}_{elr}$), our method obtains the best performance.

\newl

\section{Experiment~of~imbalance-agnostic~sf-uda}
This section evaluates \ournet~for handling imbalance-agnostic SF-UDA. Subsequently, we discuss the use of CLIP and the target label-distribution-aware classifier.  

\newl

\noindent\textbf{Datasets.} To simulate target class-distribution-agnostic scenarios, inspired by~\cite{li2021imbalanced}, we construct the following datasets. 1) \textbf{VisDA-I} is a variant of the VisDA~\cite{Peng2017VisDATV}, which is 12-class synthesis-to-real object recognition task. The source domain has two inverse distributions, \ie forward long-tailed distribution (FLT) and backward long-tailed distribution (BLT), while the target domain has three, \ie FLT, BLT and a relative balance distribution (Bal). 
Note that we term the class distribution of the original target domain in the VisDA as Bal. Hence, such a dataset has 6 tasks with different class distribution shifts. Moreover, we use an imbalance factor to measure the degree of imbalance, \ie $\mu \small{=} \frac{N_{max}}{N_{min}}$, where $N_{max}$ and $N_{min}$ denote the number of samples in the maximum class and minimum class, respectively. 2) \textbf{Office-Home-I} is a variant of the Office-Home~\cite{Venkateswara2017DeepHN}, which contains three distinct domains, \ie Clipart (Cl), Product images (Pr) and Real-World images (Rw). Each domain has three class distributions (\ie FLT, BLT and Bal), where Bal denotes the vanilla class distribution in the Office-Home. 3) \textbf{DomainNet-S} constructed by Tan \etal\cite{tan2020class} consists of four domains (Real (R), Clipart (C), Painting (P), Sketch (S)) with 40 classes. Since each domain of DomainNet-S  is imbalanced,  we directly use it for imbalance-agnostic SF-UDA.

\begin{figure*}[t]
\centering
\includegraphics[width=0.95\linewidth]{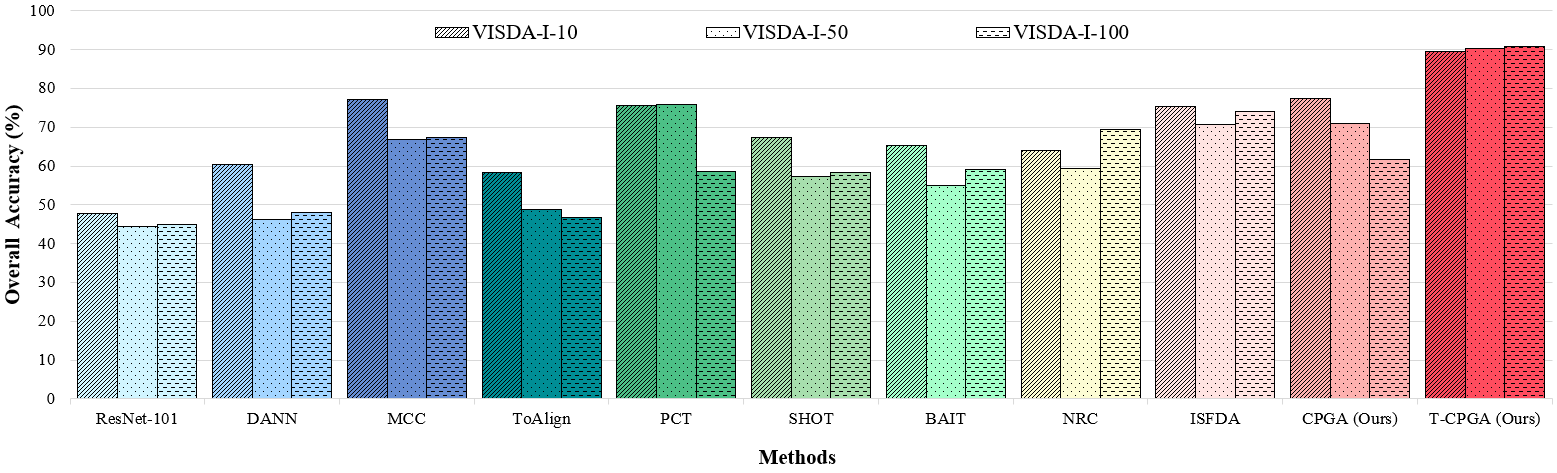}
\caption{\textbf{Overall} Accuracy (\%) on the \textbf{VisDA-I} dataset (ResNet-101). The number after VisDA-I is the imbalance ratio.}
\label{fig:visda_histograms_oa}
\end{figure*}

\begin{table*}[t]
\setlength\tabcolsep{9pt}
 \begin{center}
 \caption{\label{tab:abolation-domainnet} Ablation studies of source bias compensation and target pseudo label generation for \ournet~on the \textbf{DomainNet-S} dataset (ResNet-50) in terms of overall accuracy (\%). We first show \ournet~w/o target label-distribution-aware classifier $G_{t}$ (\ie $\mathcal{L}_{ce}$).
 Meanwhile, to validate the effectiveness of our pseudo label generation strategy, we show \ournet~with pseudo label generation only by CLIP~\cite{radford2021learning}.}
 \scalebox{0.8}{ 
 \begin{tabular}{lcccccccccccccccc}
 \toprule
 Method&
 {C$\rightarrow$P} &
 {C$\rightarrow$R} &
 {C$\rightarrow$S} &
 {P$\rightarrow$C} & 
 {P$\rightarrow$R} & 
 {P$\rightarrow$S} & 
 {R$\rightarrow$C} &
 {R$\rightarrow$P} &
 {R$\rightarrow$S} &
 {S$\rightarrow$C} & 
 {S$\rightarrow$P} & 
 {S$\rightarrow$R} &
 {Avg.} \\
 \midrule
 \ournet~(w/o target-aware classifier)
 &83.84 &91.66 &84.04 &87.13 &92.08 &83.58 &89.42 &86.15 &83.95 &85.58 &83.26 &90.22 &86.74
 \\
 \ournet~(only pseudo-labeling by CLIP)
 &85.22 &92.99 &82.12 &87.19 &92.84 &82.66 &87.31 &86.15 &82.49 &86.70 &85.42 &\textbf{93.04} &87.01
 \\
 \ournet &\textbf{86.59}	&\textbf{93.30}	&\textbf{85.08}	& \textbf{89.36}	&\textbf{92.99}	&\textbf{85.54}	&\textbf{90.10}	&\textbf{86.59}	&\textbf{85.49}	&\textbf{89.73}	&\textbf{86.73}	&{93.03}	&\textbf{88.71}

 \\
 \bottomrule
 \end{tabular}}
 \end{center}
\end{table*}

\newl

\noindent\textbf{Baselines.}
We compare \ournet~with five categories of baselines: 1) source-only model: ResNet~\cite{He2016DeepRL}; 2) UDA methods: DANN~\cite{ganin2015unsupervised}, MDD~\cite{zhang2019bridging},  MCC~\cite{jin2020minimum}, ToAlign~\cite{wei2021toalign}; 3) CI-UDA methods:  COAL~\cite{tan2020class}, PCT~\cite{tanwisuth2021prototype}; 4) SF-UDA methods: SHOT~\cite{liang2020shot}, BAIT~\cite{Yang2020UnsupervisedDA}, NRC~\cite{yang2021exploiting}, our CPGA; 5) imbalanced SF-UDA method: ISFDA~\cite{li2021imbalanced}.

\newl

\noindent\textbf{Implementation details.}
We implement all the baselines based on their official codes or reimplementation\footnote{\url{https://github.com/thuml/Transfer-Learning-Library}}. For the network architecture, we use RetNet-50, pre-trained on ImageNet, as the backbone for Office-Home-I and DomainNet-S, while adopting ResNet-101 for VisDA-I.
Due to the page limitation, we provide more implementation details in the supplementary.

\newl

\noindent\textbf{Evaluation protocol.}
We use overall accuracy to measure how well the model matches the target class distribution, and also adopt average per-class accuracy for evaluation. 
Due to the page limitation, we put the results in terms of overall accuracy in the main paper, and more detailed results in terms of per-class accuracy in Appendix \mata{D}.

\subsection{Results of Imbalance-agnostic SF-UDA}
We verify the effectiveness of our \ournet~in handling diverse class distribution shifts on three datasets, \ie{Office-Home-I, DomainNet-S and VisDA-I}.  Specifically, on Office-Home-I,  we present the results on  six types of class distribution shifts regarding the Cl$\rightarrow$Pr task in Table~\ref{tab:oh_cl2pr_oa} and those  regarding the Cl$\rightarrow$Rw  task in Table~\ref{tab:oh_cl2rw_oa}, while  the results for other tasks (\eg Pr$\rightarrow$Rw) are provided in Appendix \mata{E}. 
Moreover, we report the results on DomainNet-S in Table~\ref{tab:overall-domainnet}, where each task corresponds to a distinct type of class distribution shift.

In light of the results on Office-Home-I and DomainNet-S, we draw the following observations: 
1) UDA and CI-UDA methods are incapable to alleviate the domain discrepancy when confronted with agnostic class distribution shifts, which leads to relatively poor performance.
2) Recent state-of-the-art SF-UDA methods outperform UDA and CI-UDA methods, but they assume implicitly that the source and target domains are class-balanced. As a result, these methods exhibit inadequate performance in imbalance-agnostic SF-UDA.
3) ISFDA~\cite{li2021imbalanced} is a better SF-UDA method when compared to other SF-UDA methods. ISFDA considers two opposite class distributions (FLT$\rightarrow$BLT and BLT$\rightarrow$FLT), resulting in better performance in the two tasks than other SF-UDA baselines, as evidenced in Tables~\ref{tab:oh_cl2pr}-\ref{tab:oh_cl2rw}. However, ISFDA depends on the prior of the source class distribution to train a class-balanced model, which is infeasible in real imbalance-agnostic SF-UDA. Furthermore, ISFDA cannot perform well on other types of class distribution shifts beyond FLT$\rightarrow$BLT and BLT$\rightarrow$FLT.
4) Unlike the above baselines, our proposed method, \ournet, demonstrates superior performance, indicating that it can accurately perceive the target class distribution and effectively leverage the source model's knowledge to solve imbalance-agnostic SF-UDA.

We further investigate the effectiveness of \ournet~under various imbalance ratios and report the results on VisDA-I with three ratios (\ie{10, 50, 100}) in  Figure~\ref{fig:visda_histograms_oa}. Specifically, our \ournet~achieves the best performance on all ratios and maintains stable performance even if the imbalance ratio is 100, whereas baselines suffer from performance degradation when the imbalance ratio is high. This further demonstrates the practicability of \ournet~in handling wide imbalance ratio scenarios of imbalance-agnostic SF-UDA.

\begin{figure}[t]
\centering
 \begin{minipage}{0.32\linewidth}
 \subfigure[Source-only]{
 \includegraphics[width=1.0\linewidth]{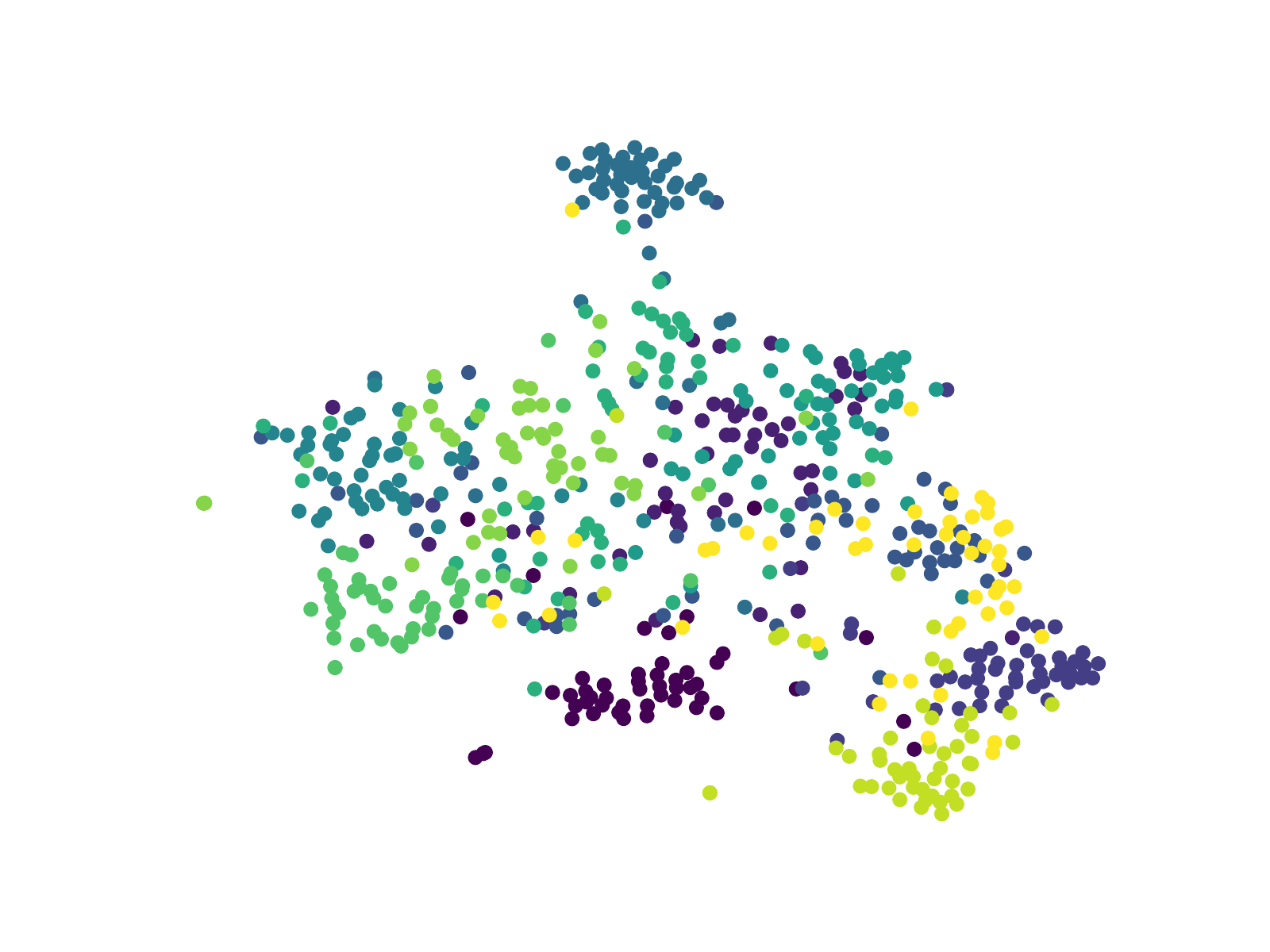}
 \label{vis:so-tsne}
 }
 \end{minipage}
 \begin{minipage}{0.32\linewidth}
 \subfigure[CLIP]{
 \includegraphics[width=1.0\linewidth]{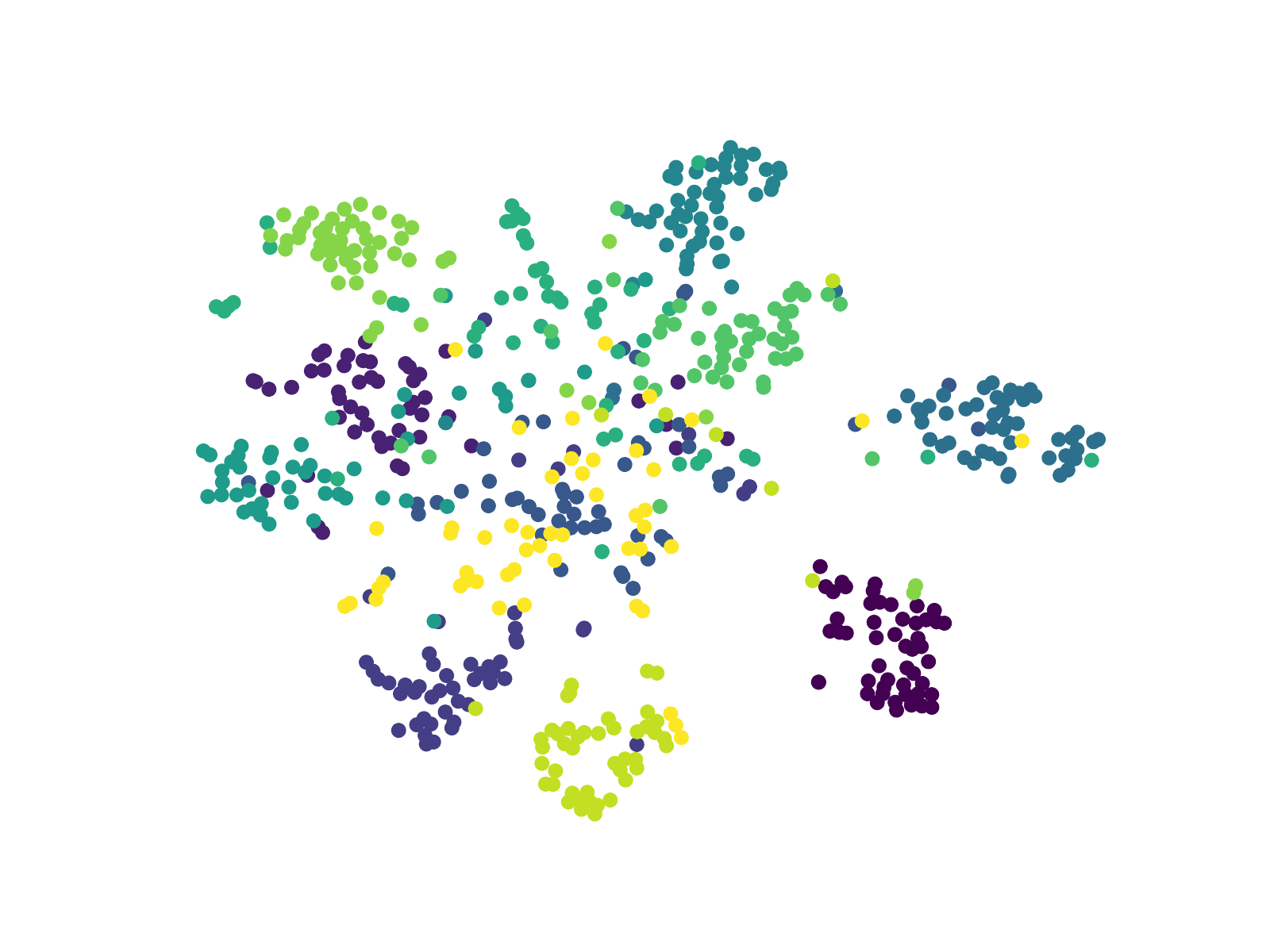}
 \label{vis:CLIP-tsne}
 }
 \end{minipage}
 \begin{minipage}{0.32\linewidth}
 \subfigure[T-CPGA]{
 \includegraphics[width=1.0\linewidth]{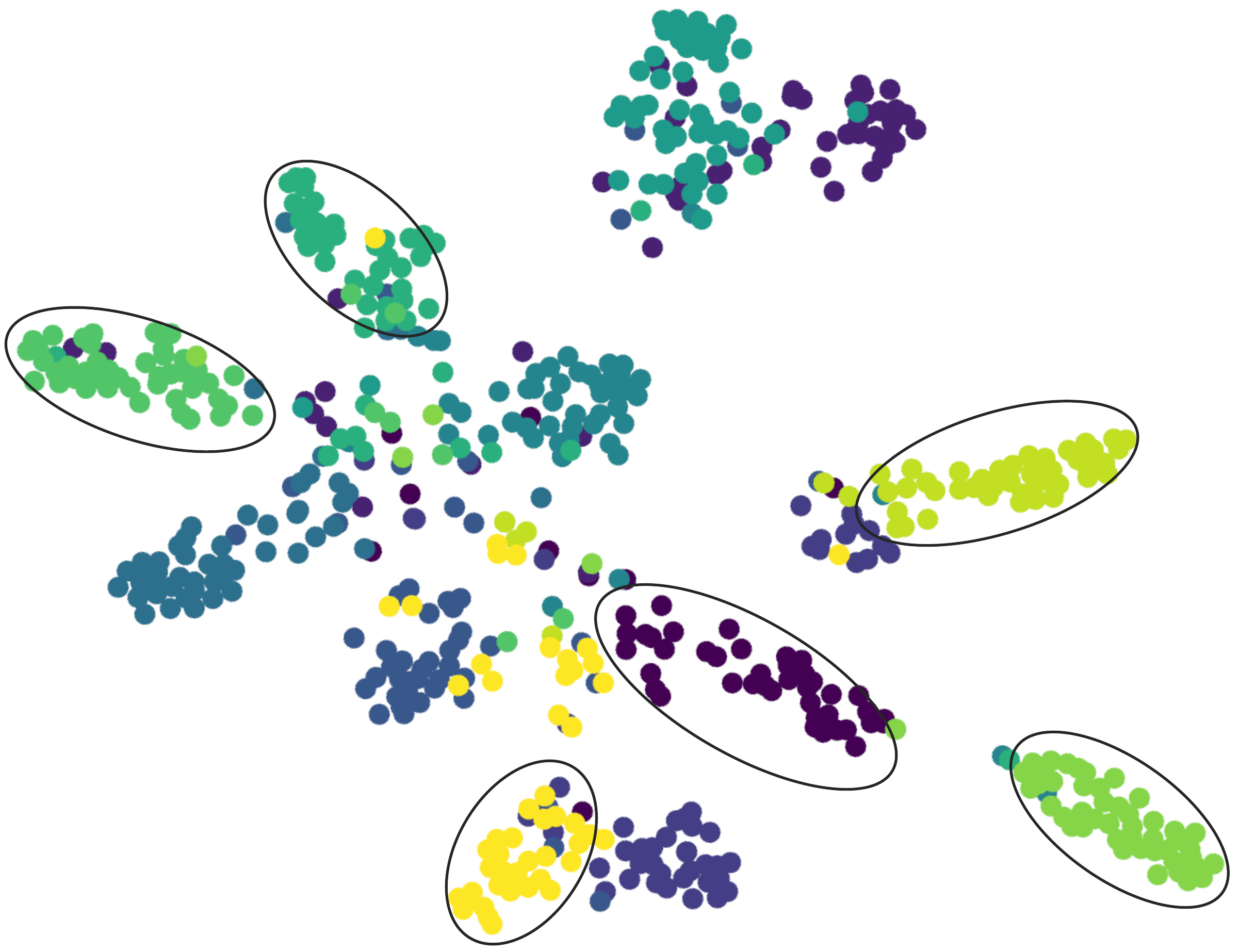}
 \label{vis:ours-tsne}
 }
 \end{minipage}
 \caption{The t-SNE visualizations on the \textbf{VisDA-I} validation set (\ie{FLT$\rightarrow$BLT, imbalance ratio 100}) generated by the source pre-trained model (ResNet-101), CLIP zero-shot prediction and our T-CPGA. Since different colors represent different classes.}
 \label{vis:tsne_total}
\end{figure}

We further use t-SNE~\cite{maaten2008visualizing} to visualize the features learned by the source-only model (ResNet-101), the CLIP model, and the model trained by our T-CPGA. We randomly selected 50 samples per class from the validation set of VisDA-I (FLT$\rightarrow$BLT, imbalance ratio 100) for visualization. As shown in Figure~\ref{vis:tsne_total}, the feature distribution of the source-only model appears chaotic, while CLIP is only slightly better than the source-only model. In contrast, the feature distribution of \ournet~is more discriminative, exhibiting both intra-class compactness and inter-class separation.
This is achieved by our target-aware contrastive prototype alignment strategy. Note that previous work~\cite{wang2021contrastive} has shown  
that learning discriminative image representations can facilitate classifier learning in imbalanced cases~\cite{wang2021contrastive}. Therefore, this visualization analysis further confirms the effectiveness of \ournet~in addressing imbalance-agnostic SF-UDA.

\subsection{Discussions}
\noindent
\textbf{{Discussion on CLIP.}}
\label{context:clip_zero-shot}
One might wonder why we do not use CLIP directly to classify target samples in imbalance-agnostic SF-UDA, given its impressive performance in other settings. However, our proposed method, \ournet, offers two significant advantages in real-world imbalance-agnostic SF-UDA applications. First,  \ournet~has better performance over CLIP in various imbalance-agnostic UDA datasets.  As shown in Table~\ref{tab:clip-zeroshot}, \ournet~is more effective than CLIP zero-shot prediction, as it can generate more discriminative feature representation for classification (cf. Figure~\ref{vis:CLIP-tsne} vs \ref{vis:ours-tsne}), and generate more accurate pseudo labels for domain alignment (cf. Table~\ref{tab:abolation-domainnet}).  It is important to note that simply fine-tuning CLIP cannot achieve better performance in imbalance-agnostic SF-UDA due to the lack of true target annotations. Figure \mata{E.1} (cf. Appendix \mata{E}) demonstrates that fine-tuning CLIP with self-training  (with inevitable noisy pseudo labels) yields declining performance compared to CLIP zero-shot prediction. In contrast, \ournet~employs target-aware contrastive prototype alignment to mitigate the risk of memorizing noisy labels, making it more suitable for imbalance-agnostic SF-UDA. Second, \ournet~can be used to train various model architectures (cf. Table \mata{E.1}, Appendix \mata{E}), making it more suitable for real-world scenarios where model size may be limited due to hardware constraints, such as mobile terminals. Unfortunately, publicly available CLIP checkpoints only support ResNet-50, ViT-B/32, or even larger models, which may not be feasible to use in these scenarios. Therefore, instead of using CLIP directly, we propose a new \ournet~that is more applicable to real imbalance-agnostic SF-UDA scenarios.

\newl 

\begin{table}[t] 
\setlength\tabcolsep{4pt}
 \begin{center}
 \caption{\label{tab:clip-zeroshot} Compare CLIP zero-shot prediction and our~\ournet~on the \textbf{Office-Home-I} (ResNet-50) and \textbf{VisDA-I} (ResNet-101) datasets in terms of \textbf{Overall} Accuracy (\%).}
 \scalebox{0.7}{ 
 \begin{tabular}{lcccccccccccccccc}
 \toprule
 Method 
 & Office-Home-I
 & VisDA-I-10
 & VisDA-I-50
 & VisDA-I-100
 & {Avg.} \\
 \midrule
 CLIP (zero-shot prediction)~\cite{radford2021learning} & 76.14 & 84.76 & 84.55 & 84.56 & 82.50 \\
 \ournet~(Ours) & \textbf{77.71} & \textbf{89.45} & \textbf{90.24} & \textbf{90.83} & \textbf{87.06} \\
 \bottomrule
 \end{tabular}}
 \end{center}
\end{table}

\noindent
\textbf{{Target label-distribution-aware classifier.}}
 As we mentioned in Section~\ref{sec5-2}, unidentified class distribution shifts would cause the fixed source classifier to provide unreliable predictions. Therefore, we devise a target label-distribution-aware classifier that enables \ournet~to match the target label distribution and accurately classify target samples. This design can be verified by the results in Table~\ref{tab:abolation-domainnet}, where  our \ournet~with the target label-distribution-aware classifier performs much better  than that without this classifier on  DomainNet-S.

\section{Conclusion}
In this paper, we have proposed a Contrastive Prototype Generation and Adaptation (CPGA) method to resolve SF-UDA. Specifically, we overcome the lack of source data by generating feature prototypes for each class via contrastive learning in the first stage. Based on the generated prototypes, we develop a robust contrastive prototype adaptation strategy to mitigate domain shifts and pseudo label noise in the second stage. Extensive experiments on three benchmark datasets have demonstrated the effectiveness of CPGA in handling SF-UDA. In addition to SF-UDA, we have explored a more practical task, namely imbalance-agnostic SF-UDA, where the class distribution does not necessarily be balanced. 
To address it, we have extended CPGA to Target-aware Contrastive Prototype Generation and Adaptation (T-CPGA). 
Like CPGA, T-CPGA consists of two stages: 1) it holds the same first stage as CPGA to handle the absence of source data. 2) To avoid the negative effect of the unidentified class distribution shift, we design a novel target-aware contrastive prototype alignment strategy.  Extensive experiments on three UDA variant datasets verify the effectiveness of \ournet~in handling imbalance-agnostic SF-UDA.
\ifCLASSOPTIONcaptionsoff
 \newpage
\fi

\bibliographystyle{IEEEtran}
{
    \bibliography{tcpga}
}

\clearpage


\section*{Supplementary Materials for ``Imbalance-Agnostic Source-Free Domain Adaptation via Avatar Prototype Alignment"}

In this supplementary, we first provide more discussions on the conventional Unsupervised Domain Adaptation (UDA) methods. In addition, we also provide more implementation details and more experimental results for both CPGA and T-CPGA.
The organization of the supplementary materials is as follows:

\begin{enumerate}[]
    \item In Appendix \mata{\ref{vanilla_UDA}}, we review the literature on vanilla unsupervised domain adaptation methods.
    \item In Appendix \mata{\ref{sec:elr}}, we provide more details of the early learning regularization term $\mathcal{L}_{elr}$.
    \item In Appendix \mata{\ref{sec:implement}}, we provide more implementation details of both CPGA and T-CPGA.
    \item In Appendix \mata{\ref{sec:exp}}, we provide more detailed experimental results of CPGA.
    \item In Appendix \mata{\ref{sec:exp_tcpga}}, we provide more detailed experimental results of T-CPGA.
\end{enumerate}

\appendices
\counterwithin{figure}{section}
\counterwithin{table}{section} 
\counterwithin{equation}{section} 

\section{Review of vanilla UDA}
\label{vanilla_UDA}
Unsupervised domain adaptation (UDA) seeks to leverage a label-rich source domain to improve the model performance on an unlabeled target domain~\cite{Yan2017LearningDC,Liang2019DistantSC,tang2020unsupervised,Zhang2020CollaborativeUD}.
In this field, Most existing methods alleviate the domain discrepancy either by adding adaptation layers to match high-order moments of distributions, \eg DDC~\cite{Tzeng2014DeepDC}, 
or by devising a domain discriminator to learn domain-invariant features in an adversarial manner, \eg DANN~\cite{ganin2015unsupervised} and MCD~\cite{saito2018maximum}. 
Recent adversarial-based approaches mainly focus on two levels, \ie feature-level and distribution-level. At the feature-level, ToAlign~\cite{wei2021toalign} proposes to select the corresponding source features to achieve task-oriented domain alignment via ignoring the task-irrelevant source features.
At the distribution-level, CLS~\cite{Liu21} proposes to align both conditional and class distribution shifts while MDD~\cite{zhang2019bridging} introduces Margin Disparity Discrepancy to measure distribution-level discrepancy which is subsequently minimized to facilitate domain alignment. 
Besides, prototypical methods and contrastive learning have also been introduced to UDA. For instance, TPN~\cite{pan2019transferrable}, PAL~\cite{hu2020panda} and PCT~\cite{tanwisuth2021prototype} attempt to align the source and target domains based on the learned prototypical feature representations. In addition, CAN~\cite{Kang2019ContrastiveAN} and CoSCA~\cite{Dai_2020_ACCV} leverage contrastive learning to minimize intra-class distance and maximize inter-class distance explicitly.
As CLIP has been successfully applied in recent studies, CLIP-based domain adaptation methods have emerged as well. 
For instance, AP~\cite{amortized21} adopts CLIP for domain generalization by combining domain prompt inference with CLIP. Additionally, StyleGAN-NADA~\cite{CLIP-guided} adopts CLIP for image generation via leveraging CLIP to discover global directions of disentangled change in the latent space.

Although conventional UDA methods continue to evolve and improve, the increasing emphasis on privacy protection laws has led to restrictions on the availability of source domain data. Furthermore, practical data may follow any class distributions rather than only relatively balanced class distributions.
To this end, we investigate a more practical task called
imbalance-agnostic SF-UDA. In this task, only a source pre-trained model and unlabeled target data are available, and the class distributions of both domains are unknown and could be arbitrarily skewed.

\section{Early Learning Regularization} \label{sec:elr}
To further prevent the model from memorizing noise, we propose to regularize the learning process via an early learning regularizer. Since DNNs first memorize the clean samples with correct labels and then the noisy data with wrong labels~\cite{arpit2017closer}, the model in the “early learning” phase can be more predictable to the noisy data. Therefore, we seek to use the early predictions of each sample to regularize learning.
To this end, we devise a memory bank $\mathcal{H}\small{=}\{\textbf{h}_{1}, \textbf{h}_{2},...,\textbf{h}_{n_t}\}$ to record non-parametric predictions of each target sample, and update them based on new predictions via a momentum strategy. Formally, for the $i$-th sample, we predict its non-parametric prediction regarding the $k$-th prototype by  
$o_{i,k} \small{=} \frac{\exp(\textbf{u}_{i}^{\top}\textbf{v}_k/\tau)}{\sum_{j=1}^{K}\exp(\textbf{u}_{i}^{\top} \textbf{v}_{j}/\tau)}$,
 and update the momentum by:
\begin{equation}
\textbf{h}_{i} \xleftarrow{} \beta \textbf{h}_{i} + (1-\beta)\textbf{o}_{i}, 
\end{equation}
where $\textbf{o}_{i}\small{=}[o_{i,1},...,o_{i,K}]$, and $\beta$ denotes the momentum coefficient. Based on the memory bank, for the $i$-th data, we further train the model via  an early learning regularizer $\mathcal{L}_{elr}$, proposed in~\cite{liu2020early}:
\begin{equation}
\label{loss:reg}
\mathcal{L}_{elr} = \log(1-\textbf{o}_{i}^{\top} \textbf{h}_{i}).
\end{equation}
This regularizer enforces the current prediction to be close to the prediction momentum, which helps to prevent overfitting to label noise. Note that the use of $\mathcal{L}_{elr}$ here is different from~\cite{liu2020early}, which focuses on classification tasks and uses parametric predictions.

\section{More Implementation Details} \label{sec:implement}
\begin{table}[t]
\renewcommand\arraystretch{1.5}
    \begin{center}
    \caption{\label{tab:generator}Detailed architecture of the generator, where $d$ denotes the output dimensions (\eg 2048) and $BS$ denotes the batch size.}
    \scalebox{0.7}{
        \begin{tabular}{c|c|c|c|c|c|c}
        \hline
        \hline
            \specialrule{0em}{1pt}{1pt}
            \multicolumn{7}{c}{ Backbone Network } \\ 
        \hline
        \hline
        Part & \multicolumn{2}{c|}{ Input $\rightarrow$ Output } & Kernel & Padding & Stride & Activation \\
        \hline
            Embedding & \multicolumn{2}{c|}{ $(BS, 1)$ $\rightarrow$ $(BS, 100)$ } & - & - & - & -\\
            \hline
            Linear & \multicolumn{2}{c|}{ $(BS, 100)$ $\rightarrow$ $(BS, 1024)$ } & - & - & - & ReLU\\
            \hline
            BatchNorm1d & \multicolumn{2}{c|}{ $(BS, 1024)$ $\rightarrow$ $(BS, 1024)$ } & - & - & - & -\\
            \hline
            Linear & \multicolumn{2}{c|}{ $(BS, 1024)$ $\rightarrow$ $(BS, \frac{d}{4}*7*7)$ } & - & - & - & ReLU\\
            \hline
            BatchNorm1d & \multicolumn{2}{c|}{ $(BS, \frac{d}{4}*7*7)$ $\rightarrow$ $(BS, \frac{d}{4}*7*7)$ } & - & - & - & -\\
            \hline
            Reshape & \multicolumn{2}{c|}{ $(BS, \frac{d}{4}*7*7)$ $\rightarrow$ $(BS, \frac{d}{4}, 7, 7)$ } & - & - & - & -\\
            \hline
            ConvTranspose2d & \multicolumn{2}{c|}{ $(BS, \frac{d}{4}, 7, 7)$ $\rightarrow$ $(BS, \frac{d}{8}, 6, 6)$ } & 2 & 1 & 2 & -\\
            \hline
            BatchNorm2d & \multicolumn{2}{c|}{ $(BS, \frac{d}{8}, 6, 6)$ $\rightarrow$ $(BS, \frac{d}{8}, 6, 6)$ } & - & - & - & ReLU\\
            \hline
            ConvTranspose2d & \multicolumn{2}{c|}{ $(BS, \frac{d}{8}, 6, 6)$ $\rightarrow$ $(BS, \frac{d}{16}, 4, 4)$ } & 3 & 1 & 2 & -\\
            \hline
            BatchNorm2d & \multicolumn{2}{c|}{ $(BS, \frac{d}{16}, 4, 4)$ $\rightarrow$ $(BS, \frac{d}{16}, 4, 4)$ } & - & - & - & ReLU\\
            \hline
            Reshape & \multicolumn{2}{c|}{ $(BS, \frac{d}{16}, 4, 4)$ $\rightarrow$ $(BS, d)$ } & - & - & - & -\\
            \hline
            \hline
         \end{tabular}}
    \end{center}
\end{table}

\noindent{\textbf{Architecture of the generator.}}
As shown in Table~\ref{tab:generator}, the generator consists of an embedding layer, two FC layers and two deconvolution layers. 
Similar to ACGAN~\cite{odena2017conditional}, given an input noise $\textbf{z}\small{\sim} U(0,1)$ and a label $\textbf{y}\small{\in}\mathbb{R}^K$, we first map the label into a vector using the embedding layer. After that, we combine the vector with the given noise by element-wise multiplication and then feed it into the following layers.
Since we propose to obtain feature prototypes instead of images, we reshape the output of the generator into a feature vector with the same dimensions as the last FC layer.

\newl

\noindent{\textbf{Training of the generator.}}
In stage one, we train the generator by optimizing $\mathcal{L}_{ce}\small{+}\mathcal{L}_{con}^{p}$.
The batch size is set to 128.
We use the SGD optimizer with learning rate $=$ 0.001.
In  stage two, to achieve class-wise domain alignment, we generate feature prototypes for K classes in each epoch.

\newl

\noindent{\textbf{Target neighborhood clustering.}}
To enhance the contrastive alignment, we further resort to feature clustering to make the target features more compact. Inspired by~\cite{Saito2020UniversalDA} that the intra-class samples in the same domain are generally more closer, we propose to close the distance between each target sample and its nearby neighbors.  
To this end, we maintain a memory bank $\mathcal{Q}\small{=} \{\textbf{q}_{1}, \textbf{q}_{2}, ...,\textbf{q}_{n_{t}}\}$ to restore all target features, which are updated when new features are extracted  in each iteration. 
Based on the bank, for the $i$-th sample's feature $\textbf{q}_i$, we can compute its normalized similarity with any feature $\textbf{q}_j$ by 
$\textbf{s}_{i,j} \small{=} \frac{\exp(\phi(\textbf{q}_i,\textbf{q}_j)/\tau)}{\sum_{l=1, l\neq i}^{n_{t}}\exp(\phi(\textbf{q}_i, \textbf{q}_{l})/\tau)}$.
Motivated by that minimizing the entropy of the normalized similarity helps to learn compact features for similar data~\cite{Saito2020UniversalDA},  we further train the extractor via a neighborhood clustering term:
\begin{equation}
\label{loss:nc}
\mathcal{L}_{nc} = -\sum_{j=1, j\neq i}^{n_t} \textbf{s}_{i, j} \log(\textbf{s}_{i, j}).
\end{equation}
Note that the entropy minimization here does not use pseudo labels, so the learned compact target features are (to some degree) robust to pseudo label noise.

\newl

\noindent{\textbf{Implementation details of CPGA.}}
We set the learning rate and epoch to 0.01 and 40 for VisDA and to 0.001 and 400 for Office-31 and Office-Home. 
For hyper-parameters, we set $\eta$, $\beta$, $\tau$ and batch size to 0.05, 0.9, 0.07 and 64, respectively. Besides, we set $\lambda \small{=}7$ for Office-31 and Office-home while $\lambda \small{=} 5$ for VisDA. Following~\cite{Xu2020GenerativeLD}, the dimension of noise $\textbf{z}$ is 100.

\newl

\noindent{\textbf{Implementation details of T-CPGA.}}
The new added target label-distribution-aware classifier $C_{lda}$ is a fully connected layer. We set the learning rate and epoch to 0.001 and 300 for Office-Home-I and to 0.01 and 400 for VisDA-I and DomainNet-S. For hyper-parameters, we set the same values as in CPGA.

\section{More Experimental Results of CPGA} \label{sec:exp}

\begin{figure}[t] 
\centering
\begin{minipage}{0.49\linewidth}
\subfigure[Total loss curve]{
\includegraphics[width=3.9cm]{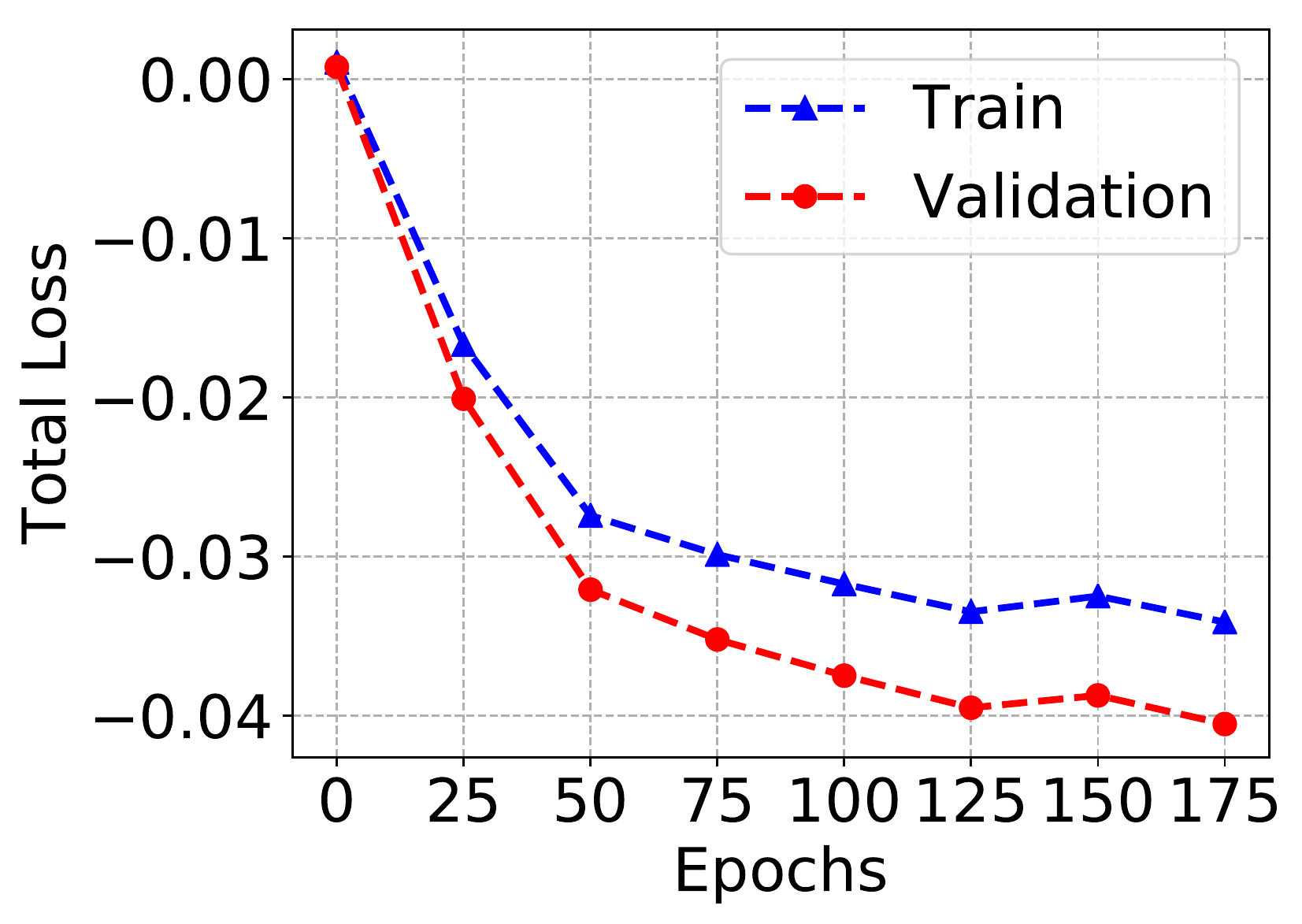}
\label{vis:lce1}
}
\end{minipage}
\begin{minipage}{0.49\linewidth}
\subfigure[Accuracy curve]{
\includegraphics[width=3.9cm]{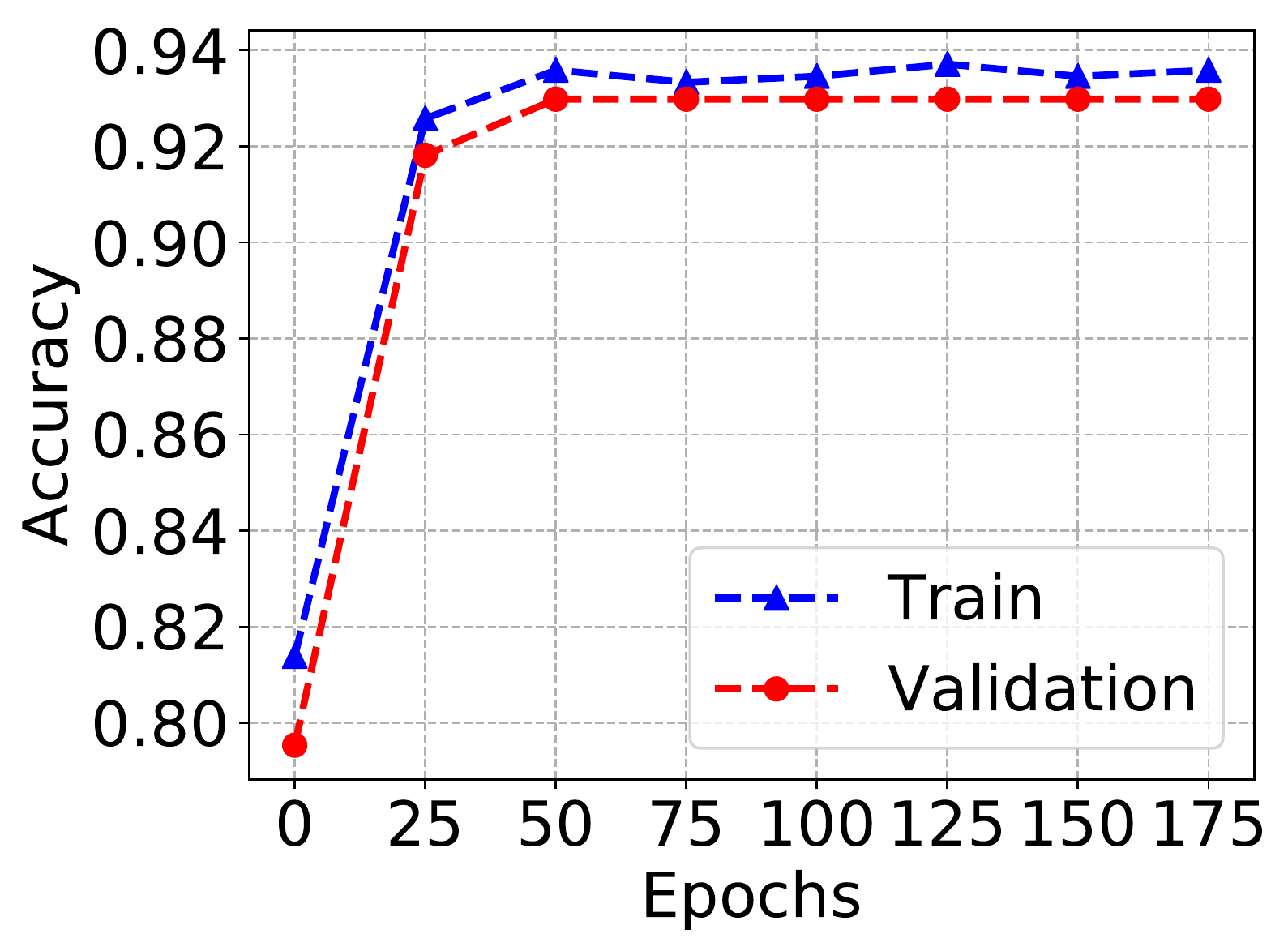}
\label{vis:lcec1}
}
\end{minipage}
\label{figdata1}
\vspace{-0.1in}
\caption{Optimization curves of CPGA on \textbf{Office-31}(A$\rightarrow$W).}
\label{vis:curve2}
\end{figure}

\begin{figure}[t]
		\begin{center}
		\includegraphics[width=4.15cm]{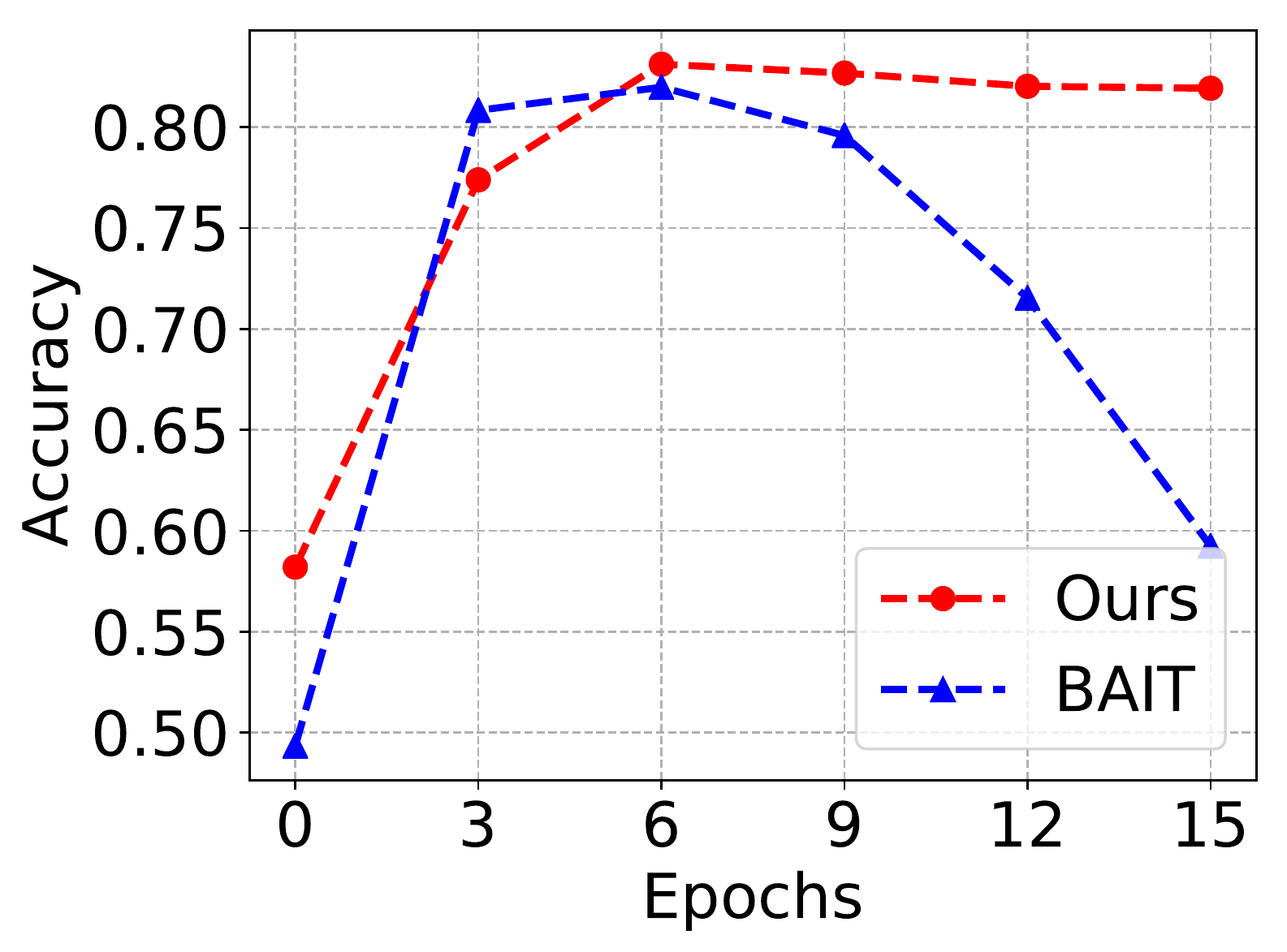}
		\vspace{-0.1in}
		\caption{Testing curves of CPGA and BAIT on \textbf{VisDA} dataset.}
		\label{fig:noise}
		\end{center}
\end{figure}

\begin{table*}[t]
\setlength\tabcolsep{5pt}
    \begin{center}
    \caption{\label{tab:visda_sup}Classification accuracies (\%) on large-scale VisDA dataset (ResNet-101). We adopt underlines to denote reimplemented results.}
    \scalebox{0.85}{
         \begin{tabular}{lcccccccccccccc}
         \toprule
         Method & Source-free & plane & bicycle & bus & car & horse & knife & mcycl & person & plant & sktbrd & train & truck & Per-class\\
         \midrule
         SHOT~\cite{liang2020shot} & \cmark & 92.6 & 81.1 & 80.1 & 58.5 & 89.7 & 86.1 & 81.5 & 77.8 & 89.5 & 84.9 & 84.3 & 49.3 & 79.6\\
         SHOT~\cite{liang2020shot} & \cmark & 88.5 & 85.9 & 77.9 & 49.8 & 90.2 & 90.8 & 82.0 & 79.0 & 88.5 & 84.4 & 85.6 & 50.5 & \underline{79.4}\\
         BAIT~\cite{Yang2020UnsupervisedDA} & \cmark & 93.7 & 83.2 & 84.5 & 65.0 & 92.9 & 95.4 & 88.1 & 80.8 & 90.0 & 89.0 & 84.0 & 45.3 & 82.7 \\
         BAIT~\cite{Yang2020UnsupervisedDA} & \cmark & 93.8 & 75.4 & \textbf{86.1} & 64.0 & \textbf{93.9} & 96.4 & 88.5 & 81.2 & 88.9 & 88.7 & 86.9 & 39.9 & \underline{82.0}\\
        \midrule
        CPGA~(Ours) & \cmark & \textbf{95.6} & \textbf{89.0} & 75.4 & 64.9 & 91.7 & \textbf{97.5} & 89.7 & \textbf{83.8} & \textbf{93.9} & \textbf{93.4} & 87.7 & \textbf{69.0} & \textbf{86.0} \\
        \bottomrule
        \end{tabular}
    }
    \end{center}
\end{table*}

\begin{table*}[t]
\setlength\tabcolsep{6pt}
    \begin{center}
    \caption{\label{tab:office-home_sup} Classification accuracies (\%) on the Office-Home dataset (ResNet-50). We adopt underlines to denote reimplemented results.}
    \scalebox{0.7}{  
         \begin{tabular}{lcccccccccccccc}
         \toprule
         Method & Source-free & Ar$\rightarrow$Cl & Ar$\rightarrow$Pr & Ar$\rightarrow$Rw & Cl$\rightarrow$Ar & Cl$\rightarrow$Pr & Cl$\rightarrow$Rw & Pr$\rightarrow$Ar & Pr$\rightarrow$Cl & Pr$\rightarrow$Rw & Rw$\rightarrow$Ar & Rw$\rightarrow$Cl & Rw$\rightarrow$Pr & Avg.\\
         \midrule
         ResNet-50~\cite{He2016DeepRL} & \xmark & 34.9 & 50.0 & 58.0 & 37.4 & 41.9 & 46.2 & 38.5 & 31.2 & 60.4 & 53.9 & 41.2 & 59.9 & 46.1 \\
         MCD~\cite{saito2018maximum} & \xmark & 48.9 & 68.3 & 74.6 & 61.3 & 67.6 & 68.8 & 57.0 & 47.1 & 75.1 & 69.1 & 52.2 & 79.6 & 64.1 \\
         CDAN~\cite{long2018conditional} & \xmark & 50.7 & 70.6 & 76.0 & 57.6 & 70.0 & 70.0 & 57.4 & 50.9 & 77.3 & 70.9 & 56.7 & 81.6 & 65.8 \\
         MDD~\cite{zhang2019bridging} & \xmark & 54.9 & 73.7 & 77.8 & 60.0 & 71.4 & 71.8 & 61.2 & 53.6 & 78.1 & 72.5 & 60.2 & 82.3 & 68.1 \\
         BNM~\cite{Cui2020TowardsDA} & \xmark & 52.3 & 73.9 & 80.0 & 63.3 & 72.9 & 74.9 & 61.7 & 49.5 & 79.7 & 70.5 & 53.6 & 82.2 & 67.9 \\
         BDG~\cite{yang2020bi} & \xmark & 51.5 & 73.4 & 78.7 & 65.3 & 71.5 & 73.7 & 65.1 & 49.7 & 81.1 & 74.6 & 55.1 & 84.8 & 68.7 \\
         SRDC~\cite{tang2020unsupervised} & \xmark & 52.3 & 76.3 & 81.0 & 69.5 & 76.2 & 78.0 & 68.7 & 53.8 & 81.7 & 76.3 & 57.1 & 85.0 & 71.3 \\
         \midrule
         PrDA~\cite{kim2020progressive} & \cmark & 48.4 & 73.4 & 76.9 & 64.3 & 69.8 & 71.7 & 62.7 & 45.3 & 76.6 & 69.8 & 50.5 & 79.0 & 65.7 \\
         SHOT~\cite{liang2020shot} & \cmark & 56.9 & 78.1 & 81.0 & 67.9 & \textbf{78.4} & \textbf{78.1} & 67.0 & 54.6 & 81.8 & 73.4 & 58.1 & \textbf{84.5} & 71.6 \\
         SHOT~\cite{liang2020shot} & \cmark & 57.5 & 77.9 & 80.3 & 66.5 & 78.3 & 76.6 & 65.8 & 55.7 & 81.7 & 74.0 & 61.2 & 84.2 & \underline{71.6} \\
         BAIT~\cite{Yang2020UnsupervisedDA} & \cmark & 57.4 & 77.5 & \textbf{82.4} & \textbf{68.0} & 77.2 & 75.1 & \textbf{67.1} & 55.5 & \textbf{81.9} & \textbf{73.9} & 59.5 & 84.2 & 71.6 \\
         BAIT~\cite{Yang2020UnsupervisedDA} & \cmark & 52.2 & 71.3 & 72.5 & 59.9 & 70.6 & 69.9 & 60.3 & 53.9 & 78.2 & 68.4 & 58.9 & 80.7 & \underline{66.4} \\
         \midrule
         CPGA(ours) & \cmark & \textbf{59.3} & \textbf{78.1} & 79.8 & 65.4 & 75.5 & 76.4 & 65.7 & \textbf{58.0} & 81.0 & 72.0 & \textbf{64.4} & 83.3 & \textbf{71.6} \\
         \bottomrule
         \end{tabular}}
    \end{center}
\end{table*}

\begin{table}[t]
\setlength\tabcolsep{20pt}
\begin{center}
    \caption{\label{tab:elrpara}
 Influence of the trade-off parameters $\beta$ and $\lambda$ in terms of per-class accuracy (\%) on \textbf{VisDA}. The value of $\beta$ is chosen from $[0.5, 0.7, 0.9, 0.99]$ and $\lambda$ is chosen from $[3, 5, 7]$. In each experiment, the rest of hyper-parameters are fixed to the values mentioned in the main paper.}
 \scalebox{0.85}{
    \begin{tabular}{ccccc}
    \toprule
    \multirow{2}{*}{$\lambda$} &
    \multicolumn{4}{c}{$\beta$}\cr
    \cmidrule(lr){2-5}
    &0.5&0.7&0.9&0.99\cr
    \midrule
    3 & 81.2 & 83.0 & 83.9 & 83.0\cr
    5 & 81.3 & 82.2 & 84.1 & 83.2\cr
    7 & 79.7 & 81.6 & 83.3 & 83.0\cr
    \bottomrule
    \end{tabular}
    }
    \end{center}
\end{table}

\begin{table}[t]
\setlength\tabcolsep{5pt}
    \begin{center}
    \caption{\label{tab:office_sup}Classification accuracies (\%) on the Office-31 dataset (ResNet-50). We adopt underlines to denote reimplemented results.}
    \scalebox{0.8}{
         \begin{tabular}{lcccccccl}
         \toprule
         Method & Source-free & A$\rightarrow$D & A$\rightarrow$W & D$\rightarrow$W & W$\rightarrow$D & D$\rightarrow$A & W$\rightarrow$A & Avg.\\
         \midrule
         SHOT~\cite{liang2020shot} & \cmark & 93.1 & 90.9 & \textbf{98.8} & 99.9 & 74.5 & 74.8 & 88.7 \\
         SHOT~\cite{liang2020shot} & \cmark & 91.4 & 90.0 & 99.1 & 100.0 & 74.8 & 73.6 & \underline{88.2} \\
         BAIT~\cite{Yang2020UnsupervisedDA} & \cmark & 92.0 & \textbf{94.6} & 98.1 & \textbf{100.0} & 74.6 & 75.2 & 89.1 \\
         BAIT~\cite{Yang2020UnsupervisedDA} & \cmark & 91.3 & 87.4 & 97.6 & 99.7 & 71.4 & 67.2 & \underline{85.8} \\
         \midrule
         CPGA~(Ours) & \cmark & \textbf{94.4} & 94.1 & 98.4 & 99.8 & \textbf{76.0} & \textbf{76.6} & \textbf{89.9} \\
         \bottomrule
         \end{tabular}
         }
    \end{center}
\end{table}

\begin{table}[t]
\renewcommand\arraystretch{1.00}
\setlength\tabcolsep{6pt}
\begin{center}
    \caption{
     \label{tab:para_sen}Influence of the trade-off parameter $\lambda$ and $\eta$ in terms of per-class accuracy (\%) on \textbf{VisDA}. The value of $\lambda$ is chosen from $[1, 3, 5, 7, 9]$ and $\eta$ is chosen from $[0.001, 0.005, 0.01, 0.05, 0.1]$. In each experiment, the rest of the hyper-parameters are fixed.
      }
  \scalebox{0.8}{
    \begin{tabular}{ccccccccccc}
    \toprule
    \multirow{2}{*}{Parameter} &
    \multicolumn{5}{c}{$\lambda$}&
    \multicolumn{5}{c}{$\eta$}\cr
    \cmidrule(lr){2-6} \cmidrule(lr){7-11}
    &1 & 3& 5& 7& 9&
    0.001& 0.005& 0.01 &0.05&0.1\cr
    \midrule
    Acc.  & 83.3 & 85.0 & \textbf{86.0} & 85.5 & 85.3 & 85.5 & 85.6 & 85.5 & \textbf{86.0} &83.0 \cr
    \bottomrule
    \end{tabular}
    }
    \end{center}
\end{table}

\noindent{\textbf{Comparison with SOTA methods of CPGA.}}
We verify the effectiveness of our method on the Office-Home dataset. From Table~\ref{tab:office-home_sup}, the results show that: (1) CPGA outperforms all the conventional unsupervised domain adaptation methods which need to use the source data. (2) CPGA achieves the competitive performance compared with the state-of-the-art source-free UDA methods, \ie SHOT~\cite{liang2020shot} and BAIT~\cite{Yang2020UnsupervisedDA}. Besides, we also provide our reimplemented results of the published source-free UDA methods on VisDA and Office-31 based on their published source codes (See Table~\ref{tab:visda_sup} and Table~\ref{tab:office_sup}).

\newl

\noindent{\textbf{Influence of hyper-parameters of CPGA.}}
On the one hand,  we evaluate the sensitivity of two hyper-parameters $\lambda$ and $\eta$ on VisDA via an unsupervised reverse validation strategy~\cite{ganin2016domain} based on the source prototypes. 
For convenience, we set $\eta=0.05$ when studying $\lambda$, and set $\lambda=5$ when studying $\eta$. As shown in Table~\ref{tab:para_sen}, the proposed method achieves the best performance when setting $\lambda = 5$ and $\eta = 0.05$ on VisDA. The results also demonstrate that our method is non-sensitive to the hyper-parameters.
On the other hand, we provide more results for the hyper-parameters $\lambda$ and $\beta$ on VisDA. As shown in Table~\ref{tab:elrpara}, our method achieves the best performance with the setting $\beta \small{=} 0.9$ and $\lambda \small{=} 5$ on VisDA.

\newl

\noindent{\textbf{Visualization of optimization curve of CPGA.}}
Figure~\ref{vis:curve2} shows our  method converges well in terms of the total loss and accuracy in the training  phase. Also, the curve on the validation set means our method does not suffer from pseudo label noise.

\newl

\noindent{\textbf{Compared CPGA with BAIT.}} 
As shown in Figure~\ref{fig:noise}, BAIT~\cite{Yang2020UnsupervisedDA} may overfit to mistaken divisions of certain and uncertain sets, leading to poor generalization abilities.
In contrast, our method is more robust and can conquer the issue of pseudo label noise.

\newl

\section{More Experimental Results of T-CPGA} \label{sec:exp_tcpga}

\begin{figure}[t]
\centering
 \begin{minipage}{0.49\linewidth}
 \subfigure[Product]{
 \includegraphics[width=0.95\linewidth]{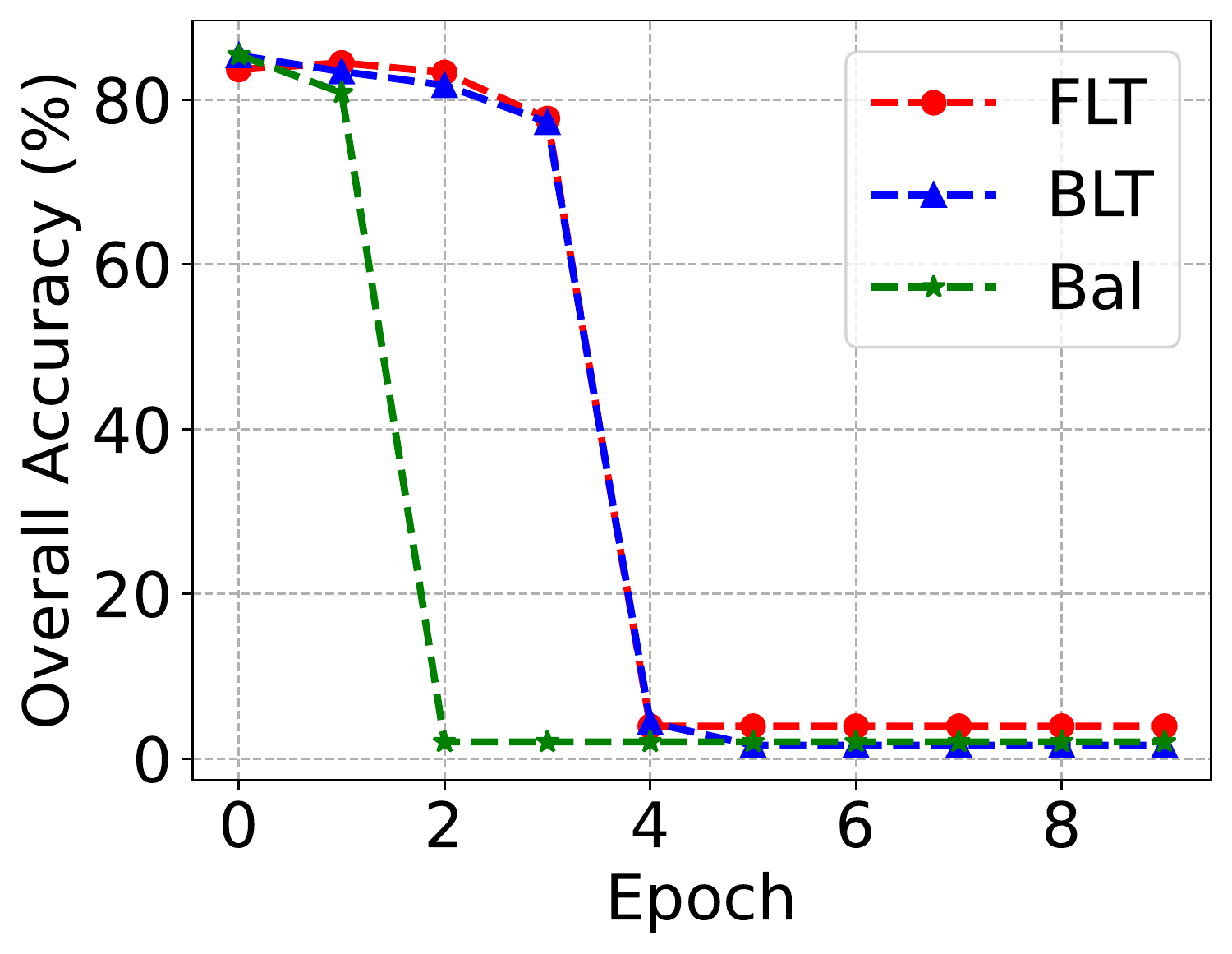}
 \label{vis:so-ft1}
 }
 \end{minipage}
 \begin{minipage}{0.49\linewidth}
 \subfigure[Real-World]{
 \includegraphics[width=0.95\linewidth]{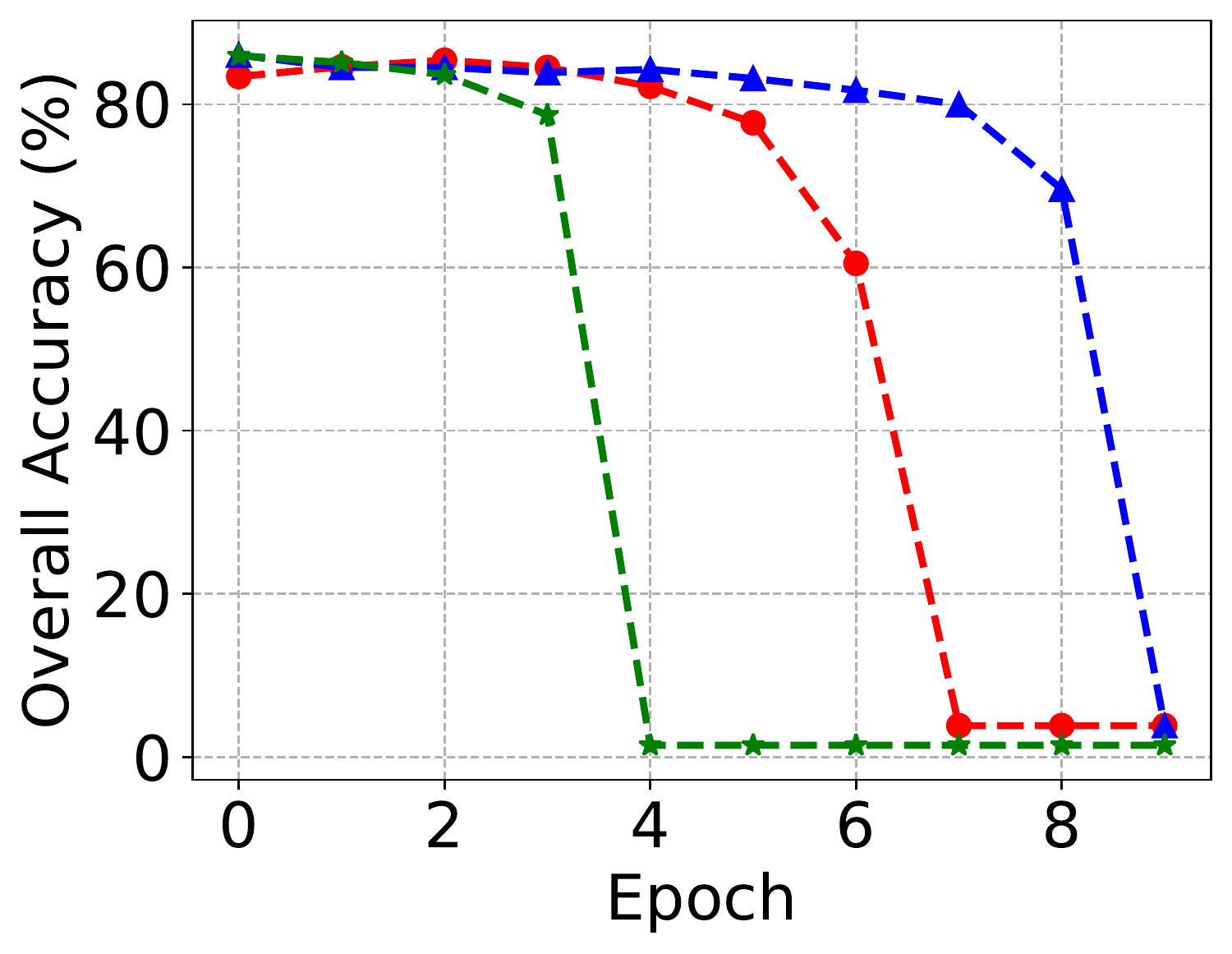}
 \label{vis:CLIP-ft2}
 }
 \end{minipage}
 \caption{Overall Accuracy of fine-tuned CLIP on the \textbf{Product} and \textbf{Real-World} domains with increasing epochs. Here, \textbf{FLT}, \textbf{BLT} and \textbf{Bal} denote the type of the label distribution.}
 \label{vis:clip_ft_total}
\end{figure}


\begin{figure*}[t]
\centering
\includegraphics[width=0.95\linewidth]{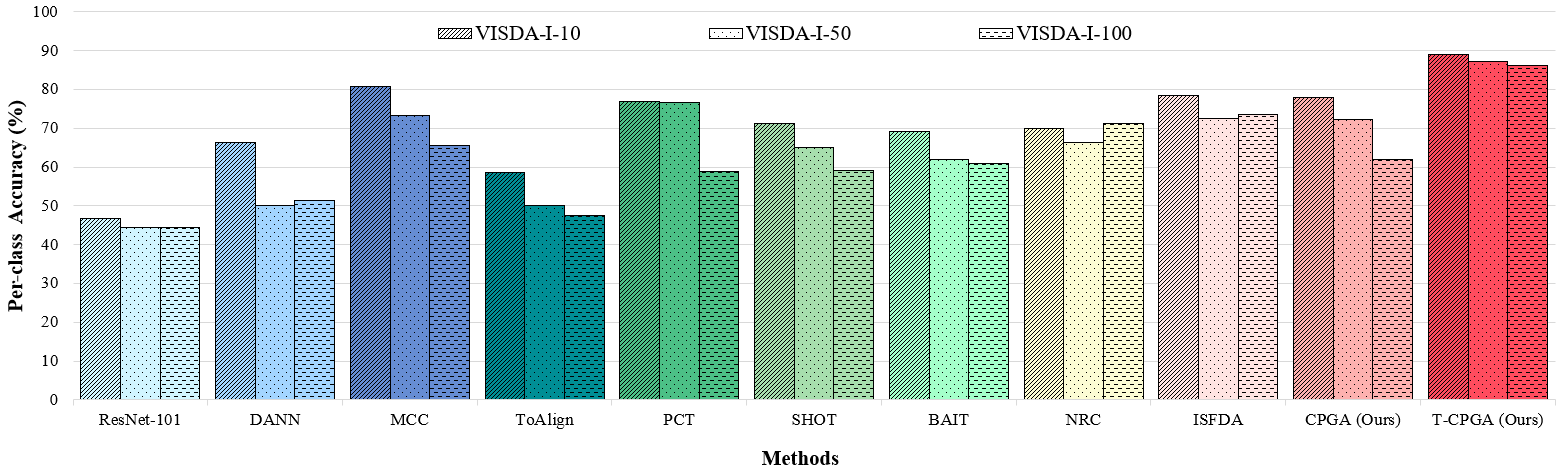}
\caption{\textbf{Per-Class} Accuracy (\%) on the \textbf{VisDA-I} dataset (ResNet-101). The number after VisDA-I is the imbalance ratio.}
\label{fig:visda_histograms}
\end{figure*}

\setlength\tabcolsep{3pt}
\begin{table*}[t]
    \caption{\label{tab:oh_mobileNet} \textbf{Overall} Accuracy (\%) of \textbf{Cl$\rightarrow$Pr} and \textbf{Cl$\rightarrow$Rw} tasks with different class distribution shifts on the \textbf{Office-Home-I} dataset (MobileNet-V2 and ResNet-50).}
    \begin{minipage}[t]{0.49\textwidth}
    \begin{center}
    \scalebox{0.7}{  
         \begin{tabular}{lcccccccccccccccc}
         \toprule
        \multirow{3}{*}{Method} &  \multicolumn{6}{c}{MobileNet-V2, Cl$\rightarrow$Pr} \\
        \cmidrule(lr){2-8}
        &
         {FLT$\rightarrow$FLT} &
         {FLT$\rightarrow$BLT} &
         {FLT$\rightarrow$Bal} &
         {BLT$\rightarrow$FLT} & 
         {BLT$\rightarrow$BLT} & 
         {BLT$\rightarrow$Bal} & 
         {Avg.} \\
         \midrule
         MobileNet-V2~\cite{sandler2018mobilenetv2} & 40.98  & 27.71  & 34.44  & 26.01  & 40.71  & 35.12  & 34.16  \\
         T-CPGA (Ours)  & 84.83 & 80.40 & 87.34 & 84.07 & 85.79 & 87.18 & 84.94 
         \\
         \midrule
        \multirow{2}{*}{Method} &  \multicolumn{6}{c}{MobileNet-V2, Cl$\rightarrow$Rw} \\
          \cmidrule(lr){2-8}
        &
         {FLT$\rightarrow$FLT} &
         {FLT$\rightarrow$BLT} &
         {FLT$\rightarrow$Bal} &
         {BLT$\rightarrow$FLT} & 
         {BLT$\rightarrow$BLT} & 
         {BLT$\rightarrow$Bal} & 
         {Avg.} \\
         \midrule
         MobileNet-V2~\cite{sandler2018mobilenetv2} &  40.54  & 30.49  & 36.22  & 28.81  & 46.37  & 38.40  & 36.81 \\
         T-CPGA (Ours)  & 
         82.04 & 85.47 & 86.67 & 83.16 & 83.32 & 86.57 & 84.54  
         \\
         \bottomrule
         \end{tabular}}
    \end{center}
    \end{minipage}
    \hfill
    \begin{minipage}[t]{0.49\textwidth}
    \begin{center}
    \scalebox{0.7}{  
         \begin{tabular}{lcccccccccccccccc}
         \toprule
        \multirow{3}{*}{Method} &  \multicolumn{6}{c}{ResNet-50, Cl$\rightarrow$Pr} \\
        \cmidrule(lr){2-8}
        &
         {FLT$\rightarrow$FLT} &
         {FLT$\rightarrow$BLT} &
         {FLT$\rightarrow$Bal} &
         {BLT$\rightarrow$FLT} & 
         {BLT$\rightarrow$BLT} & 
         {BLT$\rightarrow$Bal} & 
         {Avg.} \\
         \midrule
         ResNet-50~\cite{He2016DeepRL} & 53.88 & 43.93  & 48.19  & 44.51  & 54.26  & 51.39  & 49.36   \\
         T-CPGA (Ours)  & 84.88 & 86.25 & 87.38 & 84.78 & 86.20 & 87.20 & 86.12
         \\
         \midrule
        \multirow{2}{*}{Method} &  \multicolumn{6}{c}{ResNet-50, Cl$\rightarrow$Rw} \\
          \cmidrule(lr){2-8}
        &
         {FLT$\rightarrow$FLT} &
         {FLT$\rightarrow$BLT} &
         {FLT$\rightarrow$Bal} &
         {BLT$\rightarrow$FLT} & 
         {BLT$\rightarrow$BLT} & 
         {BLT$\rightarrow$Bal} & 
         {Avg.} \\
         \midrule
         ResNet-50~\cite{He2016DeepRL} & 54.43 & 44.93  & 50.01  & 45.41  & 58.66  & 54.05  & 51.25  \\
         T-CPGA (Ours)  & 85.16  & 85.79  & 87.15  & 85.00  & 85.87  & 87.03  & 86.00 
         \\         
         \bottomrule
         \end{tabular}}
    \end{center}
    \end{minipage}
\end{table*}

To further verify the effectiveness of T-CPGA in handling imbalance-agnostic UDA, we report Per-Class Accuracy and Overall Accuracy of each task on the Office-Home-I dataset (From Table~\ref{tab:home-perclass} to Table~\ref{tab:home11}) and different imbalance ratios on the VisDA-I dataset (From Table~\ref{tab:visda_10_per} to Table~\ref{tab:visda2}). The results in terms of Overall Accuracy on the DomainNet-S dataset (\ie{Table~\ref{tab:domain1}}) and the intuitive histograms for the VisDA-I dataset in terms of Per-Class Accuracy (\ie{Fig.~\ref{fig:visda_histograms}}) are also reported.
Experimental results show that T-CPGA outperforms all baselines (even with the source data) in terms of Per-Class Accuracy and Overall Accuracy in handling imbalance-agnostic SF-UDA, which fully demonstrates the effectiveness of T-CPGA.

\newl

\noindent{\textbf{Optimization of target-aware classifier.}}
In T-CPGA, we leverage the standard cross-entropy loss to train the additional target label-distribution-aware classifier $G_{t}$. However, there may be concerns regarding how to ensure equal treatment of each category to achieve higher Per-Class Accuracy rather than Overall Accuracy. 
To further explore this, we adopt the balanced softmax loss~\cite{ren2020balanced} and the seesaw loss~\cite{wang2021seesaw} to  train a balanced target-aware classifier and get the variants of T-CPGA (\ie T-CPGA (Bal-CE) and T-CPGA (Seasaw)). 
For the balanced softmax loss, it proposes a meta sampler to explicitly learn the current best sampling rate to prevent the model from overfitting to head (majority) classes or tail (minority) classes. As for the seesaw loss, according to the ratio of accumulated sample numbers, it adjusts the negative sample gradient which is applied to the corresponding category. In this way, it is able to effectively balance the positive and negative sample gradients of different categories, which helps the model treat samples of each class in a balanced way.

As shown in Table~\ref{tab:visda_variant}, the experiments conducted on VisDA-I datasets with varying imbalance ratios, have demonstrated that:
1) compared with the standard cross-entropy loss, the use of the balanced softmax loss or the seesaw loss results in less performance degradation as the imbalance ratio increases. 
2) However, due to the reliance on label frequency to regulate the training process, these two losses are susceptible to pseudo label noise, leading to biased model adaptation. Consequently, they fail to yield significant performance gains when dealing with imbalance agnostic SF-UDA.

\newl

\setlength\tabcolsep{2pt}
\begin{table}[t]
    \begin{center}
    \caption{\label{tab:visda_variant} \textbf{Per-Class Accuracy} (\%) of different class distribution shifts and imbalance ratios on the \textbf{VisDA-I} dataset.}
    \scalebox{0.75}{  
         \begin{tabular}{l|ccccccc}
         \toprule
          \multirow{2}{*}{Method} &  \multicolumn{6}{c}{VisDA-I-10} \\
          \cmidrule(lr){2-8}
        &
         {FLT$\rightarrow$FLT} &
         {FLT$\rightarrow$BLT} &
         {FLT$\rightarrow$Bal} &
         {BLT$\rightarrow$FLT} & 
         {BLT$\rightarrow$BLT} & 
         {BLT$\rightarrow$Bal} & 
         {Avg.} \\
         \midrule
         T-CPGA (CE) & 88.09 & 88.59 & 89.94 & 88.49 & 89.92 & 88.91 & 88.99 \\
         T-CPGA (Bal-CE) & 88.63	&88.86	&89.60	&88.95	&89.64	&88.21	&88.98
         \\
         T-CPGA (Seasaw)  & 88.67	&89.03	&89.82	&88.79	&89.95	&88.84	&89.18
         \\
        \midrule
        \multirow{2}{*}{Method} &  \multicolumn{6}{c}{VisDA-I-50} \\
        \cmidrule(lr){2-8}
        &
         {FLT$\rightarrow$FLT} &
         {FLT$\rightarrow$BLT} &
         {FLT$\rightarrow$Bal} &
         {BLT$\rightarrow$FLT} & 
         {BLT$\rightarrow$BLT} & 
         {BLT$\rightarrow$Bal} & 
         {Avg.} \\
         \midrule
         T-CPGA (CE) & 85.10 & 85.96 & 89.74 & 86.00 & 89.90 & 86.00 & 87.12 \\
         T-CPGA (Bal-CE) & 85.48	&87.03	&89.65	&86.35	&89.69	&85.55	&87.29
         \\
         T-CPGA (Seasaw)  & 85.49	&87.00	&89.80	&86.01	&89.93	&85.86	&87.35
         \\
        \midrule
        \multirow{2}{*}{Method} &  \multicolumn{6}{c}{VisDA-I-100} \\
        \cmidrule(lr){2-8}
        &
         {FLT$\rightarrow$FLT} &
         {FLT$\rightarrow$BLT} &
         {FLT$\rightarrow$Bal} &
         {BLT$\rightarrow$FLT} & 
         {BLT$\rightarrow$BLT} & 
         {BLT$\rightarrow$Bal} & 
         {Avg.} \\
         \midrule
         T-CPGA (CE) & 89.92 & 85.14 & 84.03 & 83.60 & 89.92 & 83.60 & 86.03 \\
         T-CPGA (Bal-CE) & 89.48	&87.48	&83.20	&85.59	&89.88	&86.88	&87.08
         \\
         T-CPGA (Seasaw)  & 89.86	&84.95	&86.00	&85.27	&89.84	&85.46	&86.90
         \\
         \bottomrule
         \end{tabular}}
    \end{center}
\end{table}

\setlength\tabcolsep{2pt}
\begin{table}[t]
    \begin{center}
    \caption{\label{tab:oh_ensemble} {Per-Class} Accuracy (\%) of \textbf{Cl$\rightarrow$Pr} and \textbf{Cl$\rightarrow$Rw}  tasks with {different class distribution shifts} on the \textbf{Office-Home-I} dataset.}
    \scalebox{0.7}{  
         \begin{tabular}{l|ccccccc}
         \toprule
          \multirow{2}{*}{Method} &  \multicolumn{6}{c}{Cl$\rightarrow$Pr} \\
          \cmidrule(lr){2-8}
        &
         {FLT$\rightarrow$FLT} &
         {FLT$\rightarrow$BLT} &
         {FLT$\rightarrow$Bal} &
         {BLT$\rightarrow$FLT} & 
         {BLT$\rightarrow$BLT} & 
         {BLT$\rightarrow$Bal} & 
         {Avg.} \\
         \midrule
          CLIP Zero-shot Prediction & 84.08 & 83.45 & 84.18 & 84.08 & 83.45 & 84.18 & 83.90 \\
         T-CPGA (Combination)  & 85.51 & 84.83 & 86.50 & 85.52 & 84.84 & 86.17 & 85.56    \\
         T-CPGA (Ours) & 85.55 & 84.92 & 86.50 & 85.52 & 84.84 & 86.19 & 85.59  \\
        \midrule
        \multirow{2}{*}{Method} &  \multicolumn{6}{c}{Cl$\rightarrow$Rw} \\
        \cmidrule(lr){2-8}
        &
         {FLT$\rightarrow$FLT} &
         {FLT$\rightarrow$BLT} &
         {FLT$\rightarrow$Bal} &
         {BLT$\rightarrow$FLT} & 
         {BLT$\rightarrow$BLT} & 
         {BLT$\rightarrow$Bal} & 
         {Avg.} \\
         \midrule
         CLIP Zero-shot Prediction & 85.05 & 83.46 & 84.07 & 85.05 & 83.46 & 84.07 & 84.19 \\
         T-CPGA (Combination)  & 85.67 & 83.90 & 85.22 & 85.69 & 83.93 & 85.04 & 84.91    \\
         T-CPGA (Ours) & 85.67 & 83.90 & 85.25 & 85.69 & 83.93 & 85.06 & 84.92  \\
         \bottomrule
         \end{tabular}}
    \end{center}
\end{table}

\noindent{\textbf{{More Discussions on CLIP.}}}
For the fine-tuning of CLIP, we report the results on three types of label distributions (\ie{FLT, BLT and Bal}) regarding the Cl$\rightarrow$Pr task and those regarding the Cl$\rightarrow$Pr task in Figure~\ref{vis:clip_ft_total}. 
Due to the inevitable noise in pseudo labels, the experimental results indicate that fine-tuning CLIP performance degrades in all six tasks, demonstrating that simply fine-tuning CLIP cannot achieve better performance in imbalance-agnostic SF-UDA due to the lack of true target annotations.

Since publicly available CLIP checkpoints are limited to ResNet-50, ViT-B/32, or larger models, we present the experiments of training \ournet~in a small model architecture.
Specifically, we adopt the MobileNet-V2 (pre-trained on ImageNet) as the backbone and evaluate \ournet~on the Cl$\rightarrow$Pr and Cl$\rightarrow$Rw tasks of the \textbf{Office-Home-I} dataset.
From Table~\ref{tab:oh_mobileNet}, our \ournet~is able to achieve competitive performance even with a small-sized backbone, which also demonstrates the effectiveness of the proposed methods in solving  imbalance-agnostic SF-UDA. 

To further explore the use of CLIP in the testing phase, we incorporate CLIP zero-shot prediction into the testing phase (\ie{averaging the outputs of $G_y$, $G_{t}$ and CLIP Zero-shot Prediction $\rightarrow$ T-CPGA (Combination)}) as shown in Table~\ref{tab:oh_ensemble}. Note that in the testing phase of T-CPGA, the input is first passed through the feature extractor $G_e$ and then transmitted separately to both the fixed classifier, $G_y$, and the target label-distribution-aware classifier, $G_{t}$ to obtain the final logit via averaging the output.
Experimental results show that:
1) Compared with the CLIP Zero-shot Prediction (\ie{only CLIP}), the combination of CLIP and T-CPGA achieves better performance. However,
2) compared with our T-CPGA, this variant does not bring additional performance improvement, indicating that \ournet~fully utilized the ability of CLIP when generating pseudo-labels and thus it is no need to integrate CLIP into the testing phase. 

\ifCLASSOPTIONcaptionsoff
  \newpage
\fi

\setlength\tabcolsep{7pt}
\begin{table*}[t]
    \begin{minipage}[t]{0.47\textwidth}
    \begin{center}
    \caption{\label{tab:home-perclass} \textbf{Per-class} Accuracy (\%) on the Office-Home-I dataset (ResNet-50). SF and CI indicate source-free and class-imbalanced.}
    \scalebox{0.6}{  
         \begin{tabular}{lcccccccccccccccc}
         \toprule
         Method & 
        SF & CI
         &  {Cl$\rightarrow$Pr} &
         {Cl$\rightarrow$Rw} &
         {Pr$\rightarrow$Cl} &
         {Pr$\rightarrow$Rw} & 
         {Rw$\rightarrow$Cl} & 
         {Rw$\rightarrow$Pr} & 
         {Avg.} \\
         \midrule
         ResNet-50~\cite{He2016DeepRL} & \xmark & \xmark &49.27	&50.94	&37.19	&66.89	&39.45	&69.22	&52.16
         \\
         DANN~\cite{ganin2015unsupervised} & \xmark & \xmark &56.18	&59.70	&46.48	&71.12	&51.30	&72.53	&59.55
         \\
          MDD~\cite{zhang2019bridging} & \xmark & \xmark &57.55	&61.02	&43.09	&72.41	&45.07	&74.23	&58.89
         \\
         MCC~\cite{jin2020minimum} & \xmark & \xmark &47.54	&50.47	&37.60	&69.60	&40.82	&70.64	&52.78
         \\
         ToAlign~\cite{wei2021toalign} & \xmark & \xmark &62.19	&64.64	&49.34	&75.18	&52.34	&76.44	&63.36
         \\
         \midrule
         COAL~\cite{tan2020class} & \xmark & \cmark &61.71	&64.06	&42.50	&74.93	&45.64	&75.39	&60.70
         \\
         PCT~\cite{tanwisuth2021prototype} & \xmark & \cmark &64.42	&66.01	&49.33	&77.26	&54.81	&78.77	&65.10
         \\
         \midrule
         SHOT~\cite{liang2020shot} & \cmark & \xmark &65.24	&67.58	&50.49	&76.90	&52.97	&76.98	&65.03
         \\
         BAIT~\cite{Yang2020UnsupervisedDA} & \cmark & \xmark &59.65	&60.23	&50.13	&70.82	&53.95	&72.33	&61.19
         \\
         NRC~\cite{yang2021exploiting} & \cmark & \xmark  &67.72	&66.48	&48.06	&74.10	&50.38	&77.65	&64.06
         \\
         CPGA~\cite{Qiu2021CPGA} & \cmark & \xmark &60.56	&63.73	&49.67	&73.66	&53.35	&72.62	&62.26
         \\
         \midrule
         ISFDA~\cite{li2021imbalanced} & \cmark & \cmark &67.31	&68.74	&49.64	&76.51	&53.70	&76.57	&65.41
         \\ 
         T-CPGA (Ours) & \cmark & \cmark & \textbf{85.59} 	&\textbf{84.92} 	&\textbf{60.34} 	&\textbf{84.71} 	&\textbf{60.50} 	&\textbf{85.56} 	&\textbf{76.93} 
 
         \\
 
 
         \bottomrule
         \end{tabular}}
    
    \end{center}
    \end{minipage}
    \hfill
    \begin{minipage}[t]{0.47\textwidth}
    \begin{center}
    \caption{\label{tab:home-overall} \textbf{Overall} Accuracy (\%) on the Office-Home-I dataset (ResNet-50).}
    \scalebox{0.6}{  
         \begin{tabular}{lcccccccccccccccc}
         \toprule
         Method & 
        SF & CI
         &  {Cl$\rightarrow$Pr} &
         {Cl$\rightarrow$Rw} &
         {Pr$\rightarrow$Cl} &
         {Pr$\rightarrow$Rw} & 
         {Rw$\rightarrow$Cl} & 
         {Rw$\rightarrow$Pr} & 
         {Avg.} \\
         \midrule
         ResNet-50~\cite{He2016DeepRL} & \xmark & \xmark &49.36	&51.25	&36.64	&66.90	&39.19	&69.53	&52.15
         \\
         DANN~\cite{ganin2015unsupervised} & \xmark & \xmark &55.30	&58.60	&45.07	&69.85	&49.72	&71.60	&58.36
         \\
                  MDD~\cite{zhang2019bridging} & \xmark & \xmark &58.33	&61.31	&42.47	&72.77	&44.89	&73.99	&58.96
         \\
         MCC~\cite{jin2020minimum} & \xmark & \xmark &47.49	&50.87	&37.24	&68.72	&40.61	&70.69	&52.60
         \\
         ToAlign~\cite{wei2021toalign} & \xmark & \xmark &63.29	&65.49	&49.51	&75.64	&51.90	&76.58	&63.73
         \\
         \midrule
         COAL~\cite{tan2020class} & \xmark & \cmark &61.52	&64.11	&42.14	&75.23	&44.89	&75.56	&60.57
         \\
         PCT~\cite{tanwisuth2021prototype} & \xmark & \cmark &64.58	&66.18	&48.26	&77.00	&53.73	&78.40	&64.69
         \\
         \midrule
         SHOT~\cite{liang2020shot} & \cmark & \xmark &65.64	&68.41	&50.64	&77.83	&52.55	&77.22	&65.38
         \\
         BAIT~\cite{Yang2020UnsupervisedDA} & \cmark & \xmark &60.31	&60.59	&50.80	&71.78	&54.25	&73.29	&61.84
         \\
         NRC~\cite{yang2021exploiting} & \cmark & \xmark  &68.48	&68.19	&48.34	&74.69	&50.27	&78.01	&64.66
         \\
         CPGA~\cite{Qiu2021CPGA} & \cmark & \xmark &61.03	&63.47	&49.51	&73.90	&52.72	&72.98	&62.27
         \\
         \midrule
         ISFDA~\cite{li2021imbalanced} & \cmark & \cmark &67.16	&68.56	&48.84	&76.39	&52.79	&76.10	&64.97
         \\ 
         T-CPGA (Ours) & \cmark & \cmark &\textbf{86.12} 	&\textbf{86.00} 	&\textbf{61.10} 	&\textbf{85.78} 	&\textbf{61.18} 	&\textbf{86.09} 	&\textbf{77.71}
         \\
         \bottomrule
         \end{tabular}}

    \end{center}
    \end{minipage}
\end{table*}

\setlength\tabcolsep{4pt}
\begin{table*}[t]
    \begin{minipage}[t]{0.47\textwidth}
    \begin{center}
    \caption{\label{tab:oh_cl2pr} \textbf{Per-Class} Accuracy (\%) of {Cl$\rightarrow$Pr} task with {different class distribution shifts} on the {Office-Home-I} dataset (ResNet-50). }
    \scalebox{0.6}{  
         \begin{tabular}{lcccccccccccccccc}
         \toprule
         Method & 
        SF & CI
         &  {FLT$\rightarrow$FLT} &
         {FLT$\rightarrow$BLT} &
         {FLT$\rightarrow$Bal} &
         {BLT$\rightarrow$FLT} & 
         {BLT$\rightarrow$BLT} & 
         {BLT$\rightarrow$Bal} & 
         {Avg.} \\
         \midrule
         ResNet-50~\cite{He2016DeepRL} & \xmark & \xmark &53.88	&43.93	&48.19	&44.51	&54.26	&51.39	&49.36

         \\
         DANN~\cite{ganin2015unsupervised} & \xmark & \xmark &65.90	&45.10	&51.50	&43.40	&66.90	&59.00	&55.30

         \\
         MDD~\cite{zhang2019bridging} & \xmark & \xmark &69.41	&48.92	&55.24	&46.32	&68.21	&61.86	&58.33

         \\
         MCC~\cite{jin2020minimum} & \xmark & \xmark &53.28	&42.92	&47.02	&39.11	&54.41	&48.19	&47.49

         \\
         ToAlign~\cite{wei2021toalign} & \xmark & \xmark &69.66	&56.22	&65.40	&52.42	&71.54	&64.50	&63.29

         \\
         
         
         \midrule

         COAL~\cite{tan2020class} & \xmark & \cmark &64.06	&58.74	&63.37	&57.11	&61.81	&64.05	&61.52

         \\
         PCT~\cite{tanwisuth2021prototype} & \xmark & \cmark &67.94	&59.29	&66.97	&55.34	&70.73	&67.24	&64.58

         \\
         \midrule
         SHOT~\cite{liang2020shot} & \cmark & \xmark &69.66	&58.74	&66.50	&56.35	&70.43	&72.18	&65.64

         \\
         BAIT~\cite{Yang2020UnsupervisedDA} & \cmark & \xmark &65.98	&53.20	&61.84	&54.18	&64.84	&61.84	&60.31

         \\
         NRC~\cite{yang2021exploiting} & \cmark & \xmark  &71.77	&64.58	&72.85	&59.43	&69.57	&72.70	&68.48

         \\
         CPGA~\cite{Qiu2021CPGA} & \cmark & \xmark &65.73	&56.17	&60.37	&53.78	&66.00	&64.16	&61.03

         \\
    
         \midrule
         ISFDA~\cite{li2021imbalanced} & \cmark & \cmark &67.59	&66.35	&73.15	&56.75	&68.06	&71.05	&67.16

         \\
        
         T-CPGA (Ours) & \cmark & \cmark & \textbf{85.55} & \textbf{84.92} & \textbf{86.50} & \textbf{85.52} & \textbf{84.84} & \textbf{86.19} & \textbf{85.59}

         \\

         \bottomrule
         \end{tabular}}
    \end{center}
    \end{minipage}
    \hfill
    \begin{minipage}[t]{0.47\textwidth}
    \begin{center}
    \caption{\label{tab:oh_cl2rw} \textbf{Per-Class} Accuracy (\%) of {Cl$\rightarrow$Rw} task with different class distribution shifts on the {Office-Home-I} dataset (ResNet-50).}
    \scalebox{0.6}{  
         \begin{tabular}{lcccccccccccccccc}
         \toprule
         Method & 
        SF & CI
         &  {FLT$\rightarrow$FLT} &
         {FLT$\rightarrow$BLT} &
         {FLT$\rightarrow$Bal} &
         {BLT$\rightarrow$FLT} & 
         {BLT$\rightarrow$BLT} & 
         {BLT$\rightarrow$Bal} & 
         {Avg.} \\
         \midrule
         ResNet-50~\cite{He2016DeepRL} & \xmark & \xmark &50.54	&48.54	&49.27	&52.78	&52.20	&52.32	&50.94

         \\
         DANN~\cite{ganin2015unsupervised} & \xmark & \xmark &62.50	&52.20	&56.70	&57.20	&65.70	&63.90	&59.70

         \\
         MDD~\cite{zhang2019bridging} & \xmark & \xmark &64.73	&53.61	&57.19	&59.53	&62.86	&68.19	&61.02

         \\
         MCC~\cite{jin2020minimum} & \xmark & \xmark &50.38	&47.64	&50.77	&49.67	&54.48	&49.88	&50.47

         \\
         ToAlign~\cite{wei2021toalign} & \xmark & \xmark &64.99	&61.06	&67.67	&62.33	&63.08	&68.73	&64.64

         \\
         \midrule
         COAL~\cite{tan2020class} & \xmark & \cmark &62.58	&61.52	&66.17	&64.31	&63.43	&66.33	&64.06

         \\
         PCT~\cite{tanwisuth2021prototype} & \xmark & \cmark &67.84	&62.15	&69.33	&63.77	&65.17	&67.78	&66.01

         \\
         \midrule
         SHOT~\cite{liang2020shot} & \cmark & \xmark &65.40	&64.44	&72.53	&66.35	&65.44	&71.34	&67.58

         \\
         BAIT~\cite{Yang2020UnsupervisedDA} & \cmark & \xmark &59.08	&57.74	&60.23	&62.21	&61.40	&60.74	&60.23

         \\
         NRC~\cite{yang2021exploiting} & \cmark & \xmark  &63.75	&63.47	&70.76	&62.08	&65.65	&73.17	&66.48

         \\
         CPGA~\cite{Qiu2021CPGA} & \cmark & \xmark &63.14	&62.36	&65.50	&62.37	&63.84	&65.17	&63.73

         \\
         \midrule
         ISFDA~\cite{li2021imbalanced} & \cmark & \cmark &70.76	&66.49	&70.07	&67.49	&67.24	&70.38	&68.74

         \\
         T-CPGA (Ours) & \cmark & \cmark & \textbf{85.67} & \textbf{83.90} & \textbf{85.25} & \textbf{85.69} & \textbf{83.93} & \textbf{85.06} & \textbf{84.92} 


         \\

         \bottomrule
         \end{tabular}}
    \end{center}
    \end{minipage}
\end{table*}

\setlength\tabcolsep{4pt}
\begin{table*}[t]
    \begin{minipage}[t]{0.47\textwidth}
    \begin{center}
    \caption{\label{tab:home6} \textbf{Per-Class} Accuracy (\%) of Pr$\rightarrow$Cl task on the Office-Home dataset (ResNet-50).}
    \scalebox{0.6}{  
         \begin{tabular}{lcccccccccccccccc}
         \toprule
         Method & 
        SF & CI
         &  {FLT$\rightarrow$FLT} &
         {FLT$\rightarrow$BLT} &
         {FLT$\rightarrow$Bal} &
         {BLT$\rightarrow$FLT} & 
         {BLT$\rightarrow$BLT} & 
         {BLT$\rightarrow$Bal} & 
         {Avg.} \\
         \midrule
         ResNet-50~\cite{He2016DeepRL} & \xmark & \xmark &37.48	&34.73	&35.68	&40.74	&37.07	&37.41	&37.19

         \\
         DANN~\cite{ganin2015unsupervised} & \xmark & \xmark &51.10	&38.50	&39.90	&50.10	&54.60	&44.70	&46.48

         \\
         MDD~\cite{zhang2019bridging} & \xmark & \xmark &42.11	&43.07	&39.68	&42.32	&40.40	&50.95	&43.09

         \\         
         MCC~\cite{jin2020minimum} & \xmark & \xmark &39.55	&35.19	&36.27	&40.40	&37.43	&36.77	&37.60

         \\
         ToAlign~\cite{wei2021toalign} & \xmark & \xmark &47.79	&47.51	&50.26	&50.77	&47.80	&51.89	&49.34

         \\
         \midrule

         COAL~\cite{tan2020class} & \xmark & \cmark &45.99	&42.17	&40.77	&45.53	&40.45	&40.09	&42.50

         \\
         PCT~\cite{tanwisuth2021prototype} & \xmark & \cmark &50.94	&47.00	&51.08	&48.48	&47.77	&50.73	&49.33

         \\
         \midrule
         SHOT~\cite{liang2020shot} & \cmark & \xmark &49.22	&46.44	&51.80	&52.92	&49.42	&53.14	&50.49

         \\
         BAIT~\cite{Yang2020UnsupervisedDA} & \cmark & \xmark &49.44	&49.48	&48.98	&51.24	&51.45	&50.20	&50.13

         \\
         NRC~\cite{yang2021exploiting} & \cmark & \xmark  &43.67	&44.40	&53.08	&46.51	&45.27	&55.42	&48.06

         \\
         CPGA~\cite{Qiu2021CPGA} & \cmark & \xmark &53.99	&51.05	&47.18	&53.00	&48.97	&43.82	&49.67

         \\

         \midrule
         ISFDA~\cite{li2021imbalanced} & \cmark & \cmark &52.66	&47.59	&50.43	&49.04	&48.51	&49.63	&49.64

         \\ 
         T-CPGA (Ours) & \cmark & \cmark &\textbf{61.16}	&\textbf{59.04}	&\textbf{60.58}	&\textbf{61.00}	&\textbf{59.76}	&\textbf{60.49}	&\textbf{60.34}



         \\
         \bottomrule
         \end{tabular}}
    
    \end{center}
    \end{minipage}
    \hfill
    \begin{minipage}[t]{0.47\textwidth}
    \begin{center}
\caption{\label{tab:home5} \textbf{Overall} Accuracy (\%) of Pr$\rightarrow$Cl task on the Office-Home dataset (ResNet-50).}
    \scalebox{0.6}{  
         \begin{tabular}{lcccccccccccccccc}
         \toprule
         Method & 
        SF & CI
         &  {FLT$\rightarrow$FLT} &
         {FLT$\rightarrow$BLT} &
         {FLT$\rightarrow$Bal} &
         {BLT$\rightarrow$FLT} & 
         {BLT$\rightarrow$BLT} & 
         {BLT$\rightarrow$Bal} & 
         {Avg.} \\
         \midrule
         ResNet-50~\cite{He2016DeepRL} & \xmark & \xmark &37.36	&34.02	&36.27	&35.10	&39.63	&37.46	&36.64

         \\
         DANN~\cite{ganin2015unsupervised} & \xmark & \xmark &51.60	&34.20	&40.10	&41.50	&58.00	&45.00	&45.07

         \\
         MDD~\cite{zhang2019bridging} & \xmark & \xmark &42.28	&40.51	&40.02	&36.48	&43.95	&51.55	&42.47

         \\         
         MCC~\cite{jin2020minimum} & \xmark & \xmark &40.02	&33.92	&36.01	&34.32	&42.08	&37.11	&37.24

         \\
         ToAlign~\cite{wei2021toalign} & \xmark & \xmark &51.23	&45.33	&50.63	&42.48	&54.47	&52.92	&49.51

         \\
         \midrule

         COAL~\cite{tan2020class} & \xmark & \cmark &44.64	&43.17	&40.37	&40.51	&44.05	&40.09	&42.14

         \\
         PCT~\cite{tanwisuth2021prototype} & \xmark & \cmark &47.98	&45.03	&52.21	&42.38	&50.84	&51.13	&48.26

         \\
         \midrule
         SHOT~\cite{liang2020shot} & \cmark & \xmark &49.16	&46.21	&52.26	&48.18	&54.77	&53.26	&50.64

         \\
         BAIT~\cite{Yang2020UnsupervisedDA} & \cmark & \xmark &50.84	&48.87	&50.36	&44.05	&58.21	&52.46	&50.80

         \\
         NRC~\cite{yang2021exploiting} & \cmark & \xmark  &43.07	&47.49	&53.08	&41.99	&49.75	&54.66	&48.34

         \\
         CPGA~\cite{Qiu2021CPGA} & \cmark & \xmark &53.10	&50.44	&49.05	&47.49	&53.20	&43.78	&49.51

         \\

         \midrule
         ISFDA~\cite{li2021imbalanced} & \cmark & \cmark &49.95	&47.89	&50.58	&44.05	&50.84	&49.71	&48.84

         \\ 
         T-CPGA (Ours) & \cmark & \cmark &\textbf{57.82}	&\textbf{64.01}	&\textbf{61.37}	&\textbf{57.72}	&\textbf{64.41}	&\textbf{61.26}	&\textbf{61.10}

         \\


         \bottomrule
         \end{tabular}}

    \end{center}
    \end{minipage}
\end{table*}

\setlength\tabcolsep{4pt}
\begin{table*}[t]
    \begin{minipage}[t]{0.47\textwidth}
    \begin{center}
    \caption{\label{tab:home8} \textbf{Per-Class} Accuracy (\%) of Pr$\rightarrow$Rw task on the Office-Home dataset (ResNet-50).}
    \scalebox{0.6}{  
         \begin{tabular}{lcccccccccccccccc}
         \toprule
         Method & 
        SF & CI
         &  {FLT$\rightarrow$FLT} &
         {FLT$\rightarrow$BLT} &
         {FLT$\rightarrow$Bal} &
         {BLT$\rightarrow$FLT} & 
         {BLT$\rightarrow$BLT} & 
         {BLT$\rightarrow$Bal} & 
         {Avg.} \\
         \midrule
         ResNet-50~\cite{He2016DeepRL} & \xmark & \xmark &66.29	&66.28	&66.48	&67.15	&67.92	&67.23	&66.89

         \\
         DANN~\cite{ganin2015unsupervised} & \xmark & \xmark &75.50	&65.50	&71.80	&69.10	&71.70	&73.10	&71.12

         \\
         MCC~\cite{jin2020minimum} & \xmark & \xmark &68.06	&68.31	&68.33	&68.42	&72.23	&72.27	&69.60

         \\
         MDD~\cite{zhang2019bridging} & \xmark & \xmark &73.96	&68.82	&76.52	&68.82	&74.70	&71.63	&72.41

         \\         
         ToAlign~\cite{wei2021toalign} & \xmark & \xmark &76.27	&71.55	&76.98	&73.55	&74.92	&77.83	&75.18

         \\
         \midrule

         COAL~\cite{tan2020class} & \xmark & \cmark &75.21	&73.55	&74.76	&75.01	&75.36	&75.66	&74.93

         \\
         PCT~\cite{tanwisuth2021prototype} & \xmark & \cmark &77.61	&76.08	&78.87	&74.65	&77.84	&78.49	&77.26

         \\
         \midrule
         SHOT~\cite{liang2020shot} & \cmark & \xmark &75.65	&75.48	&79.16	&76.13	&75.41	&79.56	&76.90

        \\
         BAIT~\cite{Yang2020UnsupervisedDA} & \cmark & \xmark &73.48	&71.28	&72.09	&70.04	&69.74	&68.31	&70.82

         \\
         NRC~\cite{yang2021exploiting} & \cmark & \xmark  &71.20	&71.36	&78.57	&71.94	&72.94	&78.57	&74.10

         \\
         CPGA~\cite{Qiu2021CPGA} & \cmark & \xmark &74.16	&75.11	&72.11	&73.15	&75.84	&71.59	&73.66

         \\
         \midrule
         ISFDA~\cite{li2021imbalanced} & \cmark & \cmark &75.56	&76.23	&78.74	&74.91	&76.62	&77.02	&76.51

         \\ 
         T-CPGA (Ours) & \cmark & \cmark &\textbf{85.76}	&\textbf{84.20}	&\textbf{85.07}	&\textbf{85.64}	&\textbf{83.92}	&\textbf{83.66}	&\textbf{84.71}

         \\


         \bottomrule
         \end{tabular}}
    
    \end{center}
    \end{minipage}
    \hfill
    \begin{minipage}[t]{0.47\textwidth}
    \begin{center}
    \caption{\label{tab:home7} \textbf{Overall} Accuracy (\%) of Pr$\rightarrow$Rw task on the Office-Home dataset (ResNet-50).}
    \scalebox{0.6}{  
         \begin{tabular}{lcccccccccccccccc}
         \toprule
         Method & 
        SF & CI
         &  {FLT$\rightarrow$FLT} &
         {FLT$\rightarrow$BLT} &
         {FLT$\rightarrow$Bal} &
         {BLT$\rightarrow$FLT} & 
         {BLT$\rightarrow$BLT} & 
         {BLT$\rightarrow$Bal} & 
         {Avg.} \\
         \midrule
         ResNet-50~\cite{He2016DeepRL} & \xmark & \xmark &67.44	&64.49	&67.29	&62.81	&71.11	&68.28	&66.90
         \\
         DANN~\cite{ganin2015unsupervised} & \xmark & \xmark &76.90	&57.30	&71.90	&63.20	&75.40	&74.40	&69.85
         \\
         MDD~\cite{zhang2019bridging} & \xmark & \xmark &76.94	&67.52	&76.86	&63.05	&79.17	&73.08	&72.77
         \\         
         MCC~\cite{jin2020minimum} & \xmark & \xmark &68.87	&66.40	&68.88	&62.09	&72.23	&73.84	&68.72
         \\
         ToAlign~\cite{wei2021toalign} & \xmark & \xmark &79.41	&68.72	&77.60	&68.56	&80.37	&79.18	&75.64
         \\
         \midrule

         COAL~\cite{tan2020class} & \xmark & \cmark &74.86	&73.90	&75.76	&71.67	&78.53	&76.64	&75.23
         \\
         PCT~\cite{tanwisuth2021prototype} & \xmark & \cmark &77.73	&75.02	&79.60	&70.31	&79.81	&79.53	&77.00
         \\
         \midrule
         SHOT~\cite{liang2020shot} & \cmark & \xmark &77.09	&75.42	&80.10	&75.18	&78.45	&80.77	&77.83
        \\
         BAIT~\cite{Yang2020UnsupervisedDA} & \cmark & \xmark &76.86	&69.99	&72.96	&65.76	&75.58	&69.50	&71.78
         \\
         NRC~\cite{yang2021exploiting} & \cmark & \xmark  &70.95	&71.19	&79.62	&70.31	&76.38	&79.69	&74.69
         \\
         CPGA~\cite{Qiu2021CPGA} & \cmark & \xmark &53.99	&51.05	&47.18	&53.00	&48.97	&43.82	&49.67
         \\
         \midrule
         ISFDA~\cite{li2021imbalanced} & \cmark & \cmark &73.98	&76.30	&79.62	&72.23	&78.45	&77.74	&76.39
         \\ 
         T-CPGA (Ours) & \cmark & \cmark &\textbf{85.08}	&\textbf{85.95}	&\textbf{87.03}	&\textbf{85.00}	&\textbf{85.87}	&\textbf{85.77}	&\textbf{85.78}

         \\


         \bottomrule
         \end{tabular}}

    \end{center}
    \end{minipage}
\end{table*}

 \setlength\tabcolsep{4pt}
\begin{table*}[t]
    \begin{minipage}[t]{0.47\textwidth}
    \begin{center}
    \caption{\label{tab:home10} \textbf{Per-Class} Accuracy (\%) of Rw$\rightarrow$Cl task on the Office-Home dataset (ResNet-50).}
    \scalebox{0.6}{  
         \begin{tabular}{lcccccccccccccccc}
         \toprule
         Method & 
        SF & CI
         &  {FLT$\rightarrow$FLT} &
         {FLT$\rightarrow$BLT} &
         {FLT$\rightarrow$Bal} &
         {BLT$\rightarrow$FLT} & 
         {BLT$\rightarrow$BLT} & 
         {BLT$\rightarrow$Bal} & 
         {Avg.} \\
         \midrule
         ResNet-50~\cite{He2016DeepRL} & \xmark & \xmark &41.82	&36.62	&39.16	&42.14	&38.31	&38.67	&39.45

         \\
         DANN~\cite{ganin2015unsupervised} & \xmark & \xmark &58.30	&42.90	&44.90	&51.90	&55.30	&54.50	&51.30

         \\
         MDD~\cite{zhang2019bridging} & \xmark & \xmark &52.67	&44.16	&44.73	&44.46	&41.66	&42.72	&45.07

         \\         
         MCC~\cite{jin2020minimum} & \xmark & \xmark &43.04	&40.01	&40.45	&42.20	&38.47	&40.74	&40.82

         \\
         ToAlign~\cite{wei2021toalign} & \xmark & \xmark &53.31	&50.67	&55.66	&50.83	&50.95	&52.64	&52.34

         \\
         \midrule
         COAL~\cite{tan2020class} & \xmark & \cmark &49.23	&45.19	&43.39	&50.02	&43.72	&42.28	&45.64

         \\
         PCT~\cite{tanwisuth2021prototype} & \xmark & \cmark &58.00	&51.53	&57.03	&53.63	&52.87	&55.83	&54.81

         \\
         \midrule
         SHOT~\cite{liang2020shot} & \cmark & \xmark &55.37	&52.71	&55.45	&53.74	&46.48	&54.06	&52.97

        \\
         BAIT~\cite{Yang2020UnsupervisedDA} & \cmark & \xmark &55.46	&52.91	&54.68	&56.28	&52.50	&51.88	&53.95

         \\
         NRC~\cite{yang2021exploiting} & \cmark & \xmark  &49.83	&45.72	&53.46	&48.71	&47.59	&56.98	&50.38

         \\
         CPGA~\cite{Qiu2021CPGA} & \cmark & \xmark &59.55	&53.71	&51.30	&55.91	&54.00	&45.64	&53.35

         \\
         \midrule
         ISFDA~\cite{li2021imbalanced} & \cmark & \cmark &57.21	&52.80	&54.04	&51.16	&53.02	&53.96	&53.70

         \\ 
         T-CPGA (Ours) & \cmark & \cmark & \textbf{61.27}	&\textbf{59.62}	&\textbf{60.87}	&\textbf{61.11}	&\textbf{59.39}	&\textbf{60.73}	&\textbf{60.50}

         \\


         \bottomrule
         \end{tabular}}
    
    \end{center}
    \end{minipage}
    \hfill
    \begin{minipage}[t]{0.47\textwidth}
    \begin{center}
    \caption{\label{tab:home9} \textbf{Overall} Accuracy (\%) of Rw$\rightarrow$Cl task on the Office-Home dataset (ResNet-50).}
    \scalebox{0.6}{  
         \begin{tabular}{lcccccccccccccccc}
         \toprule
         Method & 
        SF & CI
         &  {FLT$\rightarrow$FLT} &
         {FLT$\rightarrow$BLT} &
         {FLT$\rightarrow$Bal} &
         {BLT$\rightarrow$FLT} & 
         {BLT$\rightarrow$BLT} & 
         {BLT$\rightarrow$Bal} & 
         {Avg.} \\
         \midrule
         ResNet-50~\cite{He2016DeepRL} & \xmark & \xmark &42.87	&35.89	&38.67	&36.87	&42.28	&38.58	&39.19

         \\
         DANN~\cite{ganin2015unsupervised} & \xmark & \xmark &60.00	&38.60	&44.00	&42.80	&59.70	&53.20	&49.72

         \\
         MDD~\cite{zhang2019bridging} & \xmark & \xmark &55.85	&41.69	&44.42	&38.15	&46.71	&42.52	&44.89

         \\         
         MCC~\cite{jin2020minimum} & \xmark & \xmark &43.76	&38.15	&40.50	&35.99	&44.15	&41.12	&40.61

         \\
         ToAlign~\cite{wei2021toalign} & \xmark & \xmark &58.41	&45.62	&54.69	&42.38	&57.62	&52.69	&51.90

         \\
         \midrule

         COAL~\cite{tan2020class} & \xmark & \cmark &49.26	&44.05	&41.97	&46.12	&46.61	&41.33	&44.89

         \\
         PCT~\cite{tanwisuth2021prototype} & \xmark & \cmark &56.83	&48.97	&56.31	&48.87	&55.85	&55.53	&53.73

         \\
         \midrule
         SHOT~\cite{liang2020shot} & \cmark & \xmark &55.46	&50.74	&54.39	&48.28	&52.51	&53.91	&52.55

        \\
         BAIT~\cite{Yang2020UnsupervisedDA} & \cmark & \xmark &60.08	&49.56	&53.97	&50.05	&58.60	&53.26	&54.25

         \\
         NRC~\cite{yang2021exploiting} & \cmark & \xmark  &48.18	&47.59	&53.10	&44.84	&51.23	&56.66	&50.27

         \\
         CPGA~\cite{Qiu2021CPGA} & \cmark & \xmark &59.29	&52.02	&51.64	&51.43	&56.83	&45.11	&52.72

         \\
         \midrule
         ISFDA~\cite{li2021imbalanced} & \cmark & \cmark &54.38	&52.11	&54.04	&46.80	&56.05	&53.38	&52.79

         \\ 
         T-CPGA (Ours) & \cmark & \cmark &\textbf{57.82}	&\textbf{64.50}	&\textbf{61.35}	&\textbf{57.82}	&\textbf{64.21}	&\textbf{61.37}	&\textbf{61.18}

         \\


         \bottomrule
         \end{tabular}}

    \end{center}
    \end{minipage}
\end{table*}

\setlength\tabcolsep{4pt}
\begin{table*}[t]
    \begin{minipage}[t]{0.47\textwidth}
    \begin{center}
    \caption{\label{tab:home12} \textbf{Per-Class} Accuracy (\%) of Rw$\rightarrow$Cl task on the Office-Home dataset (ResNet-50).}
    \scalebox{0.6}{  
         \begin{tabular}{lcccccccccccccccc}
         \toprule
         Method & 
        SF & CI
         &  {FLT$\rightarrow$FLT} &
         {FLT$\rightarrow$BLT} &
         {FLT$\rightarrow$Bal} &
         {BLT$\rightarrow$FLT} & 
         {BLT$\rightarrow$BLT} & 
         {BLT$\rightarrow$Bal} & 
         {Avg.} \\
         \midrule
         ResNet-50~\cite{He2016DeepRL} & \xmark & \xmark &69.98	&69.69	&70.10	&68.98	&67.39	&69.15	&69.22

         \\
         DANN~\cite{ganin2015unsupervised} & \xmark & \xmark &74.50	&69.20	&73.20	&69.10	&75.60	&73.60	&72.53

         \\
         MDD~\cite{zhang2019bridging} & \xmark & \xmark &74.02	&73.35	&77.99	&71.18	&75.32	&73.51	&74.23

         \\         
         MCC~\cite{jin2020minimum} & \xmark & \xmark &77.46	&69.35	&70.77	&69.89	&67.92	&68.46	&70.64

         \\
         ToAlign~\cite{wei2021toalign} & \xmark & \xmark &77.68	&75.65	&78.18	&74.06	&76.29	&76.78	&76.44

         \\
         \midrule

         COAL~\cite{tan2020class} & \xmark & \cmark &74.13	&76.43	&77.30	&74.14	&73.58	&76.73	&75.39

         \\
         PCT~\cite{tanwisuth2021prototype} & \xmark & \cmark &77.15	&78.93	&80.61	&77.91	&77.74	&80.26	&78.77

         \\
         \midrule
         SHOT~\cite{liang2020shot} & \cmark & \xmark &77.14	&76.99	&78.62	&76.34	&74.87	&77.93	&76.98

        \\
         BAIT~\cite{Yang2020UnsupervisedDA} & \cmark & \xmark &72.73	&72.60	&74.08	&72.32	&70.09	&72.16	&72.33

         \\
         NRC~\cite{yang2021exploiting} & \cmark & \xmark  &77.19	&77.14	&80.64	&76.12	&75.18	&79.65	&77.65

         \\
         CPGA~\cite{Qiu2021CPGA} & \cmark & \xmark &74.05	&73.73	&71.82	&72.69	&71.70	&71.70	&72.61

         \\
         \midrule
         ISFDA~\cite{li2021imbalanced} & \cmark & \cmark &76.65	&76.74	&80.02	&74.51	&74.52	&76.98	&76.57

         \\ 
         T-CPGA (Ours) & \cmark & \cmark  & \textbf{85.55} & \textbf{84.83} & \textbf{86.42} & \textbf{85.49} & \textbf{84.81} & \textbf{86.26} & \textbf{85.56}

         \\


         \bottomrule
         \end{tabular}}
    
    \end{center}
    \end{minipage}
    \hfill
    \begin{minipage}[t]{0.47\textwidth}
    \begin{center}
    \caption{\label{tab:home11} \textbf{Overall} Accuracy (\%) of Rw$\rightarrow$Pr task on the Office-Home dataset (ResNet-50).}
    \scalebox{0.6}{  
         \begin{tabular}{lcccccccccccccccc}
         \toprule
         Method & 
        SF & CI
         &  {FLT$\rightarrow$FLT} &
         {FLT$\rightarrow$BLT} &
         {FLT$\rightarrow$Bal} &
         {BLT$\rightarrow$FLT} & 
         {BLT$\rightarrow$BLT} & 
         {BLT$\rightarrow$Bal} & 
         {Avg.} \\
         \midrule
         ResNet-50~\cite{He2016DeepRL} & \xmark & \xmark &70.97	&68.01	&71.28	&64.87	&71.49	&70.53	&69.53

         \\
         DANN~\cite{ganin2015unsupervised} & \xmark & \xmark &77.80	&62.40	&71.60	&63.10	&80.10	&74.60	&71.60

         \\
         MDD~\cite{zhang2019bridging} & \xmark & \xmark &77.97	&68.66	&77.34	&64.82	&80.10	&75.06	&73.99

         \\
         MCC~\cite{jin2020minimum} & \xmark & \xmark &80.95	&65.79	&71.37	&63.26	&72.80	&69.99	&70.69

         \\
         ToAlign~\cite{wei2021toalign} & \xmark & \xmark &81.40	&72.95	&77.58	&69.15	&80.55	&77.81	&76.58

         \\
         \midrule
         COAL~\cite{tan2020class} & \xmark & \cmark &75.25	&74.81	&77.29	&72.33	&75.47	&78.19	&75.56

         \\
         PCT~\cite{tanwisuth2021prototype} & \xmark & \cmark &77.62	&76.78	&80.24	&75.10	&80.30	&80.36	&78.40

         \\
         \midrule
         SHOT~\cite{liang2020shot} & \cmark & \xmark &77.82	&76.27	&78.46	&73.49	&77.82	&79.21	&77.18

        \\
         BAIT~\cite{Yang2020UnsupervisedDA} & \cmark & \xmark &77.22	&71.54	&75.33	&65.88	&76.62	&73.15	&73.29

         \\
         NRC~\cite{yang2021exploiting} & \cmark & \xmark  &76.56	&78.29	&81.21	&73.24	&77.83	&80.92	&78.01

         \\
         CPGA~\cite{Qiu2021CPGA} & \cmark & \xmark &75.66	&72.59	&72.25	&69.20	&75.52	&72.65	&72.98

         \\
         \midrule
         ISFDA~\cite{li2021imbalanced} & \cmark & \cmark &74.95	&76.52	&80.29	&72.13	&75.42	&77.27	&76.10

         \\ 
         T-CPGA (Ours) & \cmark & \cmark  & \textbf{84.98} & \textbf{86.10} & \textbf{87.25} & \textbf{84.83} & \textbf{86.10} & \textbf{87.29} & \textbf{86.09}

         \\


         \bottomrule
         \end{tabular}}

    \end{center}
    \end{minipage}
\end{table*}




\setlength\tabcolsep{4pt}
\begin{table*}[t]
    \begin{minipage}[t]{0.47\textwidth}
    \begin{center}
    \caption{\label{tab:visda_10_per} \textbf{Per-Class} Accuracy (\%) on the VisDA-I-10 dataset (ResNet-101). The number after VisDA-I is the imbalance ratio.}
    \scalebox{0.6}{  
         \begin{tabular}{lcccccccccccccccc}
         \toprule
         Method & 
        SF & CI
         &  {FLT$\rightarrow$FLT} &
         {FLT$\rightarrow$BLT} &
         {FLT$\rightarrow$Bal} &
         {BLT$\rightarrow$FLT} & 
         {BLT$\rightarrow$BLT} & 
         {BLT$\rightarrow$Bal} & 
         {Avg.} \\
         \midrule
         ResNet-50~\cite{He2016DeepRL} & \xmark & \xmark &47.88	&48.91	&48.15	&44.86	&45.73	&45.11	&46.77
         \\
         DANN~\cite{ganin2015unsupervised} & \xmark & \xmark &78.34	&56.73	&66.05	&50.93	&79.11	&66.62	&66.30
         \\
         MDD~\cite{zhang2019bridging} & \xmark & \xmark &60.41	&66.75	&75.41	&53.47	&59.70	&49.41	&60.86
         \\
         MCC~\cite{jin2020minimum} & \xmark & \xmark &80.98	&81.10	&83.24	&72.61	&83.20	&83.38	&80.75
         \\
         ToAlign~\cite{wei2021toalign} & \xmark & \xmark &74.37	&54.98	&58.53	&43.11	&66.61	&53.30	&58.48
         \\
         \midrule

         COAL~\cite{tan2020class} & \xmark & \cmark &61.03	&62.84	&64.39	&59.96	&61.36	&64.28	&62.31
         \\
         PCT~\cite{tanwisuth2021prototype} & \xmark & \cmark &80.35	&78.09	&79.80	&66.22	&82.95	&73.34	&76.79
         \\
         \midrule
         SHOT~\cite{liang2020shot} & \cmark & \xmark &81.14	&79.12	&80.82	&66.26	&46.56	&73.51	&71.23
         \\
         BAIT~\cite{Yang2020UnsupervisedDA} & \cmark & \xmark &81.04	&75.66	&52.22	&64.88	&84.01	&57.51	&69.22
         \\
         NRC~\cite{yang2021exploiting} & \cmark & \xmark &74.44	&70.83	&61.15	&71.47	&73.78	&67.52	&69.86
         \\
         CPGA~\cite{Qiu2021CPGA} & \cmark & \xmark &76.87	&79.93	&85.06	&64.47	&79.69	&81.95	&77.99
        \\
         \midrule
         ISFDA~\cite{li2021imbalanced} & \cmark & \cmark &82.19	&80.75	&82.84	&67.98	&82.74	&74.41	&78.48
         \\ 
         T-CPGA (Ours) & \cmark & \cmark  & \textbf{88.09} & \textbf{88.59} & \textbf{89.94} & \textbf{88.49} & \textbf{89.92} & \textbf{88.91} & \textbf{88.99}
         \\
         \bottomrule
         \end{tabular}}
    
    \end{center}
    \end{minipage}
    \hfill
    \begin{minipage}[t]{0.47\textwidth}
    \begin{center}
        \caption{\label{tab:visda_10_over} \textbf{Overall} Accuracy (\%) on the VisDA-I-10 dataset (ResNet-101). The number after VisDA-I is the imbalance ratio.}
    \scalebox{0.6}{  
         \begin{tabular}{lcccccccccccccccc}
         \toprule
         Method & 
        SF & CI
         &  {FLT$\rightarrow$FLT} &
         {FLT$\rightarrow$BLT} &
         {FLT$\rightarrow$Bal} &
         {BLT$\rightarrow$FLT} & 
         {BLT$\rightarrow$BLT} & 
         {BLT$\rightarrow$Bal} & 
         {Avg.} \\
         \midrule
         ResNet-50~\cite{He2016DeepRL} & \xmark & \xmark &51.39	&43.02	&54.46	&43.96	&44.57	&49.26	&47.78

         \\
         DANN~\cite{ganin2015unsupervised} & \xmark & \xmark &85.66	&37.74	&62.38	&32.24	&83.08	&60.51	&60.27

         \\
         MDD~\cite{zhang2019bridging} & \xmark & \xmark &80.49	&48.76	&73.53	&37.71	&71.25	&46.77	&59.75

         \\         
         MCC~\cite{jin2020minimum} & \xmark & \xmark &80.06	&72.25	&80.39	&73.43	&76.46	&79.75	&77.06

         \\
         ToAlign~\cite{wei2021toalign} & \xmark & \xmark &81.80	&48.88	&61.02	&34.13	&71.64	&52.24	&58.28

         \\
         \midrule
         COAL~\cite{tan2020class} & \xmark & \cmark &61.43	&53.91	&68.97	&61.22	&53.62	&69.07	&61.37
         \\
         PCT~\cite{tanwisuth2021prototype} & \xmark & \cmark &84.64	&67.93	&81.01	&65.60	&80.93	&73.56	&75.61
         \\
         \midrule
         SHOT~\cite{liang2020shot} & \cmark & \xmark &80.65	&69.78	&76.44	&62.18	&43.68	&71.87	&67.43
         \\
         BAIT~\cite{Yang2020UnsupervisedDA} & \cmark & \xmark &84.70	&63.46	&46.80	&61.29	&79.99	&54.93	&65.19
         \\
         NRC~\cite{yang2021exploiting} & \cmark & \xmark &69.47	&62.47	&58.18	&65.88	&65.38	&63.37	&64.12
         \\
         CPGA~\cite{Qiu2021CPGA} & \cmark & \xmark &84.48	&74.04	&82.00	&63.27	&81.22	&79.22	&77.37
 \\
         \midrule
         ISFDA~\cite{li2021imbalanced} & \cmark & \cmark &84.83	&71.91	&79.11	&65.42	&77.67	&73.36	&75.38
         \\ 
         T-CPGA (Ours) & \cmark & \cmark  & \textbf{91.34} & \textbf{89.08} & \textbf{87.89} & \textbf{91.31} & \textbf{87.81} & \textbf{89.29} & \textbf{89.45}
         \\
         \bottomrule
         \end{tabular}}
    \end{center}
    \end{minipage}
\end{table*}

\setlength\tabcolsep{4pt}
\begin{table*}[t]
    \begin{minipage}[t]{0.47\textwidth}
    \begin{center}
    \caption{\label{tab:visda4} \textbf{Per-Class} Accuracy (\%) on the VisDA-I-50 dataset (ResNet-101).}
    \scalebox{0.6}{  
         \begin{tabular}{lcccccccccccccccc}
         \toprule
         Method & 
        SF & CI
         &  {FLT$\rightarrow$FLT} &
         {FLT$\rightarrow$BLT} &
         {FLT$\rightarrow$Bal} &
         {BLT$\rightarrow$FLT} & 
         {BLT$\rightarrow$BLT} & 
         {BLT$\rightarrow$Bal} & 
         {Avg.} \\
         \midrule
         ResNet-50~\cite{He2016DeepRL} & \xmark & \xmark &44.83	&44.87	&44.68	&44.09	&43.65	&43.63	&44.29
         \\
         DANN~\cite{ganin2015unsupervised} & \xmark & \xmark &71.55	&36.71	&43.53	&36.82	&69.01	&43.40	&50.17
         \\
         MDD~\cite{zhang2019bridging} & \xmark & \xmark &64.37	&66.53	&75.45	&31.07	&58.92	&49.41	&57.62
         \\         
         MCC~\cite{jin2020minimum} & \xmark & \xmark &76.23	&62.28	&83.35	&63.95	&79.49	&74.73	&73.34
         \\
         ToAlign~\cite{wei2021toalign} & \xmark & \xmark &71.03	&45.34	&49.24	&38.11	&56.25	&40.24	&50.03
         \\
         \midrule
         COAL~\cite{tan2020class} & \xmark & \cmark &58.54	&60.05	&64.42	&53.28	&64.29	&64.54	&60.85
         \\
         PCT~\cite{tanwisuth2021prototype} & \xmark & \cmark &78.51	&79.28	&74.44	&66.63	&83.24	&77.21	&76.55
         \\
         \midrule
         SHOT~\cite{liang2020shot} & \cmark & \xmark &66.45	&56.48	&64.43	&56.64	&77.42	&68.82	&65.04
         \\
         BAIT~\cite{Yang2020UnsupervisedDA} & \cmark & \xmark &71.68	&56.90	&62.22	&56.17	&75.22	&49.86	&62.01

         \\
         NRC~\cite{yang2021exploiting} & \cmark & \xmark &65.31	&69.08	&70.07	&59.79	&69.72	&63.46	&66.24
         \\
         CPGA~\cite{Qiu2021CPGA} & \cmark & \xmark &75.22	&70.95	&82.30	&57.88	&73.77	&73.42	&72.26

 \\
         \midrule
         ISFDA~\cite{li2021imbalanced} & \cmark & \cmark &70.28	&70.32	&82.16	&63.14	&74.98	&73.93	&72.47
         \\ 
         T-CPGA (Ours) & \cmark & \cmark  & \textbf{85.10} & \textbf{85.96} & \textbf{89.74} & \textbf{86.00} & \textbf{89.90} & \textbf{86.00} & \textbf{87.12}
         \\
         \bottomrule
         \end{tabular}}
    
    \end{center}
    \end{minipage}
    \hfill
    \begin{minipage}[t]{0.47\textwidth}
    \begin{center}
    \caption{\label{tab:visda3} \textbf{Overall} Accuracy (\%) on the VisDA-I-50 dataset (ResNet-101).}
        \scalebox{0.6}{  
         \begin{tabular}{lcccccccccccccccc}
         \toprule
         Method & 
        SF & CI
         &  {FLT$\rightarrow$FLT} &
         {FLT$\rightarrow$BLT} &
         {FLT$\rightarrow$Bal} &
         {BLT$\rightarrow$FLT} & 
         {BLT$\rightarrow$BLT} & 
         {BLT$\rightarrow$Bal} & 
         {Avg.} \\
         \midrule
         ResNet-50~\cite{He2016DeepRL} & \xmark & \xmark &52.11	&33.29	&51.36	&40.27	&42.95	&47.12	&44.52

         \\
         DANN~\cite{ganin2015unsupervised} & \xmark & \xmark &89.11	&10.16	&44.02	&9.68	&84.66	&39.03	&46.11

         \\
         MDD~\cite{zhang2019bridging} & \xmark & \xmark &87.77	&38.76	&73.53	&11.12	&78.24	&46.77	&56.03

         \\         
         MCC~\cite{jin2020minimum} & \xmark & \xmark &74.90	&47.25	&80.73	&56.13	&67.41	&74.65	&66.84

         \\
         ToAlign~\cite{wei2021toalign} & \xmark & \xmark &87.06	&20.89	&53.52	&17.06	&75.17	&38.83	&48.76

         \\
         \midrule

         COAL~\cite{tan2020class} & \xmark & \cmark &60.30	&44.21	&67.40	&46.81	&53.26	&67.91	&56.65

         \\
         PCT~\cite{tanwisuth2021prototype} & \xmark & \cmark &84.32	&68.37	&77.13	&66.94	&80.67	&77.05	&75.75

         \\
         \midrule
         SHOT~\cite{liang2020shot} & \cmark & \xmark &77.95	&30.21	&62.49	&36.09	&69.96	&67.63	&57.39

         \\
         BAIT~\cite{Yang2020UnsupervisedDA} & \cmark & \xmark &84.49	&36.12	&52.27	&31.67	&81.05	&44.07	&54.95

         \\
         NRC~\cite{yang2021exploiting} & \cmark & \xmark &59.90	&54.38	&66.91	&51.96	&62.40	&60.71	&59.38

         \\
         CPGA~\cite{Qiu2021CPGA} & \cmark & \xmark &84.86	&58.98	&79.71	&51.85	&77.30	&73.38	&71.01

 \\
         \midrule
         ISFDA~\cite{li2021imbalanced} & \cmark & \cmark &82.95	&61.53	&77.91	&57.73	&72.02	&72.41	&70.76

         \\ 
         T-CPGA (Ours) & \cmark & \cmark  & \textbf{93.32} & \textbf{89.35} & \textbf{87.75} & \textbf{93.18} & \textbf{87.86} & \textbf{89.95} & \textbf{90.23} \\
         \bottomrule
         \end{tabular}}
    
    \end{center}
    \end{minipage}
\end{table*}

\setlength\tabcolsep{4pt}
\begin{table*}[t]
    \begin{minipage}[t]{0.47\textwidth}
    \begin{center}
        \caption{\label{tab:visda2} \textbf{Per-Class} Accuracy (\%) on the VisDA-I-100 dataset (ResNet-101).}
    \scalebox{0.6}{  
         \begin{tabular}{lcccccccccccccccc}
         \toprule
         Method & 
        SF & CI
         &  {FLT$\rightarrow$FLT} &
         {FLT$\rightarrow$BLT} &
         {FLT$\rightarrow$Bal} &
         {BLT$\rightarrow$FLT} & 
         {BLT$\rightarrow$BLT} & 
         {BLT$\rightarrow$Bal} & 
         {Avg.} \\
         \midrule
         ResNet-50~\cite{He2016DeepRL} & \xmark & \xmark &46.82	&47.32	&47.49	&41.64	&41.12	&41.56	&44.33

         \\
         DANN~\cite{ganin2015unsupervised} & \xmark & \xmark &74.52	&40.55	&47.28	&34.88	&69.56	&41.40	&51.37

         \\
         MDD~\cite{zhang2019bridging} & \xmark & \xmark &63.25	&37.58	&38.60	&37.16	&62.12	&30.38	&44.85

         \\         
         MCC~\cite{jin2020minimum} & \xmark & \xmark &58.15	&60.82	&82.67	&59.56	&54.89	&77.01	&65.52

         \\
         ToAlign~\cite{wei2021toalign} & \xmark & \xmark &69.56	&38.25	&48.69	&35.41	&57.09	&35.88	&47.48

         \\
         \midrule
         COAL~\cite{tan2020class} & \xmark & \cmark &56.25	&59.49	&66.21	&52.06	&58.41	&63.73	&59.36

         \\
         PCT~\cite{tanwisuth2021prototype} & \xmark & \cmark &62.92	&54.62	&75.03	&37.30	&58.18	&65.07	&58.85

         \\
         \midrule
         SHOT~\cite{liang2020shot} & \cmark & \xmark &65.31	&52.21	&64.33	&52.31	&53.80	&66.13	&59.02

         \\
         BAIT~\cite{Yang2020UnsupervisedDA} & \cmark & \xmark &70.50	&54.23	&69.69	&52.60	&64.16	&54.33	&60.92

         \\
         NRC~\cite{yang2021exploiting} & \cmark & \xmark &69.68	&65.12	&85.12	&65.71	&65.57	&76.12	&71.22

         \\
         CPGA~\cite{Qiu2021CPGA} & \cmark & \xmark &59.79	&56.24	&69.28	&53.15	&59.36	&73.51	&61.89

 \\
         \midrule
         ISFDA~\cite{li2021imbalanced} & \cmark & \cmark &68.63	&66.95	&82.35	&73.86	&66.80	&82.25	&73.47

         \\ 
         T-CPGA (Ours) & \cmark & \cmark  & \textbf{89.92} & \textbf{85.14} & \textbf{84.03} & \textbf{83.60} & \textbf{89.92} & \textbf{83.60} & \textbf{86.03}

         \\


         \bottomrule
         \end{tabular}}
    
    \end{center}
    \end{minipage}
    \hfill
    \begin{minipage}[t]{0.47\textwidth}
    \begin{center}
        \caption{\label{tab:visda1} \textbf{Overall} Accuracy (\%) on the VisDA-I-100 dataset (ResNet-101).}
    \scalebox{0.6}{  
         \begin{tabular}{lcccccccccccccccc}
         \toprule
         Method & 
        SF & CI
         &  {FLT$\rightarrow$FLT} &
         {FLT$\rightarrow$BLT} &
         {FLT$\rightarrow$Bal} &
         {BLT$\rightarrow$FLT} & 
         {BLT$\rightarrow$BLT} & 
         {BLT$\rightarrow$Bal} & 
         {Avg.} \\
         \midrule
         ResNet-50~\cite{He2016DeepRL} & \xmark & \xmark &53.52	&37.13	&54.58	&36.36	&42.51	&44.84	&44.82

         \\
         DANN~\cite{ganin2015unsupervised} & \xmark & \xmark &92.06	&11.31	&47.73	&9.33	&89.03	&38.97	&48.07

         \\
         MDD~\cite{zhang2019bridging} & \xmark & \xmark &89.60	&6.80	&36.68	&9.25	&84.00	&29.07	&42.57

         \\
         MCC~\cite{jin2020minimum} & \xmark & \xmark &83.57	&41.82	&80.07	&52.71	&71.10	&74.87	&67.36

         \\
         ToAlign~\cite{wei2021toalign} & \xmark & \xmark &89.66	&10.10	&51.43	&14.63	&79.81	&35.10	&46.79

         \\
         \midrule
         COAL~\cite{tan2020class} & \xmark & \cmark &59.44	&40.58	&70.07	&45.95	&49.25	&67.76	&55.51

         \\
         PCT~\cite{tanwisuth2021prototype} & \xmark & \cmark &83.15	&32.40	&74.99	&35.11	&62.42	&63.59	&58.61

         \\
         \midrule
         SHOT~\cite{liang2020shot} & \cmark & \xmark &82.13	&25.25	&64.11	&33.86	&81.57	&62.47	&58.23

         \\
         BAIT~\cite{Yang2020UnsupervisedDA} & \cmark & \xmark &87.43	&39.50	&65.69	&32.48	&77.84	&52.22	&59.19

         \\
         NRC~\cite{yang2021exploiting} & \cmark & \xmark  &76.33	&53.78	&81.07	&55.09	&74.50	&75.74	&69.42

         \\
         CPGA~\cite{Qiu2021CPGA} & \cmark & \xmark &85.34	&32.65	&69.62	&32.35	&81.71	&68.81	&61.75

         \\

         \midrule
         ISFDA~\cite{li2021imbalanced} & \cmark & \cmark &83.90	&59.55	&78.63	&84.14	&59.57	&78.15	&73.99

         \\ 
         T-CPGA (Ours) & \cmark & \cmark  & \textbf{94.16} & \textbf{90.22} & \textbf{88.00} & \textbf{94.15} & \textbf{90.57} & \textbf{87.85} & \textbf{90.83}

         \\


         \bottomrule
         \end{tabular}}
    
    \end{center}
    \end{minipage}
\end{table*}

\begin{table*}[t]
\setlength\tabcolsep{12.5pt}
    \begin{center}
    \caption{\label{tab:domain1} {Per-class} Accuracy (\%) on the \textbf{DomainNet-S} dataset (ResNet-50). SF and CI indicate source-free and class-imbalanced.}
    \scalebox{0.7}{  
         \begin{tabular}{lcccccccccccccccc}
         \toprule
         Method & 
        SF & CI
         &  {C$\rightarrow$P} &
         {C$\rightarrow$R} &
         {C$\rightarrow$S} &
         {P$\rightarrow$C} & 
         {P$\rightarrow$R} & 
         {P$\rightarrow$S} & 
         {R$\rightarrow$C} &
         {R$\rightarrow$P} &
         {R$\rightarrow$S} &
         {S$\rightarrow$C} & 
         {S$\rightarrow$P} & 
         {S$\rightarrow$R} &
         {Avg.} \\
         \midrule
         ResNet-50~\cite{He2016DeepRL} & \xmark & \xmark &57.15	&76.22	&56.00	&60.76	&84.16	&65.12	&64.31	&69.00	&59.71	&57.81	&56.28	&74.27	&65.07

         \\
         DANN~\cite{ganin2015unsupervised} & \xmark & \xmark&63.90	&82.80	&68.10	&64.80	&78.10	&60.80	&74.80	&73.30	&71.80	&74.50	&66.00	&81.60	&71.71

         \\
         MDD~\cite{zhang2019bridging} & \xmark & \xmark &68.84	&87.87	&73.62	&67.64	&86.59	&69.48	&79.38	&74.91	&74.42	&78.35	&69.84	&85.36	&76.36

         \\         
         MCC~\cite{jin2020minimum} & \xmark & \xmark &52.75	&83.54	&62.46	&66.60	&85.26	&60.64	&69.94	&61.78	&56.73	&67.24	&54.93	&77.41	&66.61

         \\
         ToAlign~\cite{wei2021toalign} & \xmark & \xmark &65.10	&87.78	&73.32	&71.67	&85.98	&73.98	&77.27	&75.78	&74.47	&76.97	&67.91	&84.76	&76.25

         \\
         \midrule
         COAL~\cite{tan2020class} & \xmark & \cmark &68.52	&85.73	&70.95	&72.06	&87.98	&68.04	&75.20	&73.76	&62.48	&77.81	&68.43	&85.21	&74.68

         \\
         PCT~\cite{tanwisuth2021prototype} & \xmark & \cmark &71.34	&89.86	&75.14	&78.17	&89.43	&77.00	&79.73	&77.16	&75.45	&81.26	&72.29	&87.59	&79.53

         \\
         \midrule
         SHOT~\cite{liang2020shot} & \cmark & \xmark &73.83	&89.67	&73.83	&77.81	&89.95	&75.03	&78.25	&75.32	&71.77	&77.31	&70.18	&87.73	&78.39

         \\
         BAIT~\cite{Yang2020UnsupervisedDA} & \cmark & \xmark &75.83	&88.73	&77.05	&77.82	&87.33	&74.63	&76.55	&76.11	&73.68	&82.07	&71.53	&88.48	&79.15

         \\
         NRC~\cite{yang2021exploiting} & \cmark & \xmark &75.32	&91.54	&76.47	&80.49	&91.29	&75.87	&82.55	&75.60	&75.13	&82.05	&77.57	&91.96	&81.32

         \\
         CPGA~\cite{Qiu2021CPGA} & \cmark & \xmark &67.73	&85.01	&65.70	&69.96	&86.14	&68.91	&71.96	&74.19	&63.81	&74.77	&69.12	&84.38	&73.47
 \\
         \midrule
         ISFDA~\cite{li2021imbalanced} & \cmark & \cmark &75.31	&90.10	&76.03	&80.83	&90.39	&72.67	&81.29	&76.32	&72.36	&79.58	&71.37	&87.98	&79.52

         \\ 
         \ournet~(Ours) & \cmark & \cmark &\textbf{80.02}	&\textbf{93.94}	& \textbf{84.41}	&\textbf{89.31}	&\textbf{93.45}	&\textbf{84.60}	&\textbf{89.24}	&\textbf{80.81}	& \textbf{85.43}	& \textbf{89.36}	& \textbf{80.36} & \textbf{93.50}	& \textbf{87.04}

         \\


         \bottomrule
         \end{tabular}}
    \end{center}
\end{table*}

\end{document}